\documentclass[runningheads]{llncs}

\usepackage{eccv}

\usepackage{eccvabbrv}
\usepackage[ruled,vlined]{algorithm2e}
\usepackage{tabularx}
\usepackage{makecell}

\usepackage{graphicx}
\usepackage{booktabs}

\usepackage[accsupp]{axessibility}  

\usepackage{amsmath}
\usepackage{amssymb}
\usepackage{multirow}
\usepackage{adjustbox}
\usepackage{xcolor}
\usepackage{comment}
\usepackage{subcaption}
\usepackage{wrapfig}
\usepackage{caption}
\usepackage{float}
\usepackage{placeins}

\AtBeginDocument{%
  \setlength{\textfloatsep}{6pt plus 2pt minus 2pt}%
  \setlength{\intextsep}{6pt plus 2pt minus 2pt}%
  \setlength{\floatsep}{6pt plus 2pt minus 2pt}%
  \setlength{\dbltextfloatsep}{6pt plus 2pt minus 2pt}%
  \setlength{\dblfloatsep}{6pt plus 2pt minus 2pt}%
  \setlength{\abovecaptionskip}{7pt}%
  \setlength{\belowcaptionskip}{0pt}%
}

\usepackage{hyperref}

\usepackage{orcidlink}

\begin{document}

\title{EquiSteer: Cross-Attention Steering Towards a Fairer Text-Guided Image Generation}

\titlerunning{EquiSteer: Cross-Attention Steering for Fairer T2I Generation}

\author{Tatiana Gaintseva\inst{1,2} \and
Akshit Achara\inst{3} \and
Gregory Slabaugh\inst{1} \and
Jiankang Deng\inst{4} \and
Ismail Elezi\inst{2}}

\authorrunning{T.~Gaintseva et al.}

\institute{Queen Mary University of London, UK \and
Huawei Noah's Ark, UK \and
King's College London, UK \and
Imperial College London, UK}

\addtocontents{toc}{\protect\setcounter{tocdepth}{-100}}

\maketitle

\begin{figure}
    \centering
    
    \includegraphics[width=\linewidth]{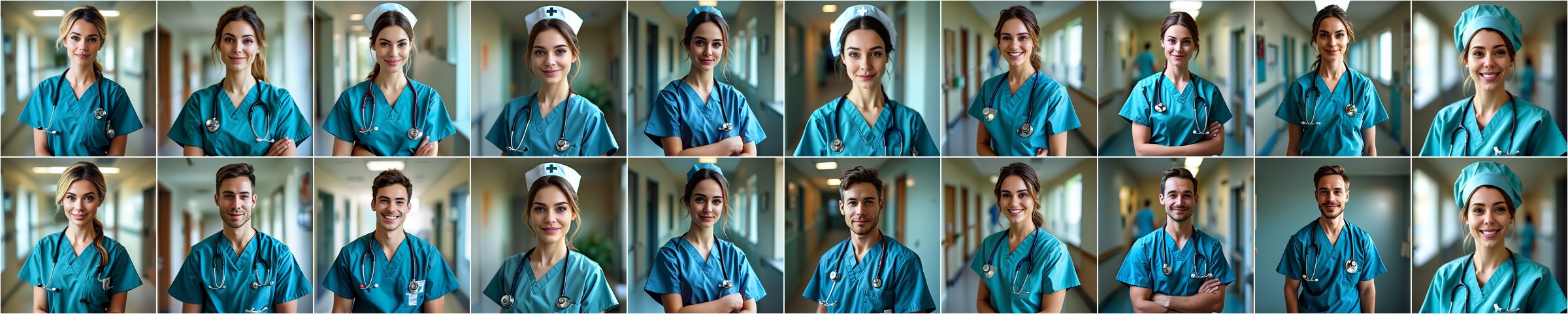}

    \vspace{0.3em}

    \includegraphics[width=\linewidth]{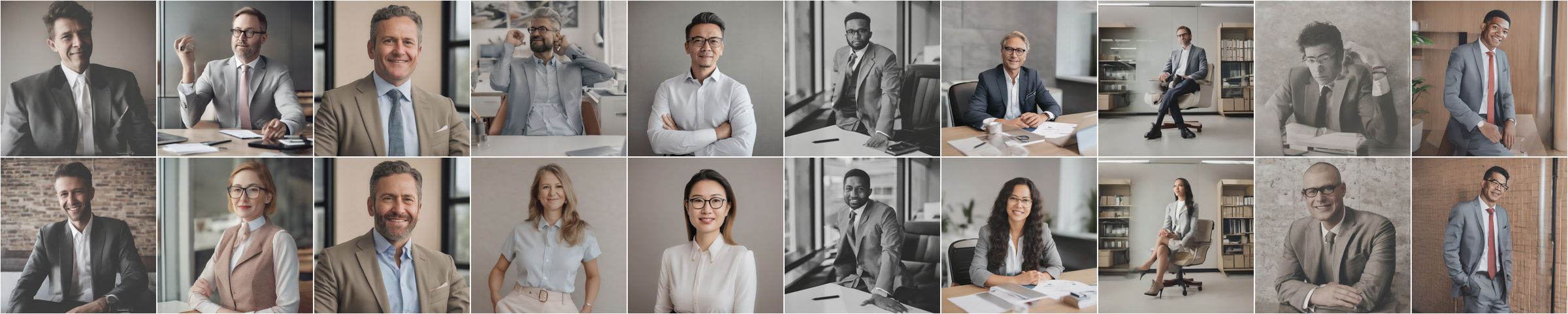}

    \caption{Examples of EquiSteer for debiasing the \textit{gender} concept. Top block: generations for the prompt ``A photo of a nurse'' with SANA, bottom block: ``A photo of a CEO'' with SDXL. In both cases, the top row corresponds to the vanilla model and the bottom row to EquiSteer, with ten generation seeds shown for each prompt.}
    \label{fig:teaser}
\end{figure}

\vspace{-1cm}

\begin{abstract}

Text-to-image diffusion models power everyday creative tasks, but they still reproduce the demographic biases in their training data.
On common prompts such as ``a photo of a nurse,'' ``a photo of a CEO'', they skew their outputs toward one gender, driven by the statistics of training data rather than anything in the text.
Existing debiasing methods show promise in narrow settings but require retraining, batch-level control, or prompt-specific tuning, limiting their scalability.
We propose \emph{EquiSteer}, a training-free method that works per sample by steering cross-attention (CA) activations at inference time.
For each target attribute, EquiSteer precomputes steering vectors from contrastive prompts. Then at generation time, a prompt-aware gate leaves attribute-specific prompts untouched, while for neutral ones it clears existing attribute signals from the CA activations and injects a target attribute.
Across SD-1.5, SD-2.1, SDXL, and SANA, EquiSteer reduces the average parity gap by up to $87\%$, with minimal effect on image quality and text-image alignment.
Code is available at \href{https://github.com/Atmyre/EquiSteer}{https://github.com/Atmyre/EquiSteer}.%
\end{abstract}


\section{Introduction}
\label{sec:intro}

Text-to-image diffusion models have become standard tools for visual content creation, but they carry a well-documented problem: they reproduce and amplify the demographic skews in their training data.
The effect is easy to observe.
Ask a state-of-the-art model for ``a photo of a nurse'' and it will almost always generate a female figure; ask for ``a CEO'' and you get a male one (Fig.~\ref{fig:teaser}).
These are not occasional glitches but stable patterns, and as these models find their way into more and more real-world applications, they can further propagate discrimination and raise ethical concerns regarding the fairness and inclusivity of generative AI systems.

To mitigate this, researchers have proposed several approaches for debiasing text-to-image (T2I) generation. 
Finetuning-based methods~\cite{friedrich2023fair,shen2023finetuning,li2025fair} can be effective but require access to model weights and additional data, which rules them out for many users and makes cross-architecture transfer difficult.
Guidance-based methods~\cite{parihar2024balancing} are training-free but operate at the batch level, using external classifiers to steer attribute distributions; they cannot debias individual samples and struggle when multiple attributes are involved.
Text-embedding methods~\cite{kim2025rethinking} modify prompt representations to suppress demographic cues, but are sensitive to prompt phrasing and offer limited spatial control over the generated output.

We instead look further inside the model. Cross-attention (CA) layers are where text tokens meet image features: they govern which spatial regions respond to which parts of the prompt, and prior work has shown they can be steered at inference time to control semantic content~\cite{hertz2022prompt,chefer2023attend,gaintseva2025casteer}.
Recent attribution analysis further finds that demographic information is encoded in these activations~\cite{chakraborty2025biasmap}, pointing to a direct and unexploited debiasing handle.
We exploit this by precomputing, for each attribute value, a \emph{steering vector} in cross-attention (CA) activation space from pairs of contrastive prompts, and at inference time uniformly sampling a target attribute and adding its steering vector to the CA outputs.
This simple mechanism is effective at reducing bias in many settings. Simple steering, however, has two failure modes.
When a prompt explicitly specifies an attribute (e.g., ``a photo of a male nurse''), the intervention should not apply, but a fixed steering strength cannot detect this. And since CA activations may already carry pre-existing attribute signals, adding a steering vector on top can yield ambiguous, mixed-attribute generations.
Thus, we introduce \emph{EquiSteer}. 
It contains a prompt-aware gate that measures how strongly early-step CA activations align with the attribute steering direction, and skips intervention when the signal exceeds a calibrated threshold.
For neutral prompts, it first orthogonalises the CA output against all attribute directions to remove pre-existing signals, then calibrates the injection strength from attribute-specific prompts and injects the target attribute.

\begin{figure}
    \centering
    
    \includegraphics[width=0.95\linewidth]{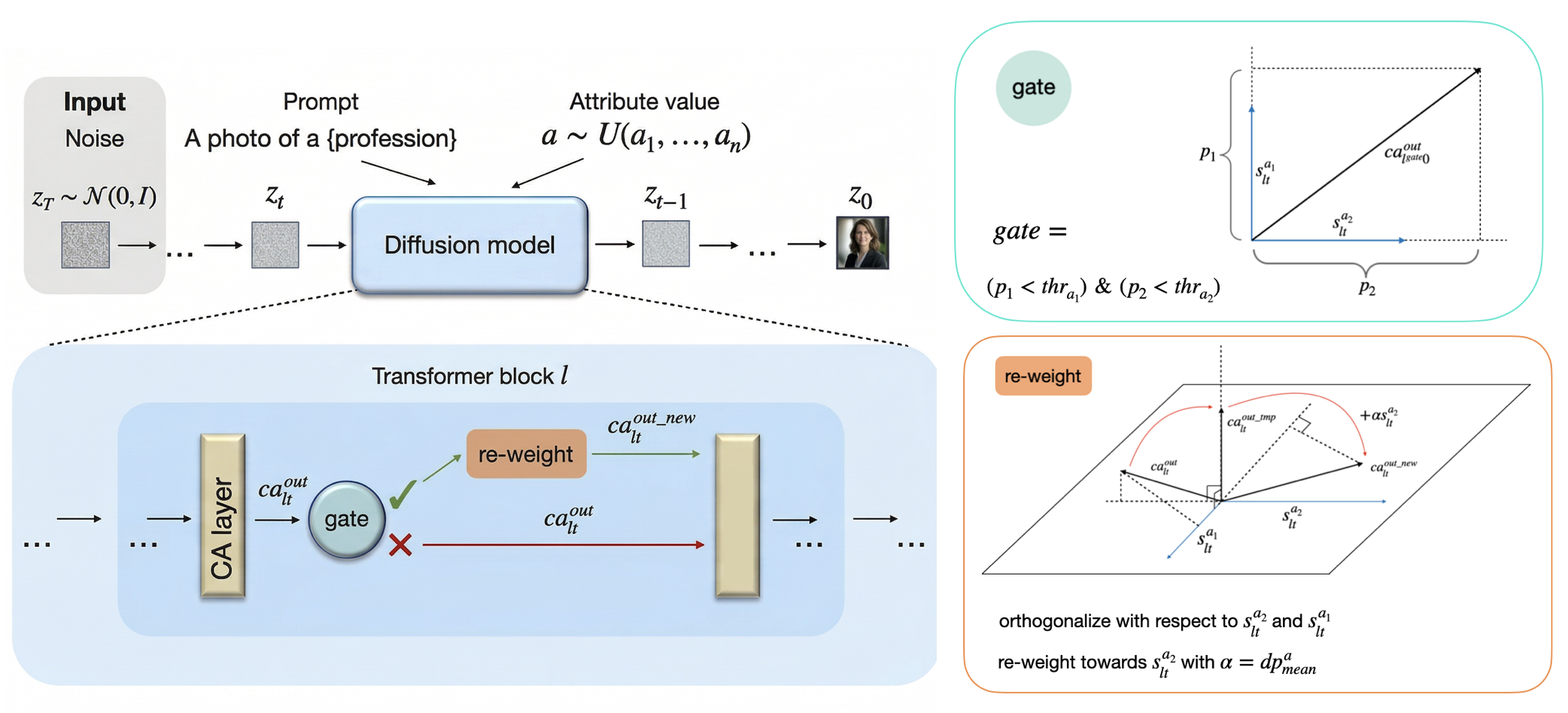}

    \caption{\noindent\textbf{Illustration of the proposed method, EquiSteer.}
At each denoising step $t$ and block $l$, EquiSteer modifies the Cross-Attention (CA) output to promote a desired attribute $a$. 
EquiSteer combines three components: (i) a \emph{gate} that estimates attribute presence in the CA output and disables debiasing for attribute-specific prompts, (ii) \emph{orthogonalisation} of the CA output w.r.t.\ $\mathrm{span}\!\left(\{s^{a_i}_{lt}\}_{i=1}^{n}\right)$ to remove pre-existing attribute signal, and (iii) \emph{re-weighting} that calibrates the shift magnitude so the target attribute is sufficiently expressed.}
    \label{fig:pipeline}
\end{figure}

We evaluate EquiSteer on four architectures (SD-1.5, SD-2.1~\cite{Rombach_2022_CVPR}, SDXL~\cite{podell2023sdxl}, and SANA~\cite{xie2024sana}), covering gender, race, and age as primary concepts, with additional experiments on categorical attributes such as eyeglasses.
EquiSteer consistently outperforms recent debiasing methods, reducing the average parity gap by up to $87\%$ relative to the undebiased model and by up to $47\%$ over the strongest available baseline.
It correctly handles attribute-specific prompts throughout and maintains competitive image quality and text-image alignment, confirming that cross-attention steering offers a practical and scalable route to fairer text-guided image generation (Fig.~\ref{fig:pipeline}).

\noindent In summary, our contributions are:
\begin{itemize}
    \item We show that CA activations encode demographic attributes in a way that makes them directly steerable for fairness control, establishing cross-attention steering as an effective approach to inference-time T2I debiasing.
    \item We introduce \textbf{EquiSteer}, a CA-steering debiasing method with prompt-aware gating, and attribute-space orthogonalisation to handle attribute-specific prompts and prevent mixed-attribute artifacts, all without changing model weights.
    \item We validate EquiSteer across four T2I backbones and multiple demographic concepts, showing large-margin improvements over recent baselines while preserving attribute-specific behavior and image quality.
\end{itemize}
\section{Related Work}

\paragraph{Auditing bias in text-to-image diffusion.}
Large-scale audits show that text-to-image (T2I) models reproduce and amplify demographic stereotypes across occupations and everyday scenes, with systematic skew toward whiteness and masculinity and under-representation of minority groups~\cite{luccioni2023stable,girrbach2025large,wu2025revealing}. Recent evaluations also link toxicity and unsafe generations with disparate impacts across demographics~\cite{schneider2025investigating}. Common practice quantifies representation parity and stereotype amplification using classifier-based (often CLIP-based~\cite{radford2021learning}) attribute or occupation predictors, with complementary human checks~\cite{d2024openbias,chinchure2024tibet,seshadri2024bias}. These findings motivate mitigation methods that work across attributes and models rather than narrow, per-task fixes. Building on these observations, we introduce a training-free, per-sample intervention: EquiSteer targets the documented skews by acting directly on CA activations without auxiliary predictors or retraining, while preserving text–image alignment.

\paragraph{Finetuning-based debiasing.}
One line of work adapts model parameters using additional supervision or balanced datasets. \emph{Fair Diffusion} instructs models toward fairness objectives~\cite{friedrich2023fair}. In~\cite{shen2023finetuning}, the authors propose distributional alignment losses and adjusted direct finetuning to reduce gender and racial bias in occupations. \emph{Fair Mapping} introduces a lightweight parameter-efficient modification that remaps intermediate features of pre-trained diffusion models~\cite{li2025fair}. While effective, they require retraining and model access, limiting scalability and transferability across diffusion backbones.
In contrast, EquiSteer operates entirely at inference time, requiring no additional data or weight updates, and can be applied selectively on a per-generation basis, working with different backbones without retraining.

\paragraph{Distributional guidance.}
Training-free guidance methods steer generation toward target attribute distributions. \emph{Balancing Act}~\cite{parihar2024balancing} achieves this by using an attribute distribution predictor (ADP) to jointly guide batches of samples toward balanced attribute marginals. However, it operates only at the batch level and faces scalability limits for general-purpose debiasing, especially under fine-grained or multi-attribute settings. Beyond its batch-only operation, reliance on external attribute estimators and sensitivity to guidance strength can introduce quality–fairness trade-offs that complicate robust deployment~\cite{parihar2024balancing,he2024debiasing}. EquiSteer instead operates at the sample level (no batch coupling), avoids external attribute estimators, and scales to multi-attribute settings with reduced quality–fairness trade-offs.

\paragraph{Text embedding intervention (TEI).}
Prompt- and embedding-level methods adjust or neutralize text embeddings to mitigate bias~\cite{kim2025rethinking,na2025diffusion}. These approaches are lightweight and often model-agnostic, but can be sensitive to phrasing and compositional prompts and offer limited spatial control during generation~\cite{he2024debiasing,friedrich2023fair}. 
LightFair~\cite{han2025lightfair} debiases T2I generation by refining the pre-trained text encoder (text-conditioning side), offering a more efficient alternative to full diffusion finetuning, but still requiring weight access and an optimization stage. FairImagen~\cite{fu2025fairimagen} performs post-hoc debiasing by projecting CLIP prompt embeddings into a fairness-aware subspace (FairPCA), optionally combined with noise injection and a unified projection for multi-attribute debiasing; while lightweight and model-agnostic, such embedding-level methods provide limited spatial control compared to cross-attention interventions.
Rather than modifying prompts or text embeddings, EquiSteer manipulates CA activations, providing spatially aware control that is robust to phrasing and compositional prompts.

\paragraph{Cross-attention control and interpretability.}
Cross-attention (CA) mechanisms have been leveraged for layout control and semantic editing~\cite{gaintseva2025casteer,chen2024training}, as well as for bias attribution and interpretability~\cite{chakraborty2025biasmap,liu2024towards}. These findings identify CA as a powerful interface for fine-grained, training-free control over generative behavior. Related training-free attention steering for controllable editing (e.g., Prompt-to-Prompt and Attend-and-Excite) demonstrates that manipulating cross-attention supports fine-grained, per-sample control and stronger word–region binding~\cite{hertz2022prompt,chefer2023attend}. Building on these controls, EquiSteer turns CA interpretability into a fairness mechanism: it detects when to abstain, removes confounding attribute directions, and re-injects a calibrated target attribute without editing prompts or weights.



\section{Method}


Our approach builds on CASteer~\cite{gaintseva2025casteer}, which suppresses target concepts in diffusion models by modifying cross-attention (CA) layer outputs with precomputed steering vectors.
We repurpose this mechanism for debiasing: instead of suppressing a concept, we use steering vectors to shift CA outputs toward desired attributes.
We review the CA steering mechanism in Sec.~\ref{sec:background} and formalize the debiasing task in Sec.~\ref{sec:task_statement}.
Sec.~\ref{sec:fair_steer_base} introduces our base debiasing method where we show that basic steering has two failure modes, motivating a gating mechanism (Sec.~\ref{sec:dot_product_indicator}) and an adaptive steering magnitude (Sec.~\ref{sec:steering_strength}); these two additions form our proposed framework, EquiSteer.

\subsection{Background: Steering for Generation Control}
\label{sec:background}

We briefly review the CA steering mechanism of CASteer~\cite{gaintseva2025casteer}, which enables inference-time concept manipulation in diffusion models.
The core idea is to modify CA layer outputs using precomputed \emph{steering vectors} associated with a target concept $X$.

For each concept $X$, CASteer constructs steering vectors for every cross-attention layer $1 \leq l \leq L$ and denoising step $1 \leq t \leq T$.
These vectors are computed from pairs of positive prompts (containing $X$) and negative prompts (excluding $X$).
Details of the construction procedure are provided in the supplementary.

Let $s_{lt}^X$ denote the steering vector for concept $X$, layer $l$, and timestep $t$.
Each $s_{lt}^X$ shares the dimensionality of the CA output and applies directly as an additive intervention during inference.
Denote by $ca^{out}_{lt}$ the output of the $l$-th CA layer at denoising step $t$.
Steering modifies the activations as
\begin{equation}
    ca^{out\_new}_{lt} = ca^{out}_{lt} + \alpha s^X_{lt}.
\end{equation}

Here, $\alpha \in \mathbb{R}$ controls the strength and direction of intervention.
The vector $s^X_{lt}$ encodes a direction in CA activation space associated with concept $X$:
adding $\alpha s^X_{lt}$ with $\alpha > 0$ amplifies the presence of $X$ in the generation, while $\alpha < 0$ suppresses it.

CASteer demonstrated that such activation-level interventions enable effective concept suppression without retraining.
In this work, we build on this mechanism and show that cross-attention steering extends naturally from concept suppression to controlled debiasing.

\subsection{Debiasing Task Statement}
\label{sec:task_statement}

We consider the task of \emph{training-free, inference-time debiasing} of a diffusion model with respect to a concept $X$ that has $n$ discrete attributes $\{a_i\}_{i=1}^n$.
For example, when $X=\textit{gender}$, we have $(a_1, a_2) = (\textit{male}, \textit{female})$.

Our goal is to design an inference-time intervention that adjusts the generative process so that the resulting images exhibit attributes $\{a_i\}_{i=1}^n$ according to a prescribed target distribution $p(a_1, \dots, a_n)$.
In practice, we choose a uniform target $p = U(a_1, \dots, a_n)$, so that each attribute is generated with probability $1/n$.
For instance, in the binary gender setting, this means generating male and female images with equal probability.

The intervention must also operate selectively.
We aim to modify generations from \emph{attribute-neutral prompts} (e.g., ``a photo of a doctor''), while leaving the model's behavior on \emph{attribute-specific prompts} (e.g., ``a photo of a male doctor'') unchanged.
The objective is to enforce the desired marginal attribute distribution for neutral prompts, without altering generations where the prompt already specifies an attribute.

\subsection{Debiasing via Cross-Attention Steering}
\label{sec:fair_steer_base}

We now present our inference-time debiasing method based on CA steering. Consider a diffusion model with $L$ CA layers and $T$ denoising steps, and a target concept $X$ with discrete attribute set $\{a_i\}_{i=1}^n$. For each attribute $a_i$, we precompute steering vectors $s_{lt}^{a_i}$ for all CA layers $l$ and timesteps $t$. Each steering vector $s_{lt}^{a}$ captures attribute-specific information in CA activation space, so adding $s_{lt}^{a}$ to the CA output $ca^{out}_{lt}$ biases the generation toward attribute $a$. This motivates the following debiasing mechanism:

\medskip
\noindent\textbf{Basic steering mechanism for debiasing.}
At inference time, we sample a target attribute
$a \sim \mathrm{Uniform}\{a_i\}_{i=1}^n, $
and then modify each CA output according to
\begin{equation}
    ca^{out\_new}_{lt} = ca^{out}_{lt} + \alpha s_{lt}^{a},
    \label{eq:fairsteer_base}
\end{equation}
where $\alpha > 0$ controls the steering strength. The CA output $ca^{out}_{lt}$ has shape $(bs, \text{seq\_len}, \text{emb\_dim})$, corresponding to batch size, number of image tokens, and embedding dimension. In Eq.~\ref{eq:fairsteer_base}, the same steering vector $\alpha s_{lt}^{a}$ is added to each image token, shifting the activations toward the selected attribute direction and thereby increasing the likelihood that attribute $a$ appears in the final image.

Finally, for stability we re-normalize the updated CA output to preserve its original $L_2$ norm:
\begin{equation}
ca^{out\_new\_renormed}_{lt} = \|ca^{out}_{lt}\|_2\,\frac{ca^{out\_new}_{lt}}{\|ca^{out\_new}_{lt}\|_2}.
\label{eq:renorm_base}
\end{equation}

\medskip



\begin{wrapfigure}{l}{0.5\textwidth}
    \centering
    \vspace{-10pt}
    \includegraphics[width=0.48\textwidth]{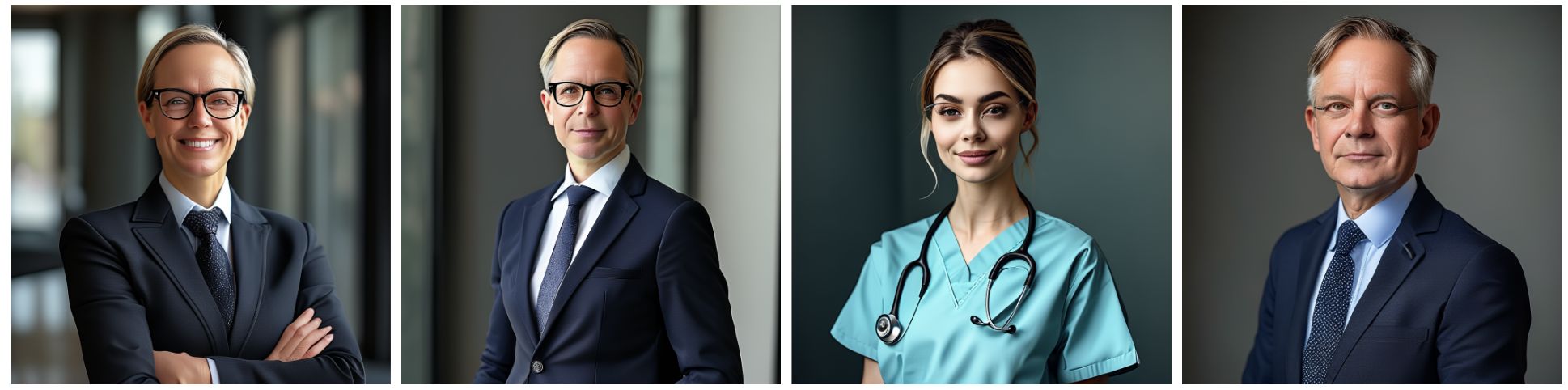}
    \caption{Failure cases of the basic steering mechanism for \textit{gender} (left) and \textit{eyeglasses} (right).}
    \label{fig:base_fail}
    \vspace{-11pt}
\end{wrapfigure}

\noindent\textbf{However, the basic mechanism is insufficient.}
We aim to debias \emph{attribute-neutral} prompts while preserving generations for \emph{attribute-specific} prompts. A single global $\alpha$ cannot satisfy both goals: small $\alpha$ may be too weak to counteract pre-existing bias (Fig.~\ref{fig:base_fail}, right), while large $\alpha$ may override explicitly specified attributes. Moreover, when multiple attribute signals are present, direct addition can lead to mixed or ambiguous generations (Fig.~\ref{fig:base_fail}, left).

These limitations motivate two additional components. First, we introduce a gating mechanism that determines whether to apply debiasing at all. Second, we refine the steering update by removing existing attribute directions from the activation and setting the steering magnitude adaptively.

\subsection{Gated Steering}
\label{sec:dot_product_indicator}

We first introduce a gating mechanism that determines whether debiasing applies to the current prompt. CA activations already encode whether an attribute is explicitly specified in the prompt, and we exploit this to build an attribute-specificity detector.

\begin{figure}[t]
    \centering
    \begin{subfigure}[t]{0.42\textwidth}
        \centering
        \includegraphics[width=\textwidth]{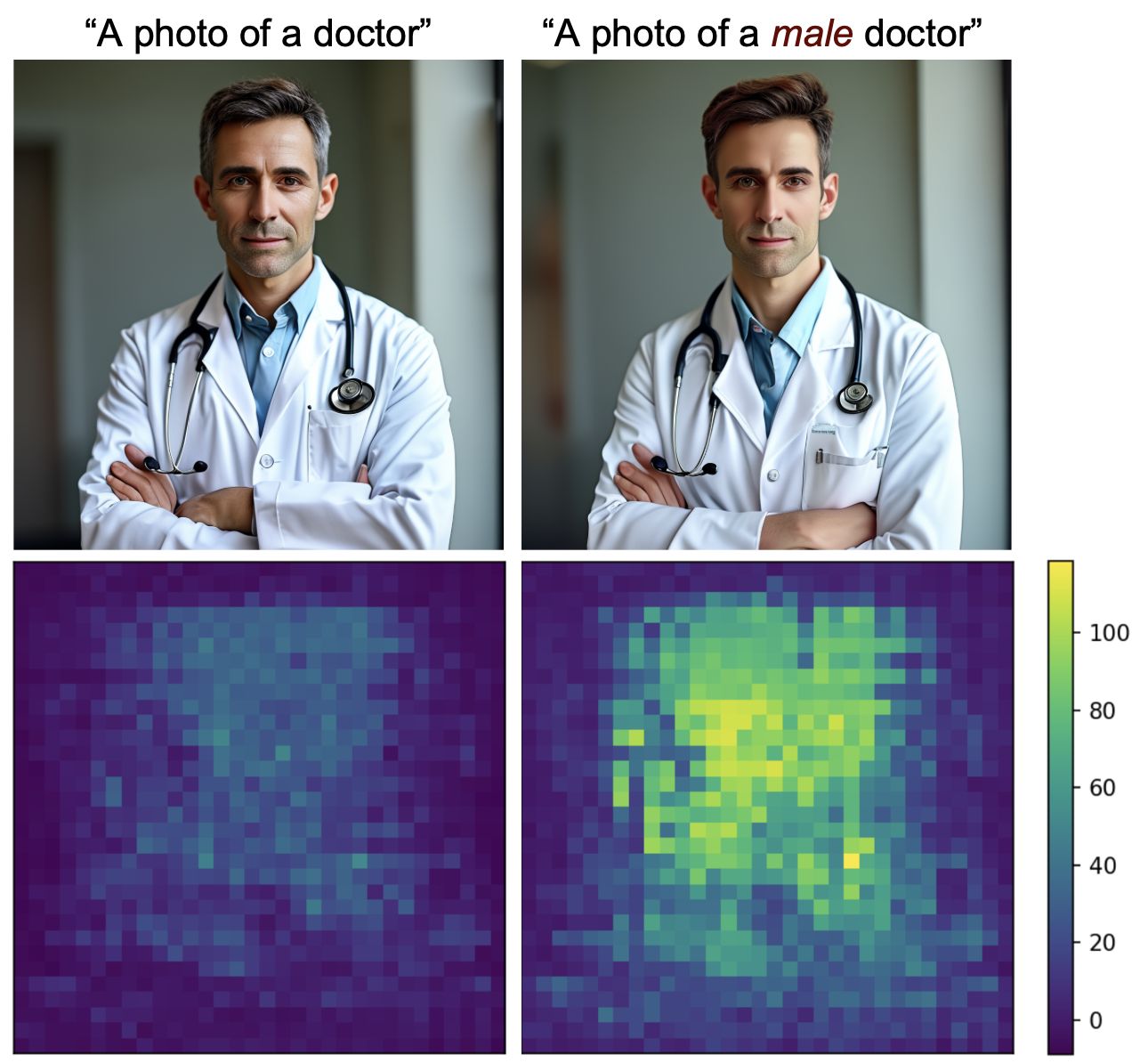}
        \caption{Example generated images and corresponding dot-product heatmaps (SANA).}
        \label{fig:ca_maps}
    \end{subfigure}
    \hfill
    \begin{subfigure}[t]{0.52\textwidth}
        \centering
        \includegraphics[width=\textwidth]{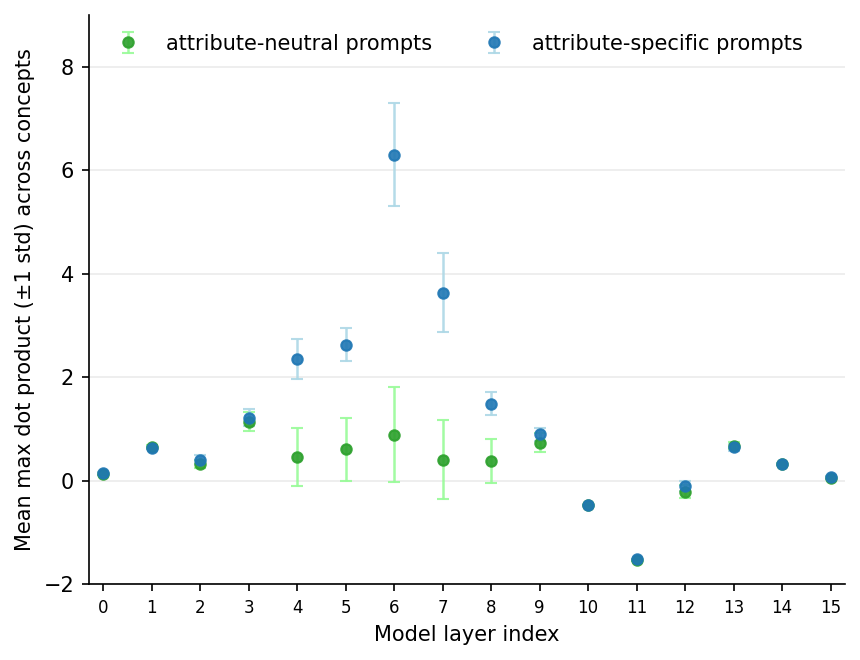}
        \caption{Layer-wise maximal dot-product statistics for SD-1.5 at $t=0$.}
        \label{fig:dot_prods_sd15}
    \end{subfigure}
    \caption{Dot-product statistics between cross-attention outputs and the steering vector of the \textit{male} attribute. \textbf{(a)} For SANA, attribute-specific prompts induce substantially stronger dot-product responses than attribute-neutral prompts, as seen in the heatmaps of layer $l=5$ at denoising step $t=0$. \textbf{(b)} For SD-1.5, the maximal dot-product statistic separates attribute-specific from attribute-neutral prompts in intermediate layers, with the clearest gap around layers 4 to 6. This motivates its use as a signal for the gating mechanism.}
    \label{fig:gating_combined}
    \vspace{-15pt}
\end{figure}

\medskip
\noindent\textbf{Attribute expression as a dot-product signal.}
For a target attribute $X$, let $ca^{out}_{ltk}$ denote the CA output for image token $k$ at layer $l$ and denoising step $t$, and let $s_{lt}^{X}$ be the corresponding steering vector. Since $s_{lt}^{X}$ encodes a direction associated with attribute $X$, the dot product $\langle ca^{out}_{ltk}, s_{lt}^{X} \rangle$
measures how strongly attribute $X$ is expressed in that token. Figure~\ref{fig:ca_maps} illustrates this: for attribute-specific prompts, the dot products are substantially higher than for attribute-neutral ones.
To obtain a layer-level statistic, we compute the maximal token response
\begin{equation}
    dp_{lt} = \max_{0 \leq k \leq K_l} \langle ca^{out}_{ltk}, s_{lt}^{X} \rangle,
    \label{eq:3}
\end{equation}
where $K_l$ is the number of image tokens in layer $l$. We use the maximum because attribute evidence may be spatially localized, and $dp_{lt}$ captures the strongest attribute signal at that layer.

\medskip
\noindent\textbf{CA layers capture attribute specificity in the prompt.} We use Eq.~\ref{eq:3} to detect attribute-specific prompts.
We evaluate $dp_{lt}$ on attribute-neutral prompts (e.g., ``A photo of a doctor'') and attribute-specific prompts (e.g., ``A photo of a male doctor''). For SD-1.5, at timestep $t=0$, intermediate CA layers (typically $l \in [4,8]$) show substantially larger $dp_{lt}$ values for attribute-specific prompts (Fig.~\ref{fig:dot_prods_sd15}). The same pattern holds for SD-2.1, SDXL, and SANA (see the supplementary). This shows that early-step CA activations encode whether the prompt explicitly specifies an attribute.

\medskip


\noindent\textbf{Gating rule.}
Based on this observation, we choose for each backbone a \emph{gating layer} $l^{gate}$ at which attribute presence can be reliably detected. For each attribute $a$, we define a gating threshold as the midpoint between the empirical means of the dot-product statistic on attribute-specific and attribute-neutral prompts:
\begin{equation}
    thr^{a}
    =
    \frac{dp^{a}_{\text{specific}} + dp^{a}_{\text{neutral}}}{2},
    \label{eq:threshold}
\end{equation}
where $dp^{a}_{\text{specific}}$ and $dp^{a}_{\text{neutral}}$ denote the empirical means of $dp^{a}_{l^{gate}0}$ computed over attribute-specific and attribute-neutral prompts, respectively (both measured at layer $l^{gate}$ and timestep $t{=}0$). During generation, we evaluate $dp^{a}_{l^{gate}0}$ once, at $(l^{gate},t{=}0)$. If
\begin{equation}
dp^{a}_{l^{gate}0} > thr^{a}
\quad \text{for any } a \in \{a_i\}_{i=1}^n,
\label{eq:gate}
\end{equation}
we treat the prompt as attribute-specific and perform vanilla generation without intervention; otherwise, we apply the EquiSteer update.

The gate thus prevents unintended modification of prompts that already specify an attribute, while enabling debiasing for neutral ones. \textbf{Note} that Eq.~\ref{eq:gate} is evaluated only once per inference run (at $(l^{gate}, t{=}0)$), and therefore introduces negligible computational overhead.

\subsection{Choosing Steering Strength for Debiasing}
\label{sec:steering_strength}

We now refine the steering update for when the gate decides to apply debiasing, addressing the two main limitations of the basic mechanism: mixed attribute signals and insufficiently pronounced target attributes.

\medskip
\noindent\textbf{Orthogonalisation with respect to attribute directions.}
Let $\{a_i\}_{i=1}^n$ be the attribute set for the debiasing concept, and let $s_{lt}^{a_i}\in\mathbb{R}^{d}$ denote the steering vector for attribute $a_i$ at layer $l$ and timestep $t$. Before injecting the target attribute, we remove from the current cross-attention activation the components lying in the subspace spanned by attribute directions $\mathrm{span}\!\left(\{s_{lt}^{a_i}\}_{i=1}^{n}\right)$.

We construct an orthonormal basis $\{u^{j}_{lt}\}_{j=1}^{m}$ for this subspace via Gram--Schmidt with a small $\varepsilon>0$ for numerical stability. Then we stack the basis vectors as columns of $U_{lt}\in\mathbb{R}^{d\times m}$. Orthogonalization then becomes a projection removal:
\begin{equation}
\label{eq:orth_matmul}
ca^{out\_tmp}_{lt} \;=\; \bigl(I-U_{lt}U_{lt}^{\top}\bigr)\,ca^{out}_{lt}.
\end{equation}
By construction, $ca^{out\_tmp}_{lt}\perp \mathrm{span}\!\left(\{s_{lt}^{a_i}\}_{i=1}^{n}\right)$, which removes any pre-existing attribute signal and makes mixed-attribute generations less likely. \textbf{Note} that $U_{lt}$ (and thus $I-U_{lt}U_{lt}^{\top}$) can be precomputed offline; at inference time this step requires only one matrix multiplication per CA layer.

\medskip


\noindent\textbf{Adaptive steering magnitude.}
After orthogonalisation, we choose the steering strength so that the injected attribute signal matches the typical signal induced by \emph{attribute-specific} prompts at the same layer. 
Recall from Sec.~\ref{sec:dot_product_indicator} (Eq.~\ref{eq:3}) that for an attribute $a$ we measure its expression in CA outputs via the maximal dot-product statistic
$dp^{a}_{lt}=\max_{k}\langle ca^{out}_{ltk}, s^{a}_{lt}\rangle$.
For each layer $l$ (and timestep $t$), we precompute
\[
dp^{a}_{\text{mean}}(l,t)
\;=\;
\mathbb{E}_{\text{attr-spec prompts}}\!\left[\,dp^{a}_{lt}\,\right],
\]
i.e., the mean maximal dot product observed on attribute-specific prompts.
We then set the steering magnitude \(\alpha = dp^{a}_{\text{mean}}(l,t)\) and update
\begin{equation}
    ca^{out\_new}_{lt}
    =
    ca^{out\_tmp}_{lt} + \alpha\, s_{lt}^{a}.
    \label{eq:thr_add}
\end{equation}
Since \(ca^{out\_tmp}_{lt}\perp \mathrm{span}(\{s^{a_i}_{lt}\}_{i=1}^{n})\), this update injects only the selected attribute direction \(s^{a}_{lt}\).
Choosing \(\alpha\) in this way ensures that the target attribute is expressed at a strength consistent with attribute-specific prompts, reducing both weak and ambiguous attribute expression.

\medskip
\noindent\textbf{Summary of EquiSteer.}
At inference time, EquiSteer first checks whether the prompt already specifies an attribute via the gating signal and skips debiasing if so. For neutral prompts, it orthogonalises the CA output against all attribute directions and reinjects the target attribute at a calibrated strength. This allows the method to debias neutral prompts while leaving attribute-specific ones intact, eliminating both mixed-attribute and weak-attribute failure modes of basic steering.

\section{Experiments}
\label{sec:experiments_main}

\subsection{Experimental Setup}

We follow the evaluation protocol of TEI~\cite{kim2025rethinking}. We adopt this setup as it explicitly evaluates debiasing performance on both \emph{attribute-neutral} prompts and \emph{attribute-specific} prompts. This is particularly important in our setting: EquiSteer is designed not only to improve fairness on attribute-neutral prompts, but also to preserve the model’s behavior when the attribute is explicitly specified.

In our main experiments, we evaluate EquiSteer on the \textit{gender} concept with the binary attribute set (\textit{male}, \textit{female}). In Sec.~\ref{sec:exps_additional_concepts} we additionally evaluate EquiSteer on 4 more concepts with varying numbers of attributes: \textit{race} (5 attributes: \textit{White}, \textit{Black}, \textit{Asian}, \textit{Indian}, \textit{Latino}), \textit{age} (3 attributes: \textit{young}, \textit{middle-aged}, \textit{elderly}), \textit{body type} (3 attributes: \textit{slim}, \textit{average build}, \textit{heavy}), \textit{eyeglasses} (2 attributes: \textit{with eyeglasses} and \textit{without eyeglasses}). Because diffusion models handle prompt negation poorly (prompts like \textit{``a man with no eyeglasses''} can still produce eyeglasses), we do not compute a separate steering vector for the negated value for the \textit{eyeglasses} concept; instead, we use a single steering vector for the \textit{eyeglasses} direction and implement the \textit{no eyeglasses} target through a CASteer-style erasure step at inference time. More details are given in Sec.~\ref{sec:eyeglasses_supp}. We consider four widely used backbones: SD-1.5, SD-2.1, SDXL, and SANA.

The prompts used to compute steering vectors and to estimate the gating thresholds and adaptive steering magnitudes do not overlap with testing prompts and are described in the supplementary Sec.~\ref{sec:steering_vectors_prompts}. For the gating mechanism, we use layer $l^{gate} = 4$ for SD-1.5 and SD-2.1, $l^{gate} = 17$ for SDXL, and $l^{gate} = 5$ for SANA. For each model, the gating layer $l^{gate}$ is chosen as the first layer that shows clear separation in maximal token response $dp_{l0}$ on a validation set of prompts (see Sec.~\ref{sec:dot_product_indicator} and Sec.~\ref{sec:gate_analysis_supp} in the supplementary).

\vspace{0.5em}
\noindent\textbf{Evaluation setup.}
We adopt the standard auditing protocol from prior work~\cite{friedrich2023fair,parihar2024balancing,kim2025rethinking}, using both \emph{attribute-neutral} and \emph{attribute-specific} prompts over a fixed set of professions: ``CEO'', ``doctor'', ``pilot'', ``technician'', ``fashion designer'', ``librarian'', ``teacher'', and ``nurse''. These professions are known to induce strong attribute biases in text-to-image diffusion models~\cite{kim2025rethinking}.

For attribute-neutral prompts, we use the template \textit{``A photo of a \{profession\}''}. For attribute-specific prompts, we use the template \textit{``A photo of a \{attribute\} \{profession\}''}. For each model and profession, we generate 1{,}000 images for attribute-neutral prompts and 300 images for attribute-specific prompts.

Following~\cite{kim2025rethinking}, we evaluate attribute bias for all concepts except \textit{eyeglasses} using a CLIP ViT-L/14 zero-shot classifier. For \textit{gender}, we use the prompt template \textit{``A photo of a \{attribute\}''}. For \textit{race}, \textit{age}, and \textit{body type}, we use the profession-conditioned template \textit{``A photo of a \{attribute\} \{profession\}''}, which provides a more specific comparison for occupational prompts.

For the \textit{eyeglasses} concept, we find that this CLIP setup tends to favor the more generic prompt \textit{``A photo of a person''}, which systematically underestimates the presence of eyeglasses. To obtain a more reliable measurement, we instead report eyeglasses results using the VQA-based classifier BLIP-VQA~\cite{li2022blip} \texttt{capfilt-large}. In the supplementary Sec.~\ref{sec:classifier_calibration_supp}, we report agreement rates between BLIP-, CLIP-, GPT-4o-, and human-based evaluations, supporting our choice of the BLIP-based classifier as a more reliable measure of \textit{eyeglasses} presence.
We report both per-profession results and macro-averaged results across all evaluation prompts (see Tables~\ref{tab:debias_perf_sd15} and~\ref{tab:gender_higher_models_main_v2}).

For attribute-specific prompts, the target ratio for the specified attribute is \(1.0\). For attribute-neutral prompts, the target ratio is \(1/n\), where \(n\) is the number of debiasing attributes. We summarise performance using $\Delta = \left|p - \frac{1}{n}\right|$, where \(p\) is the observed attribute ratio; lower values indicate better parity. Thus, we define fairness as closeness of the generated attribute distribution to the target uniform distribution.

In addition to fairness, we evaluate general image--text alignment using CLIPScore~\cite{hessel2021clipscore} with CLIP-ViT-L/14, and image fidelity using CMMD~\cite{jayasumana2024rethinking}. Both metrics are computed on 30{,}000 images generated from prompts in the MS-COCO-2014 validation set~\cite{lin2014microsoft} (see Table~\ref{tab:fid}) on EquiSteer applied to the \textit{gender} concept.

\setlength{\tabcolsep}{3.5pt}
\begin{table*}[t]
  \centering
  \adjustbox{max width=0.95\textwidth}{
  \begin{tabular}{cc|cccccccc}
    \toprule
        Minor attribute & Profession 
    & Vanilla SD
    & FairDiff~\cite{friedrich2023fair}
    & UCE~\cite{gandikota2024unified}$^\dagger$ 
    & FTDiff~\cite{shen2023finetuning}
    & SelfDisc~\cite{li2024self}$^\dagger$
    & TEI & EquiSteer\\
    \midrule
    
    \multirow{4}{*}{\rotatebox[origin=c]{0}{Female}} & CEO & 0.030 & \underline{0.452} & 0.027 &0.190 &0.445 & 0.389  &\textbf{0.483}\\
    & Doctor & 0.081 & \underline{0.502} &0.049 &0.198 &\underline{0.502} & 0.334 & \textbf{0.500} \\
    & Pilot & 0.150 & 0.739 &0.244 &0.260 &\textbf{0.568}	 &  0.408 & \underline{0.416} \\
    & Technician & 0.007  & \textbf{0.553} & 0.005 &0.168 & 0.347 & 0.164 & \underline{0.375}  \\
   \midrule
   \multirow{4}{*}{\rotatebox[origin=c]{0}{Male}} & Fashion designer & 0.078 & 0.333 & 0.018 &0.167 &0.067 &\underline{0.451} & \textbf{0.504}\\
    & Librarian & 0.194 & 0.300 &0.297 &\underline{0.538} &0.174	 &  0.421 & \textbf{0.473}\\
    & Teacher & 0.222 & 0.205& 0.155 &0.231 &0.081 & \textbf{0.492} & \underline{0.421}\\
    & Nurse &  0.007 & 0.162 &0.003 &\underline{0.208} &0.004 & {0.039} &  \textbf{0.432}\\
    \midrule
    &Avg. $\Delta$ ($\downarrow$) &0.403&\underline{0.167}&0.400&0.264&0.244&\underline{0.167}  & \textbf{0.051}\\
    \bottomrule
  \end{tabular}}
  \caption{Ratio of minor attributes within 1,000 images generated with SD1.5, where a value closer to 0.5 is preferred.
  Avg. $\Delta$ represents the average absolute difference from the target ratio of 0.5 across all professions.}
  \label{tab:debias_perf_sd15}
\end{table*}

\noindent\textbf{Baselines.}
We compare EquiSteer against recent debiasing methods, including FairDiffusion~\cite{friedrich2023fair}, Unified Concept Editing (UCE)~\cite{gandikota2024unified}, Fine-Tuning Diffusion (FTDiff)~\cite{shen2023finetuning}, Self-Discovering Latent Direction (SelfDisc)~\cite{li2024self}, FairImagen~\cite{fu2025fairimagen} (for SDXL only), and the method of~\cite{kim2025rethinking}, which we denote as TEI. TEI is our primary point of comparison, as it is also training-free and is explicitly designed to improve fairness on attribute-neutral prompts while preserving behavior on attribute-specific prompts.

\subsection{Experimental Results}

\subsubsection{Debiasing Gender Concept}

Following TEI~\cite{kim2025rethinking}, we report our main results for debiasing the \textit{gender} concept. 
In this case, the set of attributes is binary: (\textit{male}, \textit{female}).


\begin{figure}[!htbp]
      \centering
      \begin{minipage}[t]{0.55\textwidth}
          \centering
          \vspace{0pt}
          \resizebox{0.8\textwidth}{!}{
          \begin{tabular}{lrr}\toprule
          & SDXL & SANA \\
          \midrule
          Vanilla model & 0.381 & 0.473 \\
          TEI & \underline{0.242} & - \\
          EquiSteer & \textbf{0.075} & \textbf{0.097} \\
          \bottomrule
          \end{tabular}
          }
          \captionof{table}{Average $\Delta$ across SDXL and SANA on attribute-neutral prompts (lower is better).}
          \label{tab:gender_higher_models_main_v2}

          \vspace{0.8em}

          \includegraphics[width=0.85\textwidth]{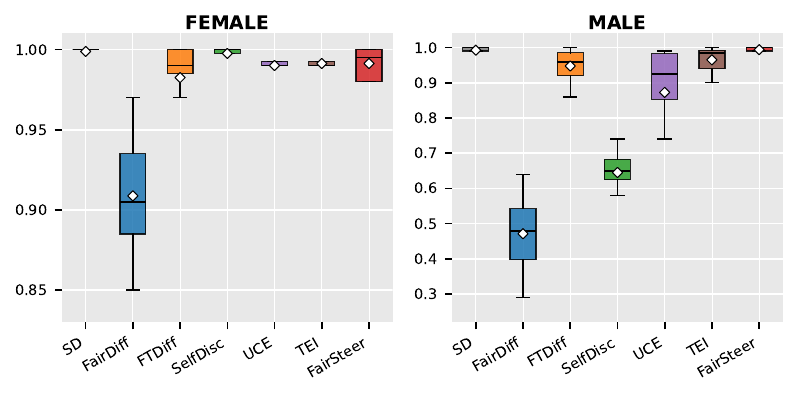}
          \captionof{figure}{Attribute-preservation results on SD-1.5 for attribute-specific prompts. The target value is $1.0$, meaning that the gender explicitly stated in the prompt should be preserved in the generated image. Aggregated over 8 professions.}
          \label{fig:gender_box_v2}
      \end{minipage}
      \hfill
      \begin{minipage}[t]{0.42\textwidth}
          \centering
          \vspace{0pt}
          \scriptsize
          \setlength{\tabcolsep}{4pt}
          \renewcommand{\arraystretch}{1.15}
          \begin{tabular}{cl|cc}
            \toprule
            Model & Method & CLIP\,$\uparrow$ & CMMD\,$\downarrow$ \\
            \midrule
            \multirow{7}{*}{\rotatebox[origin=c]{0}{SD 1.5}}
              & Vanilla SD & 26.42 & 0.532 \\
              & FairDiff & 26.03 & 0.586 \\
              & UCE & 18.10 & 1.240 \\
              & FTDiff & 25.61 & 0.783 \\
              & SelfDisc & 24.97 & 0.900 \\
              & TEI & 26.56 & 0.509 \\
              & EquiSteer & 26.63 & 0.519 \\
            \midrule
            \multirow{4}{*}{\rotatebox[origin=c]{0}{SD 2}}
              & Vanilla SD & 26.55 & 0.524 \\
              & FairDiff & 25.91 & 0.602 \\
              & TEI & 26.29 & 0.549 \\
              & EquiSteer & 26.36 & 0.555 \\
            \midrule
            \multirow{3}{*}{\rotatebox[origin=c]{0}{SDXL}}
              & Vanilla SD & 26.51 & 0.794 \\
              & TEI & 26.58 & 0.757 \\
              & EquiSteer & 26.75 & 0.796 \\
            \midrule
            \multirow{2}{*}{\rotatebox[origin=c]{0}{SANA}}
              & Vanilla SANA & 26.93 & 0.890 \\
              & EquiSteer & 26.77 & 0.905 \\
            \bottomrule
          \end{tabular}
          \captionof{table}{Comparison of text--image alignment (CLIP Score) and image fidelity (CMMD). Our method retains or improves generation quality while enhancing fairness.}
          \label{tab:fid}
      \end{minipage}
  \end{figure}

\noindent\textbf{Attribute-neutral prompts.}
Table~\ref{tab:debias_perf_sd15} reports results for attribute-neutral prompts on SD-1.5. EquiSteer substantially improves over the vanilla model and outperforms the training-free baseline TEI, achieving a markedly lower $\Delta$ score. The improvement is consistent across professions, indicating that EquiSteer effectively reduces gender imbalance in biased occupational prompts.

Table~\ref{tab:gender_higher_models_main_v2} reports the average $\Delta$ across the eight professions for SD-2.1, SDXL, and SANA. EquiSteer again consistently improves over the vanilla model across all backbones, showing that the method transfers well beyond SD-1.5. Detailed per-profession results are provided in the supplementary Sec.~\ref{sec:gender_details_supp}.

For the SDXL model, we additionally compare against FairImagen~\cite{fu2025fairimagen}, which is evaluated only for SDXL. For this comparison, we report evaluation details and results on their setup in the supplementary Sec.~\ref{sec:additional_comparisons_supp}.

\noindent\textbf{Attribute-specific prompts.}
Figure~\ref{fig:gender_box_v2} shows results on attribute-specific prompts for the SD-1.5 model, where the objective is to preserve the attribute explicitly specified in the prompt (target \(=1.0\)). Detailed per-profession breakdowns are provided in the supplementary Sec.~\ref{sec:gender_details_supp}.

For female-specified prompts, most methods already perform strongly, and EquiSteer preserves this behavior, achieving near-perfect results across all eight professions. Male-specified prompts are more challenging, but EquiSteer remains competitive with TEI and outperforms the remaining baselines. Results for SD-2.1, SDXL, and SANA, reported in the supplementary Sec.~\ref{sec:gender_details_supp}, show the same overall pattern: EquiSteer preserves attribute-specific generations while improving fairness on attribute-neutral prompts.

\noindent\textbf{Highly biased concepts.}
Among the evaluated professions, \textit{nurse} and \textit{technician} are particularly challenging, as they are strongly biased toward the \textit{female} and \textit{male} attributes, respectively. These cases are difficult because stronger pre-existing bias makes it harder to move the generated attribute distribution toward parity. Even in these settings, EquiSteer yields clear improvements over the vanilla model, and the detailed results in the supplementary Sec.~\ref{sec:gender_details_supp} show that it remains effective on strongly biased prompts.

\medskip
\noindent\textbf{Overall,} across all four backbones, EquiSteer consistently improves debiasing performance on the gender concept while preserving fidelity to attribute-specific prompts. As shown in Table~\ref{tab:fid}, it also maintains strong image fidelity and text--image alignment, remaining competitive with recent debiasing approaches on both quality metrics.

\vspace{-1pt}
\subsubsection{Debiasing Additional Concepts}
\label{sec:exps_additional_concepts}

\begin{table}[t]
\centering
\fontsize{6.5pt}{7.5pt}\selectfont
\begin{minipage}[t]{0.47\textwidth}
\centering
\setlength{\tabcolsep}{3pt}
\begin{tabular}{ll|cc|cc}
\toprule
& & \multicolumn{2}{c|}{\textbf{SDXL}} & \multicolumn{2}{c}{\textbf{SANA}} \\
\cmidrule(lr){3-4}\cmidrule(lr){5-6}
Attribute & Clf. & Van. & ES & Van. & ES \\
\midrule
\textit{Race}       & CLIP & $0.172$ & $\mathbf{0.042}$ & $0.178$ & $\mathbf{0.039}$ \\
\textit{Age}        & CLIP & $0.276$ & $\mathbf{0.115}$ & $0.387$ & $\mathbf{0.060}$ \\
\textit{Body type}  & CLIP & $0.317$ & $\mathbf{0.220}$ & $0.390$ & $\mathbf{0.231}$ \\
\textit{Eyeglasses} & BLIP & $0.270$ & $\mathbf{0.085}$ & $0.234$ & $\mathbf{0.175}$ \\
\bottomrule
\end{tabular}
\caption{EquiSteer (ES) applied on four additional attributes beyond gender. Parity gap $\Delta$ on attribute-neutral prompts (mean over 8 professions, lower is better).}
\label{tab:main_attributes_summary}
\end{minipage}
\hspace{0.04\textwidth}
\begin{minipage}[t]{0.47\textwidth}
\centering
\setlength{\tabcolsep}{4pt}
\begin{tabular}{l|cc|cc}
\toprule
& \multicolumn{2}{c|}{\textbf{SDXL}} & \multicolumn{2}{c}{\textbf{SANA}} \\
\cmidrule(lr){2-3}\cmidrule(lr){4-5}
Attribute & Van. & ES & Van. & ES \\
\midrule
\textit{Gender}    & $0.381$ & $\mathbf{0.012}$ & $0.473$ & $\mathbf{0.009}$ \\
\textit{Race}      & $0.192$ & $\mathbf{0.047}$ & $0.237$ & $\mathbf{0.111}$ \\
\textit{Age}       & $0.276$ & $\mathbf{0.158}$ & $0.387$ & $\mathbf{0.032}$ \\
\textit{Body type} & $0.317$ & $\mathbf{0.162}$ & $0.391$ & $\mathbf{0.178}$ \\
\bottomrule
\end{tabular}
\caption{Joint-4 debiasing of gender + race + age + body type simultaneously. Parity gap $\Delta$ per attribute; lower is better.}
\label{tab:main_multiconcept}
\end{minipage}
\end{table}

\begin{figure}[t]
    \centering
    \begin{subfigure}[t]{0.49\textwidth}
        \centering
        \includegraphics[width=\textwidth]{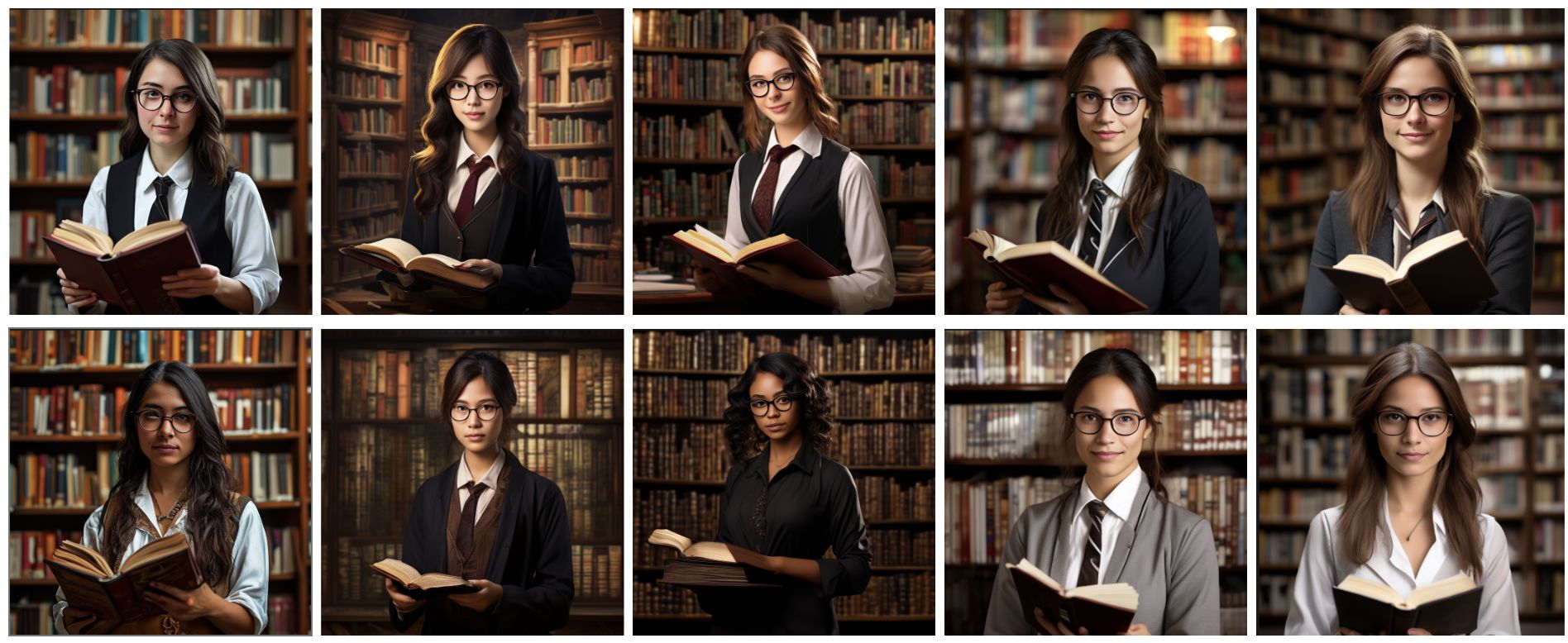}
        \caption{Prompt: ``A photo of a librarian'', SANA}
        \label{fig:qual_race_sana_main}
    \end{subfigure}
    \hfill
    \begin{subfigure}[t]{0.49\textwidth}
        \centering
        \includegraphics[width=\textwidth]{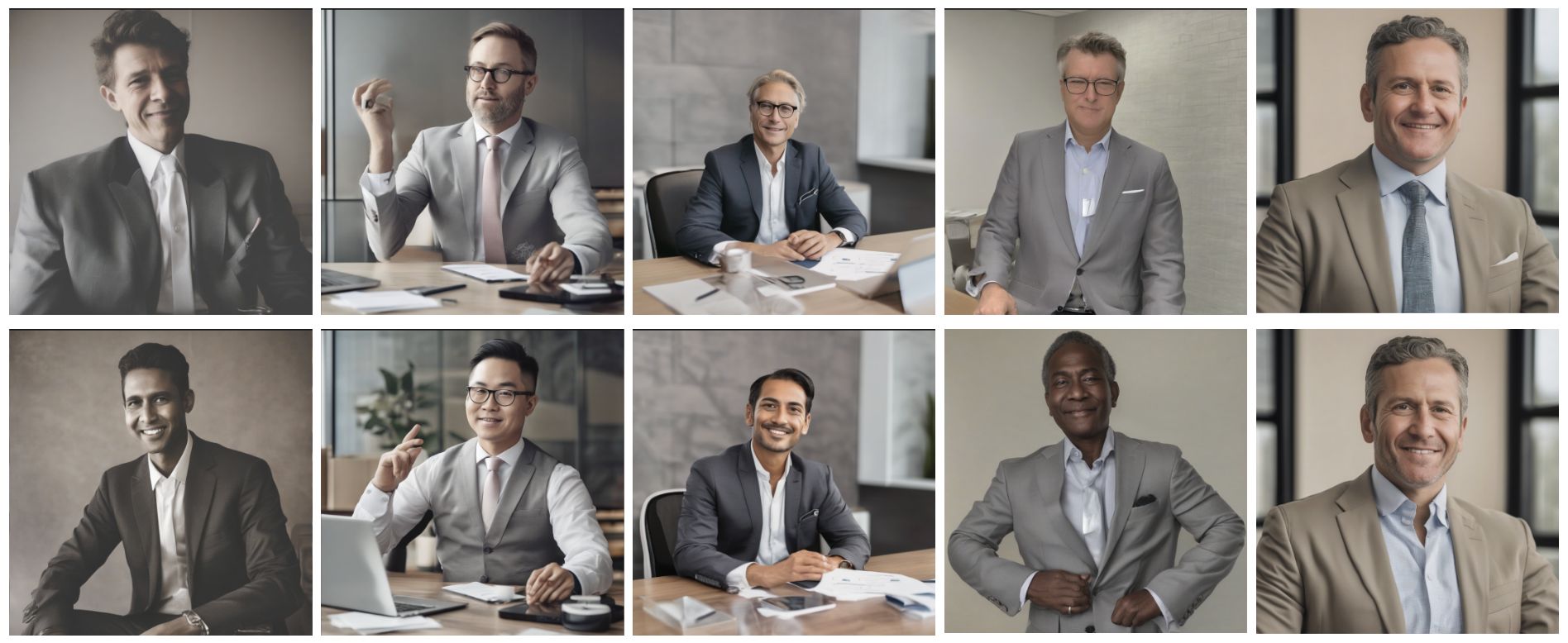}
        \caption{Prompt: ``A photo of a CEO'', SDXL}
        \label{fig:qual_race_sdxl_main}
    \end{subfigure}
    \caption{Qualitative examples of EquiSteer on the \textit{race} concept. Top row in each panel: vanilla model; bottom row: EquiSteer.}
    \label{fig:qualitative_race}
\end{figure}

Tab.~\ref{tab:main_attributes_summary} reports the parity gap before and after EquiSteer applied on SDXL and SANA models for each of the four additional debiasing concepts. EquiSteer reduces the parity gap on every cell, with the strongest reductions on \textit{race} ($76\!-\!78\%$) and on \textit{age} on SANA ($85\%$). Fig.~\ref{fig:qualitative_race} shows qualitative examples of \textit{race} debiasing on SANA and SDXL: the vanilla generations skew toward a single demographic per profession, while EquiSteer distributes generations across the five race classes while preserving scene composition. We report per-profession breakdowns and more qualitative results in Sec.~\ref{sec:additional_attributes} in the supplementary.

\vspace{-1pt}
\subsubsection{Multi-concept debiasing}
\label{sec:exps_multi_concept}
We further apply EquiSteer to several concepts within the same generation, by running the full intervention (gate, orthogonalisation, target sampling, adaptive re-injection) sequentially per concept at every cross-attention layer and step. Each attribute retains its own threshold and steering vectors. Tab.~\ref{tab:main_multiconcept} reports \emph{joint} debiasing of 4 concepts: \textit{gender + race + age + body type} simultaneously. All four attributes are substantially debiased in one pass on both backbones. Per-profession breakdowns, qualitative results and additional multi-concept debiasing results are reported in Sec.~\ref{sec:multi_concept_supp} in the supplementary.

\subsection{Ablations}
\label{sec:ablations}

\noindent\textbf{EquiSteer components} We ablate the three components of EquiSteer namely, adaptive-magnitude steering (\textit{add}; Eq.~\ref{eq:fairsteer_base},~\ref{eq:thr_add}), subspace orthogonalisation (\textit{erase}; Eq.~\ref{eq:orth_matmul}), and the prompt-aware gate (\textit{gate}; Eq.~\ref{eq:gate}). We use EquiSteer applied to SD-1.5 on gender concept. Results presented in Tab.~\ref{tab:main_ablation} show that each component matters: \textit{add} alone gives only modest debiasing ($\Delta_{\text{neutral}}{=}0.165$); adding orthogonalisation tightens parity ($\Delta_{\text{neutral}}{=}0.095$) but corrupts attribute preservation on attribute-specific prompts (male-specific recall drops to $0.230$). Finally, the gate restores attribute preservation ($\Delta_{\text{male}}{=}0.495$, $\Delta_{\text{female}}{=}0.492$) while keeping the neutral parity gap at its tightest ($\Delta_{\text{neutral}}{=}0.051$). Per-profession ablation details are in the supplementary Sec.~\ref{sec:ablation_suppl}.

\begin{table}[t]
\centering
\setlength{\tabcolsep}{6pt}
\begin{tabular}{l|ccc}
\toprule
SD-1.5 / Gender & Neutral $\Delta$ ($\downarrow$) & Female-spec.\ $\Delta$ ($\uparrow$) & Male-spec.\ $\Delta$ ($\uparrow$) \\
\midrule
\textit{add}                          & $0.165$ & $0.481$ & $0.379$ \\
\textit{add \& erase}                 & $0.095$ & $0.394$ & $0.230$ \\
\textit{add \& erase \& gate}  & $\mathbf{0.051}$ & $\mathbf{0.492}$ & $\mathbf{0.495}$ \\
\bottomrule
\end{tabular}
\caption{Ablation of EquiSteer components on SD-1.5 gender (avg.\ over eight professions; $\Delta = |R - 0.5|$). Each row adds one component to the previous. \textit{add} = basic adaptive-magnitude steering (Eq.~\ref{eq:fairsteer_base},~\ref{eq:thr_add}); \textit{erase} = subspace orthogonalisation (Eq.~\ref{eq:orth_matmul}); \textit{gate} = the prompt-aware gate (Eq.~\ref{eq:gate}).}
\label{tab:main_ablation}
\end{table}

\noindent\textbf{Transferability across prompt families.}
The standard evaluation uses the template \textit{``A photo of a \{profession\}''}. We test whether EquiSteer transfers to 13 held-out templates grouped into three prompt families: paraphrased, long contextual, and compositional/multi-subject prompts. We use the same steering vectors, gate layers, and thresholds for all templates, without prompt-family-specific recalibration. The full prompt list and per-template results are provided in the supplementary Sec.~\ref{sec:transferability_supp}.

Table~\ref{tab:main_transferability} reports the aggregate gender parity gap on SDXL and SANA. EquiSteer reduces $\Delta$ across all three prompt families on both backbones, with reductions ranging from $48\%$ to $89\%$. Paraphrased and long-context prompts behave similarly to the standard template, while the weakest case is SANA under compositional prompts ($48\%$ reduction), likely because the current image-level gate cannot separately control two subjects in the same scene. We discuss this limitation further in the supplementary Sec.~\ref{sec:transferability_supp}.


\begin{table}[t]
\centering
\setlength{\tabcolsep}{6pt}
\begin{tabular}{l|cc|cc}
\toprule
& \multicolumn{2}{c|}{\textbf{SDXL}} & \multicolumn{2}{c}{\textbf{SANA}} \\
\cmidrule(lr){2-3}\cmidrule(lr){4-5}
Prompt Family & Van. & ES & Van. & ES \\
\midrule
Standard (\textit{``A photo of a \{prof\}''})  & $0.381$ & $\mathbf{0.075}$ & $0.473$ & $\mathbf{0.097}$ \\
Paraphrased                                    & $0.376$ & $\mathbf{0.100}$ & $0.457$ & $\mathbf{0.100}$ \\
Long contextual                                & $0.333$ & $\mathbf{0.120}$ & $0.444$ & $\mathbf{0.049}$ \\
Compositional / multi-subject                  & $0.262$ & $\mathbf{0.045}$ & $0.295$ & $\mathbf{0.155}$ \\
\bottomrule
\end{tabular}
\caption{EquiSteer (ES) transferability across prompt families. \textit{Gender}-debiasing parity gap $\Delta$ averaged over (template $\times$ profession) cells; lower is better. The same steering vectors and gate thresholds (calibrated once on generic prompts) are reused without re-tuning across every regime. Per-template breakdowns are in Sec.~\ref{sec:transferability_supp} of the supplementary.}
\label{tab:main_transferability}
\end{table}


\noindent\textbf{Gate analysis} We perform a detailed analysis of EquiSteer's gating mechanism in the supplementary Sec.~\ref{sec:gate_analysis_supp}. We measure the realised inference-time gate's AUROC across every (backbone $\times$ attribute) cell (Sec.~\ref{sec:gate_analysis_auroc}), describe an automated procedure for selecting the gating layer $l^{gate}$ that recovers our manually chosen layers (Sec.~\ref{sec:gate_analysis_lgate}), test the gate's separability on subtle / low-saliency attributes such as religion, disability, and socio-economic status (Sec.~\ref{sec:subtle_attributes_supp}), and sweep the threshold multiplier $m$ to study sensitivity of debiasing performance to the gate threshold (Sec.~\ref{sec:thrshift_supp}). The gate is highly separable, with AUROC $\geq 0.988$ for every evaluated cell. The default threshold multiplier $m{=}1$ provides a favorable trade-off between attribute-neutral debiasing and attribute-specific preservation.

\noindent\textbf{Classifier calibration and human evaluation} We assess the reliability of the default CLIP zero-shot evaluator in the supplementary Sec.~\ref{sec:classifier_calibration_supp}. We collect labels from two human annotators on a 600-image stratified subset of generations, and use their agreement with GPT-4o labels to validate GPT-4o as a scalable visual oracle. We then recompute the parity gaps using this oracle as an additional evaluator. The resulting $\Delta$ values show that the CLIP-based evaluation is conservative: for every evaluated attribute, the same EquiSteer interventions yield equal or larger reductions under GPT-4o. This effect is especially pronounced for race and eyeglasses, where CLIP exhibits per-class recall failures.

\section{Conclusion}
\label{sec:conclusion}
We presented \emph{EquiSteer}, a training-free inference-time framework for debiasing text-to-image diffusion models via cross-attention modulation. EquiSteer uses precomputed steering vectors with adaptive magnitude to shift generations toward desired attributes, and a prompt-aware \textit{gating mechanism} to skip intervention on attribute-specific prompts. Across multiple concepts and backbones, EquiSteer consistently improves fairness while maintaining text--image alignment and visual quality; it requires no retraining, operates per sample, and integrates naturally with standard sampling pipelines.

\section*{Acknowledgments}
This work was supported by a Google DeepMind PhD Studentship, and the work utilized Queen Mary's Andrena HPC facility, supported by QMUL Research-IT. This work was also supported by the Engineering and Physical Sciences Research Council [grant number EP/Y009800/1], through funding from Responsible AI UK (KP0016).

\bibliographystyle{splncs04}
\bibliography{main}

\clearpage

\pagenumbering{arabic}
\renewcommand{\thepage}{P\arabic{page}} 

\setcounter{section}{0}
\setcounter{subsection}{0}
\renewcommand{\thesection}{S\arabic{section}} 
\renewcommand{\thesubsection}{\thesection.\arabic{subsection}}

\addtocontents{toc}{\protect\setcounter{tocdepth}{2}}

\renewcommand{\contentsname}{Supplementary Material -- Contents}

\begingroup
\makeatletter
\let\l@title\@gobbletwo
\let\l@author\@gobbletwo
\makeatother

\tableofcontents
\endgroup

\section{Limitations}

While EquiSteer is a general-purpose inference-time framework that can be applied across different backbones and concepts, it also introduces several practical challenges and directions for future work. 
First, EquiSteer adds inference-time overhead due to the per-layer interventions in Eqs.~\ref{eq:fairsteer_base},~\ref{eq:thr_add}, and the re-normalization in Eq.~\ref{eq:renorm_base}. Although this overhead is modest, it may still matter in latency-critical generation systems. 
Second, EquiSteer relies on concept-specific calibration, including the choice of steering magnitudes (via $dp_{\text{mean}}^{a}$) and gating thresholds $thr^{a}$. In practice, the gate can be overly conservative, suppressing debiasing for some attribute-neutral prompts (false positives), and it can also be too permissive, allowing some attribute-specific prompts to pass through and be modified (false negatives). 
Improving the reliability of gating (e.g. via better attribute detectors, uncertainty-aware thresholds, or adaptive prompt-dependent calibration) is an important avenue for future work, especially for deployment settings with strict constraints on attribute preservation.

\clearpage
\section{Future Work}
\label{sec:future_work}

While EquiSteer provides a general-purpose inference-time framework for debiasing along discrete attributes, our experiments focus on relatively simple prompt templates. Extending the method to more complex and compositional prompts will likely require adapting the calibration procedure (e.g., thresholds and steering magnitudes) to account for richer prompt structure. Such extensions could be integrated into real-world generative pipelines to improve fairness in practical deployments.
\paragraph{Entity-aware gating for complex prompts.}
A promising direction is to apply EquiSteer’s gating mechanism at the level of \emph{individual prompt entities}. For example, for prompts that mention multiple people, the system could decide separately for each person whether debiasing should be applied. Regions corresponding to different entities can be approximated using cross-attention map analysis as in Prompt-to-Prompt~\cite{hertz2022prompt}, enabling per-entity attribute detection and more robust handling of long or multi-entity prompts.
\paragraph{Targeted (localized) debiasing.}
Another line of work is \emph{targeted} debiasing, where intervention is applied only to specific entities or concepts in the prompt (e.g., debias \textit{doctors} while leaving other professions unchanged). One approach is to localize the relevant image tokens by analyzing token-wise scores such as $\langle ca^{out}_{ltk}, s^{a}_{lt}\rangle$ (cf.\ Fig.~\ref{fig:qual_race_sana_main}) and applying the update only to the subset of tokens associated with the target entity, potentially combined with cross-attention-based region extraction~\cite{hertz2022prompt}. We leave a systematic study of such localized interventions to future work.
\paragraph{Limitations on compositional prompts.}
Finally, the current gate operates at the image level, which can be overly conservative for compositional prompts that mix attribute-specific and attribute-neutral content (e.g., prompts describing multiple people with different levels of attribute specification). In such cases, the presence of any attribute-specific signal may suppress debiasing altogether. Extending the gate to be entity or region-specific, using cross-attention-based localization~\cite{hertz2022prompt}, is a natural direction to improve robustness to compositionality.

\clearpage
\section{Construction of steering vectors}
\label{sec:steering_vectors_prompts}

In this section we describe construction of steering vectors that are used in EquiSteer (Sec.~\ref{sec:fair_steer_base},~\ref{sec:dot_product_indicator}).

We closely follow CASteer~\cite{gaintseva2025casteer} and construct steering vectors using pairs of contrastive prompts for each attribute of each concept. Below we give sets of prompts that we use for each attribute of each concept.

\subsection{\textit{Gender} concept}
\label{sec:prompts_steering_vecs}

As \textit{gender} is binary, with attributes \textit{male} and \textit{female}, we construct steering vectors from prompt pairs of the form
\[
(b_m\, c,\; b_f\, c),
\]
where $b_m \in B_m$, $b_f \in B_f$, and $c \in C$.

The male prompt templates are:
\begin{itemize}
    \item ``a boy''
    \item ``two men''
    \item ``two male people''
    \item ``a man''
    \item ``an old man''
    \item ``boys''
    \item ``men''
    \item ``group of male people''
    \item ``a male human''
\end{itemize}

The female prompt templates are:
\begin{itemize}
    \item ``a girl''
    \item ``two women''
    \item ``two female people''
    \item ``a woman''
    \item ``an old woman''
    \item ``girls''
    \item ``women''
    \item ``group of female people''
    \item ``a female human''
\end{itemize}

The shared contexts are:
\begin{itemize}
    \item ``''
    \item ``gloomy image''
    \item ``zoomed in''
    \item ``talking''
    \item ``on the street''
    \item ``in a strange pose''
    \item ``realism''
    \item ``colorful background''
    \item ``on a beach''
    \item ``playing guitar''
    \item ``enjoying nature''
    \item ``smiling''
    \item ``in a futuristic spaceship''
    \item ``with kittens''
\end{itemize}

Since $|B_m| = |B_f| = 9$ and $|C| = 14$, this yields a total of $9 \times 14 = 126$ prompt pairs for each steering direction, i.e. for \textit{male}$\rightarrow$\textit{female} and \textit{female}$\rightarrow$\textit{male}.

\subsection{\textit{Race} concept}
\label{sec:prompts_race}

For \textit{race}, we construct steering vectors from prompt pairs of the form
\[
(b\ \text{of}\ r\ \text{race}\ c,\; b\, c),
\]
where $b \in B$, $c \in C$, and $r$ denotes the race attribute of interest.

The base prompt templates are:
\begin{itemize}
    \item ``a girl''
    \item ``a boy''
    \item ``two men''
    \item ``two women''
    \item ``two people''
    \item ``a man''
    \item ``a woman''
    \item ``an old man''
    \item ``an old woman''
    \item ``boys''
    \item ``girls''
    \item ``men''
    \item ``women''
    \item ``group of people''
    \item ``a human''
\end{itemize}

The shared contexts are:
\begin{itemize}
    \item ``''
    \item ``gloomy image''
    \item ``zoomed in''
    \item ``talking''
    \item ``on the street''
    \item ``in a strange pose''
    \item ``realism''
    \item ``colorful background''
    \item ``on a beach''
    \item ``playing guitar''
    \item ``enjoying nature''
    \item ``smiling''
    \item ``in a futuristic spaceship''
    \item ``with kittens''
\end{itemize}

Since $|B| = 15$ and $|C| = 14$, this yields a total of $15 \times 14 = 210$ prompt pairs for each race attribute.

\subsection{\textit{Age} concept}
\label{sec:prompts_age}

For \textit{age}, we construct steering vectors from prompt pairs of the form
\[
(a\,b\,c,\; b\,c),
\]
where $a$ denotes the age attribute of interest, $a \in \{\textit{young}, \textit{middle-aged}, \textit{elderly}\}$, and the base subject templates $b \in B$ and shared contexts $c \in C$ are the same sets defined in Sec.~\ref{sec:prompts_race}. 

Since $|B| = 15$ and $|C| = 14$, this yields $15 \times 14 = 210$ prompt pairs for each age attribute.

\subsection{\textit{Body type} concept}
\label{sec:prompts_body}

For \textit{body type}, we construct steering vectors from prompt pairs of the form
\[
(a\,b\,c,\; b\,c),
\]
where $a \in \{\textit{slim}, \textit{average build}, \textit{heavy}\}$, and the base subject templates $b \in B$ and shared contexts $c \in C$ are the same sets defined in Sec.~\ref{sec:prompts_race}. 

Since $|B| = 15$ and $|C| = 14$, this yields $15 \times 14 = 210$ prompt pairs for each body-type attribute.

\subsection{\textit{Eyeglasses} concept}
\label{sec:prompts_eyeglasses}

\textit{Eyeglasses} is a binary attribute with values $\{\textit{eyeglasses}, \textit{no eyeglasses}\}$. As diffusion models handle prompt negation poorly (prompts like \textit{``a man with no eyeglasses''} can still produce eyeglasses), we do not compute a separate steering vector for the negated value; instead, we use a single steering vector for the \textit{eyeglasses} direction and implement the \textit{no eyeglasses} target through a CASteer-style erasure step at inference time (Sec.~\ref{sec:eyeglasses_supp}).

The \textit{eyeglasses} steering vector is constructed from prompt pairs of the form
\[
(b\ \text{wearing eyeglasses}\ c,\; b\,c),
\]
where $b \in B$ and $c \in C$ are the same subject-template and context sets defined in Sec.~\ref{sec:prompts_race}.

Since $|B| = 15$ and $|C| = 14$, this yields $15 \times 14 = 210$ prompt pairs for the single \textit{eyeglasses} direction.

\clearpage
\section{Gate threshold choice}
\label{sec:gate_threshold_supp}

The per-attribute gate threshold $thr^{a}$ (Eq.~\ref{eq:threshold}) is the only quantity EquiSteer uses at inference time to decide whether a prompt is already attribute-specific. It is set once per (backbone $\times$ attribute) cell as the midpoint between the empirical means of the maximal token response statistic $dp^{a}_{l^{gate} 0}$ (Eq.~\ref{eq:3}) on a small set of \emph{neutral} calibration prompts and a corresponding set of \emph{attribute-specific} calibration prompts. This section lists the calibration prompts we use per attribute, how the means are estimated, and one practical note on the gender calibration that matters for reproducibility.

\noindent \textbf{Estimation protocol.}
For each attribute $a$, each calibration prompt $p$ is rendered with $10$ random seeds on the target backbone. For each generated image we extract the CA outputs at the chosen gating layer $l^{gate}$ at $t{=}0$, compute the per-direction maximal token response $dp^{a}_{l^{gate} 0}$ (Eq.~\ref{eq:3}), and take the empirical means over (i) all images generated from neutral prompts and (ii) all images generated from attribute-specific prompts. The threshold $thr^{a}$ is then the midpoint of these two means (Eq.~\ref{eq:threshold}).

\noindent \textbf{Per-attribute calibration prompts.}
Tab.~\ref{tab:gate_threshold_prompts} lists the calibration prompts per attribute. For all the concepts except gender, we use concept-neutral subject pair (\textit{``a man''}, \textit{``a woman''}). For the gender concept this subject pair is not neutral, so we use a pair of professions that are not present in test prompts instead: (\textit{cleaner}, \textit{counselor}). The attribute-specific prompts are obtained by prepending the attribute value to the neutral prompt (e.g.\ \textit{``A photo of a male cleaner''}, \textit{``A photo of an Asian man''}, \textit{``A photo of a man wearing eyeglasses''}, \textit{``A photo of an elderly woman''}, \textit{``A photo of a heavy man''}). For multi-class attributes (race, age, body type) one threshold is fitted per attribute value $a$ from the (neutral, $a$-specific) pair; the inference-time gate fires when \emph{any} per-direction threshold is exceeded (Eq.~\ref{eq:gate}).

\begin{table}[th!]
  \centering
  \caption{Calibration templates used to estimate the per-direction gate threshold $thr^{a}$ via Eq.~\ref{eq:threshold}. For each attribute value $a$, we instantiate neutral and attribute-specific prompt pairs and render each prompt with 10 random seeds per backbone. Here $p$ denotes a profession and $s$ denotes a subject word; their instantiated values are listed below the table.}
  \label{tab:gate_threshold_prompts}
  \scriptsize
  \setlength{\tabcolsep}{5pt}
  \begin{tabular}{l|l|l|l}
  \toprule
  Attribute & Neutral template & Attribute-specific template & Values $a$ \\
  \midrule
  Gender      & \textit{``A photo of a $p$''} & \textit{``A photo of a $a$\ $p$''} & \textit{male}, \textit{female} \\
  Race        & \textit{``A photo of a $s$''} & \textit{``A photo of a $a$\ $s$''} & \textit{white}, \textit{black}, \textit{asian}, \textit{indian}, \textit{latino} \\
  Eyeglasses  & \textit{``A photo of a $s$''} & \makecell[l]{\textit{``A photo of a $s$}\\ \textit{ wearing eyeglasses''}} & \textit{eyeglasses} (1 direction) \\
  Age         & \textit{``A photo of a $s$''} & \textit{``A photo of a $a$\ $s$''} & \textit{young}, \textit{middle-aged}, \textit{elderly} \\
  Body type   & \textit{``A photo of a $s$''} & \textit{``A photo of a $a$\ $s$''} & \textit{slim}, \textit{average build}, \textit{heavy} \\
  \bottomrule
  \end{tabular}
  \vspace{2pt}\\
  \scriptsize $p \in \{$\textit{cleaner}, \textit{counselor}$\}$ for gender; $s \in \{$\textit{man}, \textit{woman}$\}$ for the other attributes.
\end{table}

\clearpage
\section{Additional Comparisons}
\label{sec:additional_comparisons_supp}

We additionally compare FairImagen and EquiSteer on the FairImagen evaluation setup. FairImagen prompts follow the template \texttt{Generate a photo of a face of a \{concept\}}, and we use the same prompt format when evaluating EquiSteer. For FairImagen, we use FairPCA with hidden dimension $512$ and empirical noise parameter $\epsilon=0.2$. These hyperparameters were selected on the FairImagen development set using their default protocol, namely SDXL\footnote{The Fairimagen paper is based on SD3 whereas our work compares with the shared Fairimagen \href{https://github.com/fuzihaofzh/FairImagen}{implementation} of SDXL on the same setup} with batch size $12$ and $10$ diffusion steps.

 We report a comparison on the FairImagen setup by aligning the prompt template, guidance scale, profession set, and number of diffusion steps between FairImagen and EquiSteer. Since FairImagen uses $10$ diffusion steps in its main evaluation and hyperparameter search for efficiency, we show both $10$-step and $28$-step results.

We report both pooled $\Delta$, computed from the aggregate demographic proportions over all generated images in a run, and macro-averaged $\Delta$, where we first compute $\Delta$ separately for each profession and then average across professions.

\begin{table}[thpb!]
\centering
\begin{tabular}{llccc}
\hline
Method / Setup & Attribute & Steps & Pooled $\Delta$ & Macro $\Delta$ \\
\hline
FairImagen on FairImagen setup & Gender & 10 & 0.098 & 0.372 \\
FairImagen on FairImagen setup & Gender & 28 & 0.105 & 0.382 \\
EquiSteer on FairImagen setup & Gender & 10 & 0.008 & 0.173 \\
EquiSteer on FairImagen setup & Gender & 28 & 0.003 & 0.180 \\
\hline
\end{tabular}
\caption{Gender comparison on the FairImagen prompt setup. Lower is better.}
\label{tab:fairimagen_delta_steps}
\end{table}

\clearpage
\section{Algorithm}

In this section, we present Algorithm~\ref{alg:fairsteer_full}, which summarizes the EquiSteer inference-time procedure. Note that for a fixed layer $l$ and denoising step $t$, the EquiSteer update is applied to all image tokens in parallel (i.e. via a batched tensor operation).

\begin{algorithm}[H]
\footnotesize
\caption{\textsc{EquiSteer}: inference-time debiasing via cross-attention steering}
\label{alg:fairsteer_full}
\KwIn{
Prompt $p$; attribute set $\{a_i\}_{i=1}^{n}$; denoising steps $t=0,\dots,T-1$; CA layers $l=1,\dots,L$;\\
CA outputs $\{ca^{out}_{ltk}\in\mathbb{R}^{d}\}$ (image token index $k$);\\
Steering vectors $\{s^{a}_{lt}\in\mathbb{R}^{d}\}$ for all $a,l,t$;\\
Gating layer $l^{gate}$; thresholds $\{thr^{a}\}$;\\
Adaptive magnitudes $\{dp^{a}_{\text{mean}}(l,t)\}$ (or $\{dp^{a}_{\text{mean}}(l)\}$ if timestep-independent);\\
Orthogonalization operators $\{P_{lt}=I-U_{lt}U_{lt}^{\top}\in\mathbb{R}^{d\times d}\}$ (precomputed offline);\\
Renormalization flag \texttt{renorm}.
}
\KwOut{Modified CA outputs used during denoising.}

\BlankLine
\textbf{(1) Gate evaluation (once per run).}\;
\For{$a \in \{a_i\}_{i=1}^{n}$}{
    $dp^{a}_{l^{gate}0} \leftarrow \max_{k}\ \langle ca^{out}_{l^{gate}0k},\ s^{a}_{l^{gate}0}\rangle$\;
}
\If(\tcp*[f]{attribute-specific prompt}){$\exists a \in \{a_i\}_{i=1}^{n}:\ dp^{a}_{l^{gate}0} > thr^{a}$}{
    \Return \tcp*[f]{Vanilla generation (no intervention)}\;
}

\BlankLine
\textbf{(2) Sample target attribute for neutral prompt.}\;
Sample $a \sim \mathrm{Uniform}(\{a_i\}_{i=1}^{n})$\;

\BlankLine
\textbf{(3) Apply EquiSteer at each denoising step and CA layer.}\;
\For{$t \leftarrow 0$ \KwTo $T-1$}{
  \For{$l \leftarrow 1$ \KwTo $L$}{
    \ForEach{image token $k$}{
      \tcp{(3a) Orthogonalize w.r.t. attribute subspace}
      $ca^{out\_tmp}_{ltk} \leftarrow P_{lt}\, ca^{out}_{lt}$\;

      \tcp{(3b) Adaptive magnitude (layer-/timestep-specific)}
      $\alpha \leftarrow dp^{a}_{\text{mean}}(l,t)$\;

      \tcp{(3c) Inject target attribute direction}
      $ca^{out\_new}_{ltk} \leftarrow ca^{out\_tmp}_{ltk} + \alpha\, s^{a}_{lt}$\;

      \tcp{(3d) Optional renormalization for stability}
      \If{\textnormal{\texttt{renorm}}}{
        $ca^{out\_new}_{ltk} \leftarrow \|ca^{out}_{ltk}\|_2\,
        \dfrac{ca^{out\_new}_{ltk}}{\|ca^{out\_new}_{ltk}\|_2+\varepsilon}$\;
      }

      \tcp{Write back modified CA output}
      $ca^{out}_{ltk} \leftarrow ca^{out\_new}_{ltk}$\;
    }
  }
}
\end{algorithm}

\clearpage
\section{Debiasing additional concepts}
\label{sec:additional_attributes}

In this section, we report results on applying EquiSteer to three further concepts beyond the \textit{gender}: \textit{age} (3-way: \textit{young}, \textit{middle-aged}, \textit{elderly}), \textit{body type} (3-way: \textit{slim}, \textit{average build}, \textit{heavy}), and \textit{eyeglasses} (binary). Age and body type use the same multi-class pipeline as the race experiments in Sec.~\ref{sec:exps_additional_concepts}; \textit{eyeglasses} requires a small modification for the negated value, described in Sec.~\ref{sec:eyeglasses_supp}.

For all three attributes, we use the standard evaluation setup from Sec.~\ref{sec:experiments_main}.

\subsection{Age}
\label{sec:age_supp}

\noindent \textbf{Method.}
Steering vectors and gate thresholds for the three age values are constructed as described in Sec.~\ref{sec:prompts_age} and Sec.~\ref{sec:gate_threshold_supp}, respectively. We report results on two most capable diffusion backbones: SDXL and SANA-1.5

\noindent \textbf{Evaluation protocol.}
Following our evaluation protocol, we use the CLIP zero-shot classifier with the templates \textit{``A photo of a young person''}, \textit{``A photo of a middle-aged person''}, and \textit{``A photo of an elderly person''}. Each generated image is assigned to the class with the highest CLIP score.

\noindent \textbf{Quantitative results.}
Tab.~\ref{tab:age_per_concept} reports the per-profession class ratios and parity gaps. On SDXL, EquiSteer reduces the mean parity gap from $0.276$ to $0.115$ ($58\%$ reduction); on SANA-1.5, from $0.387$ to $0.060$ ($85\%$ reduction). Note that vanilla SANA-1.5 is essentially mode-collapsed to \textit{middle-aged} on five of eight professions (Tab.~\ref{tab:age_per_concept}): CEO, doctor, pilot, teacher, and nurse all have $\geq 87.5\%$ middle-aged generations, with $0\%$ elderly. After EquiSteer applied, every profession on SANA-1.5 except \textit{teacher} converges to within $0.07$ of uniform across the three classes. The single resistant cell is \textit{teacher} on SANA-1.5 ($\Delta = 0.244$, still pulled toward middle-aged), reflecting an unusually strong profession-attribute prior that our gate-then-redistribute mechanism partially but not fully overcomes.

\begin{table*}[th!]
  \centering
  \caption{Per-profession \textit{age} class ratios on SDXL and SANA-1.5 (CLIP ViT-L/14 zero-shot, $n{=}1{,}000$ images per cell, $v2$ pipeline). Three classes: Young (Y), Middle-Aged (MA), Elderly (E). Parity gap $\Delta = \tfrac{1}{3}\sum_{c}|R_c - \tfrac{1}{3}|$ (target is $\Delta{=}0$); $\Delta$-rows in bold mark improvement over vanilla.}
  \label{tab:age_per_concept}
  \scriptsize
  \setlength{\tabcolsep}{4pt}
  \begin{tabular}{cl|cccc|cccc}
  \toprule
  & & \multicolumn{4}{c|}{SDXL} & \multicolumn{4}{c}{SANA-1.5} \\
  \cmidrule(lr){3-6}\cmidrule(lr){7-10}
  Mode & Profession & Y & MA & E & $\Delta$ ($\downarrow$) & Y & MA & E & $\Delta$ ($\downarrow$) \\
  \midrule
  \multirow{9}{*}{\rotatebox[origin=c]{90}{Vanilla}}
    & CEO              & 0.117 & 0.849 & 0.034 & 0.344 & 0.000 & 1.000 & 0.000 & 0.444 \\
    & Doctor           & 0.011 & 0.946 & 0.043 & 0.408 & 0.000 & 0.999 & 0.001 & 0.444 \\
    & Pilot            & 0.341 & 0.609 & 0.050 & 0.189 & 0.001 & 0.999 & 0.000 & 0.444 \\
    & Technician       & 0.231 & 0.746 & 0.023 & 0.275 & 0.204 & 0.796 & 0.000 & 0.308 \\
    & Teacher          & 0.182 & 0.793 & 0.025 & 0.306 & 0.007 & 0.993 & 0.000 & 0.440 \\
    & Librarian        & 0.280 & 0.590 & 0.130 & 0.171 & 0.934 & 0.046 & 0.020 & 0.400 \\
    & Nurse            & 0.197 & 0.771 & 0.032 & 0.292 & 0.125 & 0.875 & 0.000 & 0.361 \\
    & Fashion designer & 0.553 & 0.446 & 0.001 & 0.222 & 0.720 & 0.269 & 0.011 & 0.258 \\
  \cmidrule(lr){2-10}
    & Avg.\ $\Delta$   &       &       &       & 0.276 &       &       &       & 0.387 \\
  \midrule
  \multirow{9}{*}{\rotatebox[origin=c]{90}{EquiSteer}}
    & CEO              & 0.180 & 0.412 & 0.408 & \textbf{0.102} & 0.283 & 0.439 & 0.278 & \textbf{0.070} \\
    & Doctor           & 0.053 & 0.696 & 0.251 & \textbf{0.242} & 0.325 & 0.374 & 0.301 & \textbf{0.027} \\
    & Pilot            & 0.209 & 0.426 & 0.365 & \textbf{0.083} & 0.336 & 0.328 & 0.336 & \textbf{0.004} \\
    & Technician       & 0.138 & 0.547 & 0.315 & \textbf{0.142} & 0.356 & 0.341 & 0.303 & \textbf{0.020} \\
    & Teacher          & 0.229 & 0.409 & 0.362 & \textbf{0.070} & 0.146 & 0.699 & 0.155 & \textbf{0.244} \\
    & Librarian        & 0.297 & 0.287 & 0.416 & \textbf{0.055} & 0.329 & 0.275 & 0.396 & \textbf{0.042} \\
    & Nurse            & 0.135 & 0.527 & 0.338 & \textbf{0.132} & 0.347 & 0.346 & 0.307 & \textbf{0.018} \\
    & Fashion designer & 0.192 & 0.449 & 0.359 & \textbf{0.094} & 0.413 & 0.280 & 0.307 & \textbf{0.053} \\
  \cmidrule(lr){2-10}
    & Avg.\ $\Delta$   &       &       &       & \textbf{0.115} &       &       &       & \textbf{0.060} \\
  \bottomrule
  \end{tabular}
\end{table*}

\begin{figure}[t]
\centering
\begin{subfigure}[t]{0.48\linewidth}\centering\includegraphics[width=\linewidth]{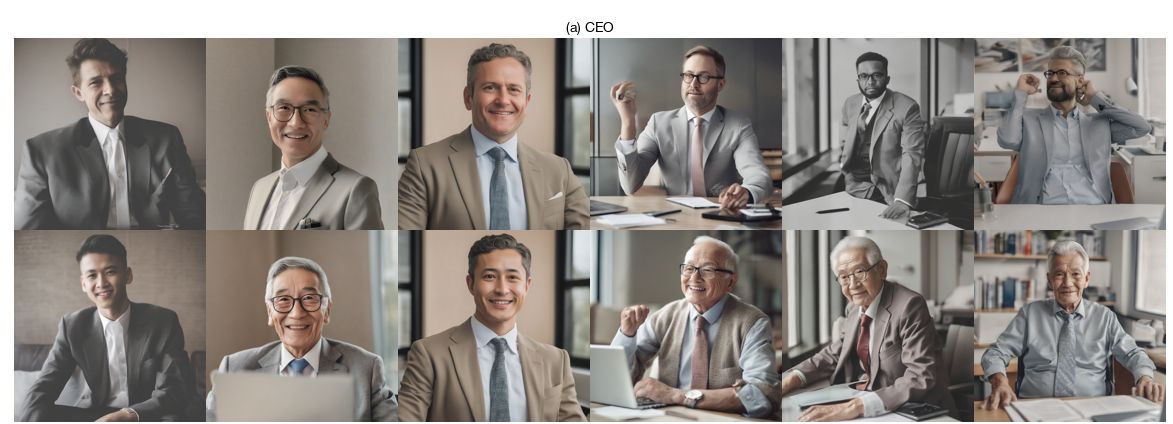}\caption{\textit{CEO}}\end{subfigure}\hfill
\begin{subfigure}[t]{0.48\linewidth}\centering\includegraphics[width=\linewidth]{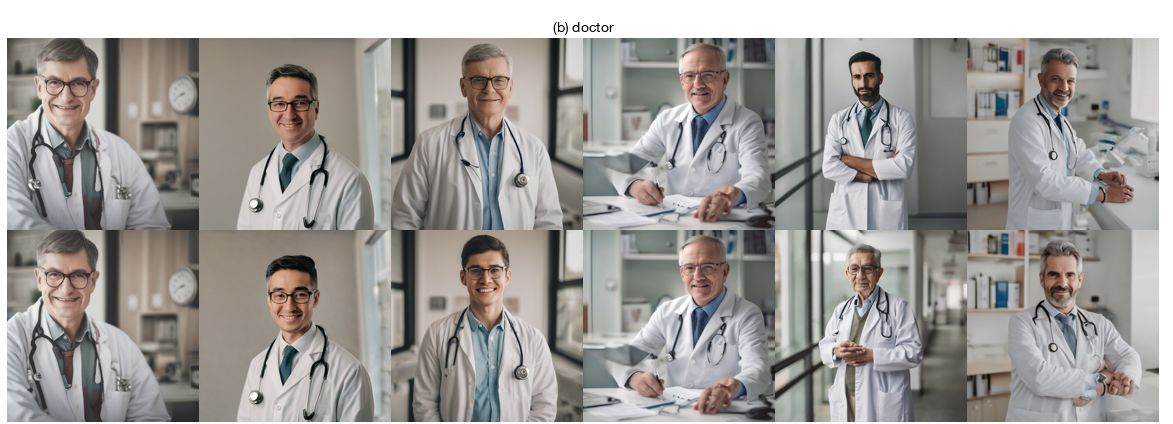}\caption{\textit{Doctor}}\end{subfigure}

\medskip
\begin{subfigure}[t]{0.48\linewidth}\centering\includegraphics[width=\linewidth]{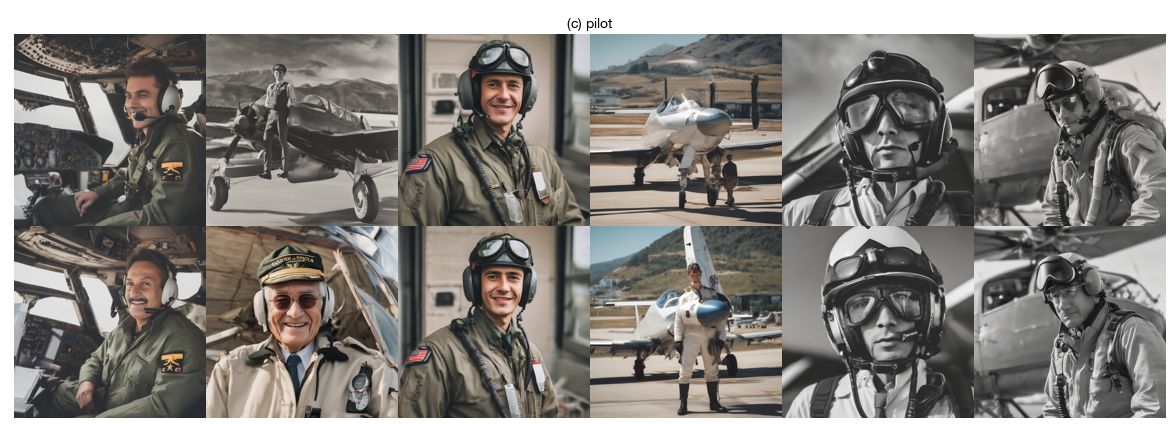}\caption{\textit{Pilot}}\end{subfigure}\hfill
\begin{subfigure}[t]{0.48\linewidth}\centering\includegraphics[width=\linewidth]{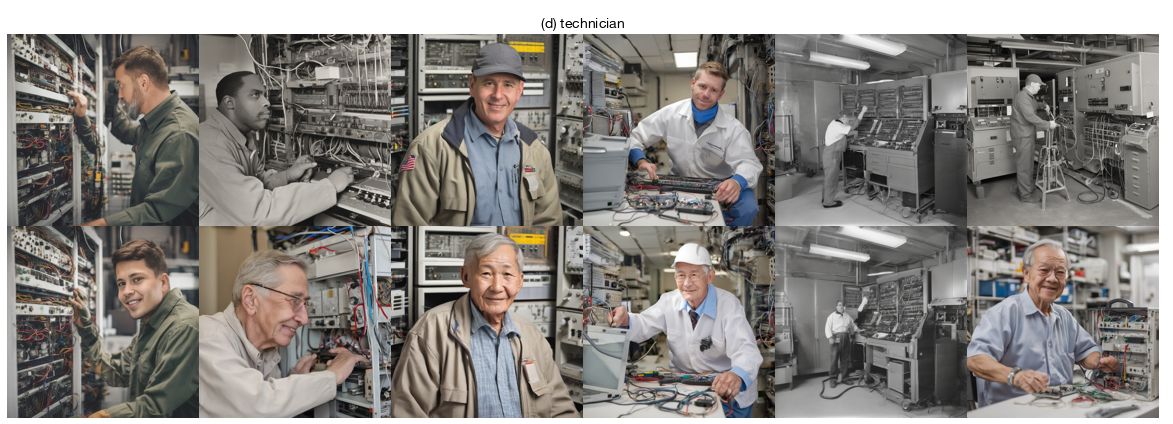}\caption{\textit{Technician}}\end{subfigure}

\medskip
\begin{subfigure}[t]{0.48\linewidth}\centering\includegraphics[width=\linewidth]{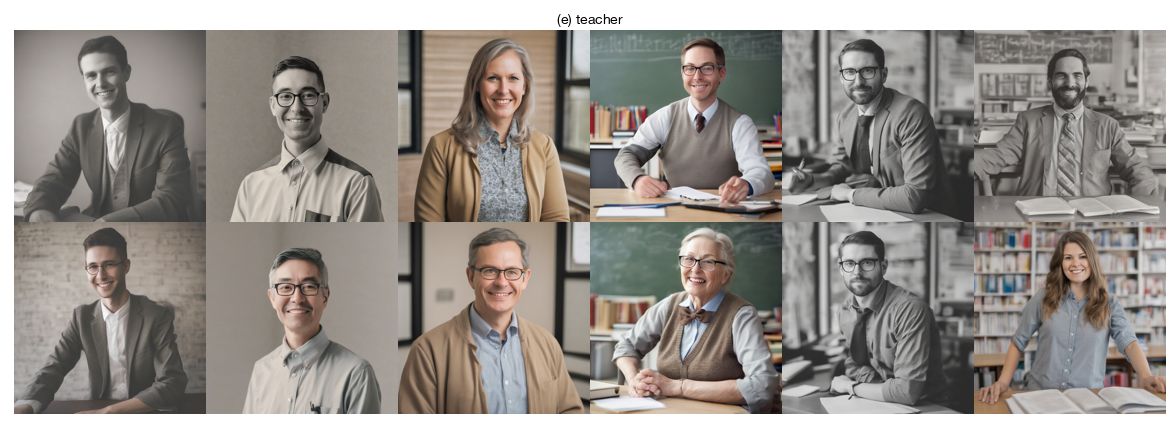}\caption{\textit{Teacher}}\end{subfigure}\hfill
\begin{subfigure}[t]{0.48\linewidth}\centering\includegraphics[width=\linewidth]{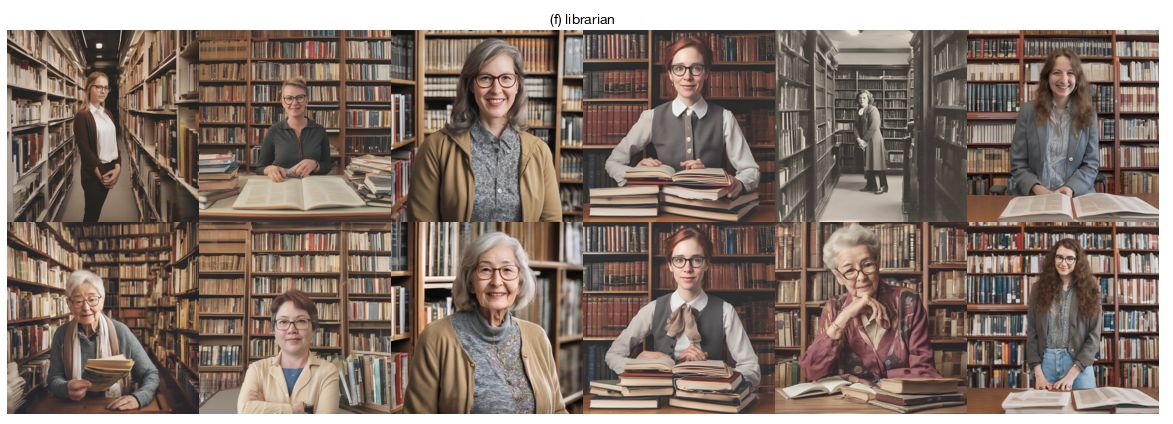}\caption{\textit{Librarian}}\end{subfigure}

\medskip
\begin{subfigure}[t]{0.48\linewidth}\centering\includegraphics[width=\linewidth]{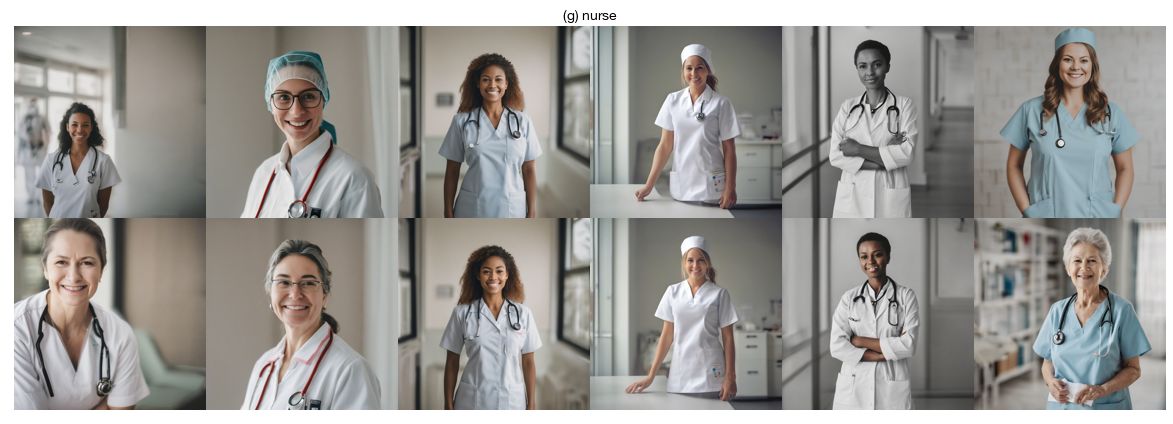}\caption{\textit{Nurse}}\end{subfigure}\hfill
\begin{subfigure}[t]{0.48\linewidth}\centering\includegraphics[width=\linewidth]{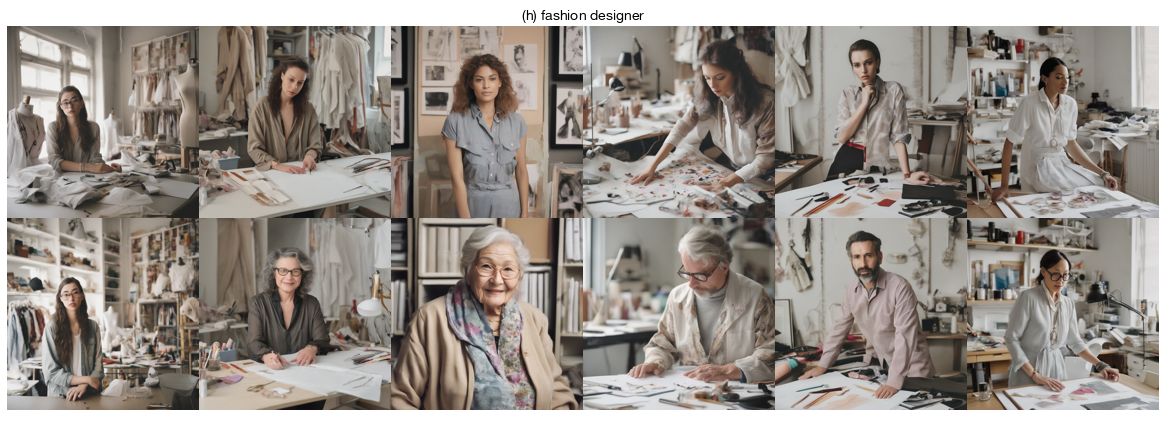}\caption{\textit{Fashion Designer}}\end{subfigure}

\caption{Debiasing of the \textit{age} attribute on SDXL. For each profession, top row: vanilla SDXL; bottom row: EquiSteer.}
\label{fig:qual_age_sdxl}
\end{figure}

\begin{figure}[t]
\centering
\begin{subfigure}[t]{0.48\linewidth}\centering\includegraphics[width=\linewidth]{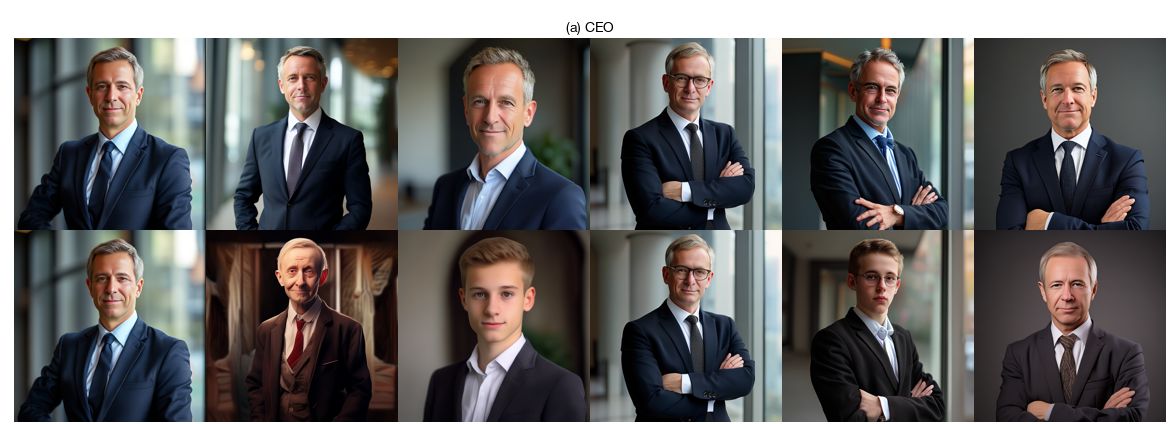}\caption{\textit{CEO}}\end{subfigure}\hfill
\begin{subfigure}[t]{0.48\linewidth}\centering\includegraphics[width=\linewidth]{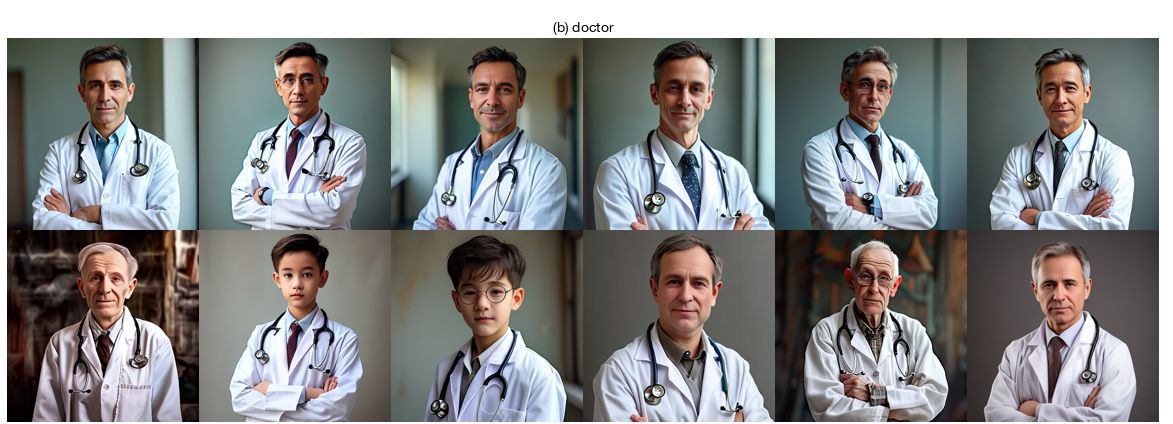}\caption{\textit{Doctor}}\end{subfigure}

\medskip
\begin{subfigure}[t]{0.48\linewidth}\centering\includegraphics[width=\linewidth]{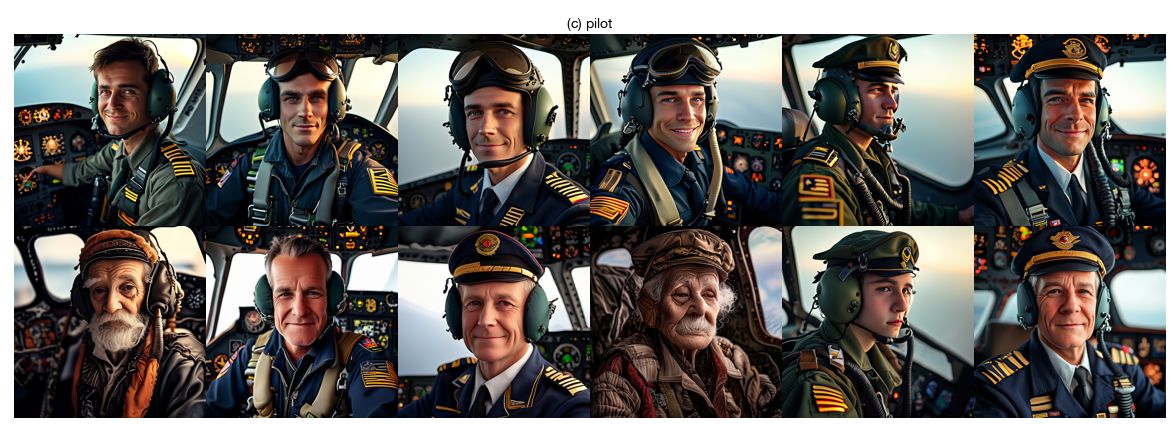}\caption{\textit{Pilot}}\end{subfigure}\hfill
\begin{subfigure}[t]{0.48\linewidth}\centering\includegraphics[width=\linewidth]{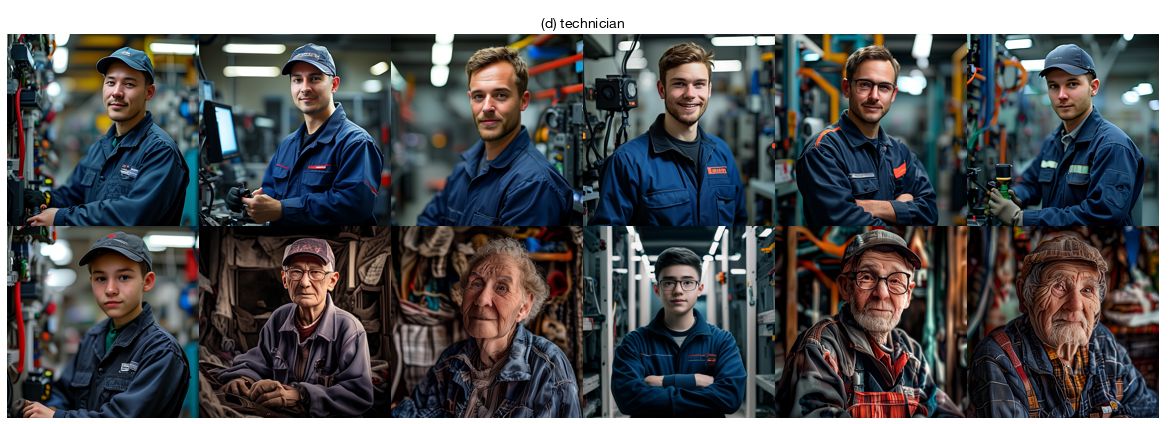}\caption{\textit{Technician}}\end{subfigure}

\medskip
\begin{subfigure}[t]{0.48\linewidth}\centering\includegraphics[width=\linewidth]{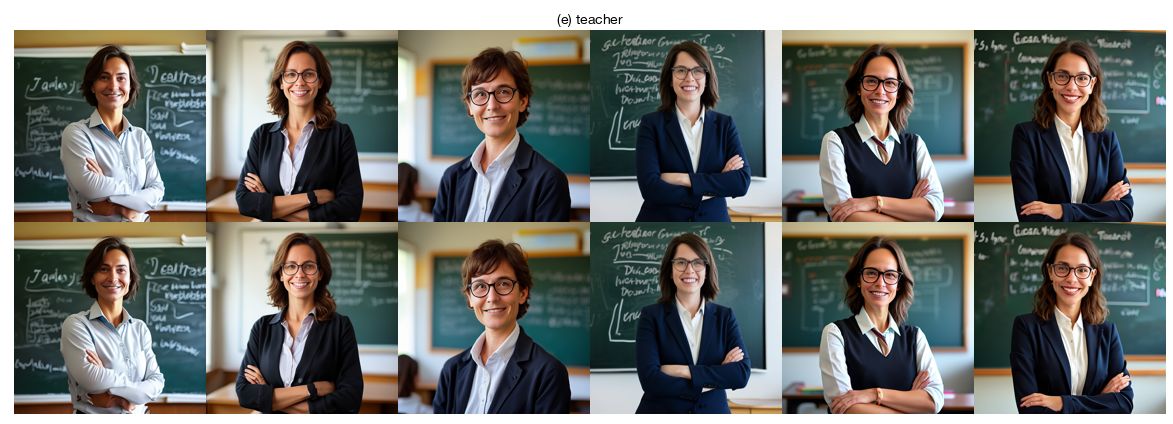}\caption{\textit{Teacher}}\end{subfigure}\hfill
\begin{subfigure}[t]{0.48\linewidth}\centering\includegraphics[width=\linewidth]{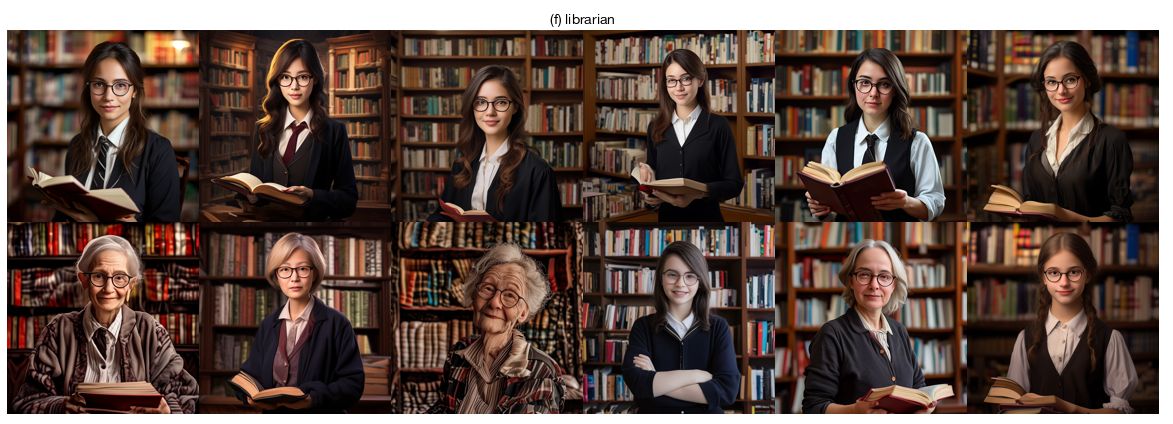}\caption{\textit{Librarian}}\end{subfigure}

\medskip
\begin{subfigure}[t]{0.48\linewidth}\centering\includegraphics[width=\linewidth]{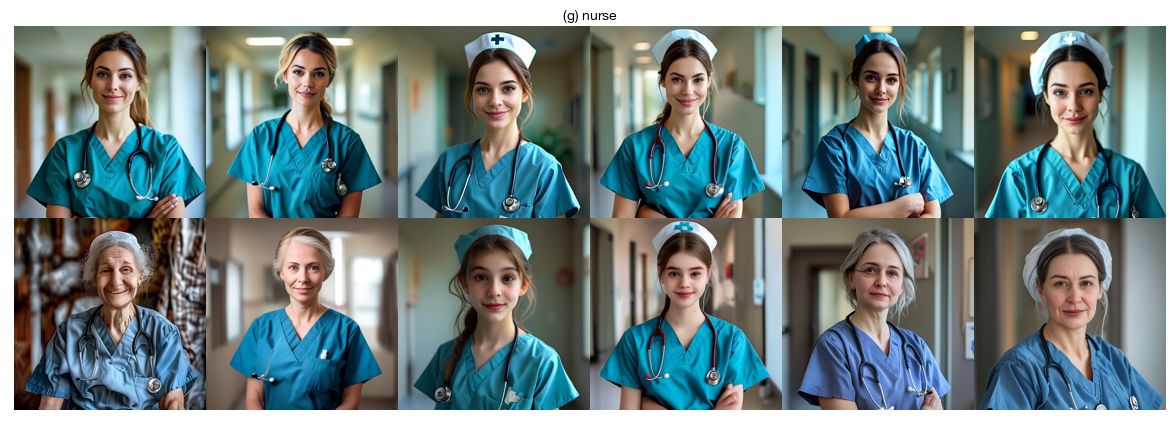}\caption{\textit{Nurse}}\end{subfigure}\hfill
\begin{subfigure}[t]{0.48\linewidth}\centering\includegraphics[width=\linewidth]{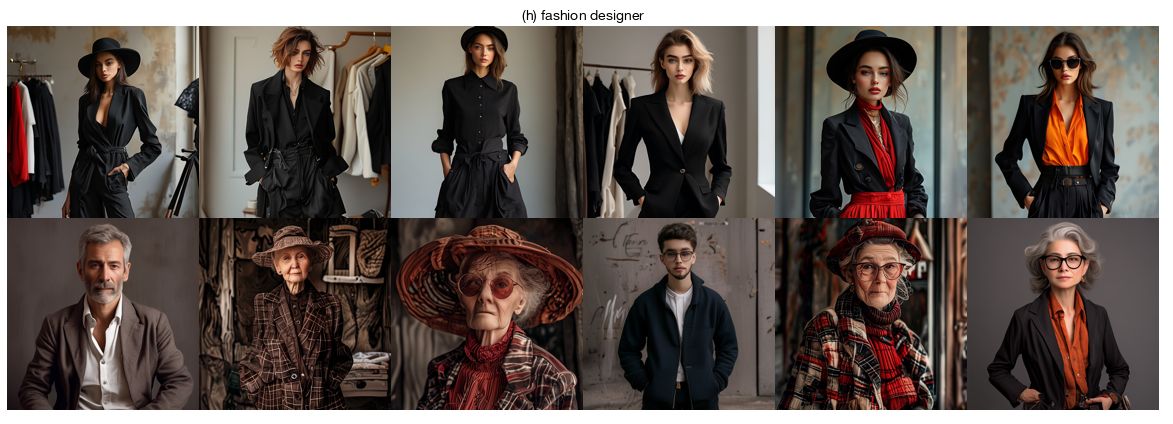}\caption{\textit{Fashion Designer}}\end{subfigure}

\caption{Debiasing of the \textit{age} attribute on SANA-1.5. For each profession, top row: vanilla SANA-1.5; bottom row: EquiSteer.}
\label{fig:qual_age_sana15}
\end{figure}

\noindent \textbf{Qualitative results.}
Fig.~\ref{fig:qual_age_sdxl} and Fig.~\ref{fig:qual_age_sana15} show side-by-side vanilla vs.\ EquiSteer generations for all eight professions on SDXL and SANA-1.5 respectively. For each profession, the top row shows vanilla generations and the bottom row shows EquiSteer. EquiSteer visibly increases the proportion of elderly subjects on professions where vanilla generates almost exclusively middle-aged people (e.g., CEO, doctor, pilot), and increases middle-aged / elderly subjects on professions where vanilla skews young (e.g., librarian, fashion designer on SANA-1.5).

\FloatBarrier
\subsection{Body type}
\label{sec:body_supp}

\noindent \textbf{Method.}
Steering vectors and gate thresholds for the three body-type values are constructed as described in Sec.~\ref{sec:prompts_body} and Sec.~\ref{sec:gate_threshold_supp}, respectively. We report results on the two most capable diffusion backbones, SDXL and SANA-1.5.

\noindent \textbf{Evaluation protocol.}
Following our evaluation protocol, we use the CLIP zero-shot classifier with the templates \textit{``A photo of a slim person''}, \textit{``A photo of a average build person''}, and \textit{``A photo of an heavy person''}. Each generated image is assigned to the class with the highest CLIP score.

\noindent \textbf{Quantitative results.}
Tab.~\ref{tab:body_per_concept} reports per-profession results. EquiSteer reduces the mean parity gap from $0.317$ to $0.220$ on SDXL ($31\%$ reduction) and from $0.390$ to $0.231$ on SANA-1.5 ($41\%$ reduction). Note that across all professions, EquiSteer increases the previously near-zero \textit{heavy} class to between $3\%$ and $36\%$ on SDXL.

In Sec.~\ref{sec:classifier_calibration_supp} we additionally calibrate CLIP classifier with human evaluation and GPT-4o oracle. The results show that (i) GPT-4o oracle is well-agreed with human evaluation, and (ii) that when the CLIP labels are replaced with a GPT-4o oracle on the same generations for body type concept, the measured EquiSteer reduction is substantially larger: $0.257 \to 0.098$ on SDXL ($62\%$ reduction) and $0.365 \to 0.210$ on SANA-1.5 ($42\%$ reduction). This suggests, that CLIP classifier is less reliable on body labels, and CLIP numbers in Tab.~\ref{tab:body_per_concept} are therefore best read as conservative estimates of the underlying debiasing effect.

\begin{table*}[th!]
  \centering
  \caption{Per-profession \textit{body type} class ratios on SDXL and SANA-1.5 (CLIP ViT-L/14 zero-shot, $n{=}1{,}000$ images per cell). Three classes: Slim (S), Average build (A), Heavy (H). Parity gap $\Delta = \tfrac{1}{3}\sum_{c}|R_c - \tfrac{1}{3}|$ (target $\Delta{=}0$); $\Delta$-rows in bold mark improvement over vanilla.}
  \label{tab:body_per_concept}
  \scriptsize
  \setlength{\tabcolsep}{4pt}
  \begin{tabular}{cl|cccc|cccc}
  \toprule
  & & \multicolumn{4}{c|}{SDXL} & \multicolumn{4}{c}{SANA-1.5} \\
  \cmidrule(lr){3-6}\cmidrule(lr){7-10}
  Mode & Profession & S & A & H & $\Delta$ ($\downarrow$) & S & A & H & $\Delta$ ($\downarrow$) \\
  \midrule
  \multirow{9}{*}{\rotatebox[origin=c]{90}{Vanilla}}
    & CEO              & 0.257 & 0.731 & 0.012 & 0.265 & 0.900 & 0.100 & 0.000 & 0.378 \\
    & Doctor           & 0.781 & 0.215 & 0.004 & 0.298 & 0.969 & 0.031 & 0.000 & 0.424 \\
    & Pilot            & 0.028 & 0.916 & 0.056 & 0.388 & 0.001 & 0.999 & 0.000 & 0.444 \\
    & Technician       & 0.000 & 1.000 & 0.000 & 0.444 & 0.000 & 1.000 & 0.000 & 0.444 \\
    & Teacher          & 0.289 & 0.690 & 0.021 & 0.238 & 0.396 & 0.604 & 0.000 & 0.222 \\
    & Librarian        & 0.213 & 0.756 & 0.031 & 0.282 & 0.102 & 0.898 & 0.000 & 0.376 \\
    & Nurse            & 0.930 & 0.060 & 0.010 & 0.398 & 0.933 & 0.067 & 0.000 & 0.400 \\
    & Fashion designer & 0.481 & 0.513 & 0.006 & 0.218 & 0.988 & 0.012 & 0.000 & 0.436 \\
  \cmidrule(lr){2-10}
    & Avg.\ $\Delta$   &       &       &       & 0.317 &       &       &       & 0.391 \\
  \midrule
  \multirow{9}{*}{\rotatebox[origin=c]{90}{EquiSteer}}
    & CEO              & 0.143 & 0.639 & 0.218 & \textbf{0.209} & 0.628 & 0.131 & 0.241 & \textbf{0.196} \\
    & Doctor           & 0.475 & 0.387 & 0.138 & \textbf{0.130} & 0.636 & 0.334 & 0.030 & \textbf{0.202} \\
    & Pilot            & 0.094 & 0.546 & 0.360 & \textbf{0.160} & 0.004 & 0.732 & 0.264 & \textbf{0.266} \\
    & Technician       & 0.001 & 0.971 & 0.028 & \textbf{0.425} & 0.000 & 0.853 & 0.147 & \textbf{0.346} \\
    & Teacher          & 0.197 & 0.671 & 0.132 & \textbf{0.225} & 0.335 & 0.586 & 0.079 & \textbf{0.170} \\
    & Librarian        & 0.099 & 0.856 & 0.045 & 0.348 & 0.144 & 0.767 & 0.089 & \textbf{0.289} \\
    & Nurse            & 0.643 & 0.045 & 0.312 & \textbf{0.206} & 0.671 & 0.066 & 0.263 & \textbf{0.225} \\
    & Fashion designer & 0.417 & 0.247 & 0.336 & \textbf{0.058} & 0.578 & 0.100 & 0.322 & \textbf{0.163} \\
  \cmidrule(lr){2-10}
    & Avg.\ $\Delta$   &       &       &       & \textbf{0.220} &       &       &       & \textbf{0.232} \\
  \bottomrule
  \end{tabular}
\end{table*}

\begin{figure}[t]
\centering
\begin{subfigure}[t]{0.48\linewidth}\centering\includegraphics[width=\linewidth]{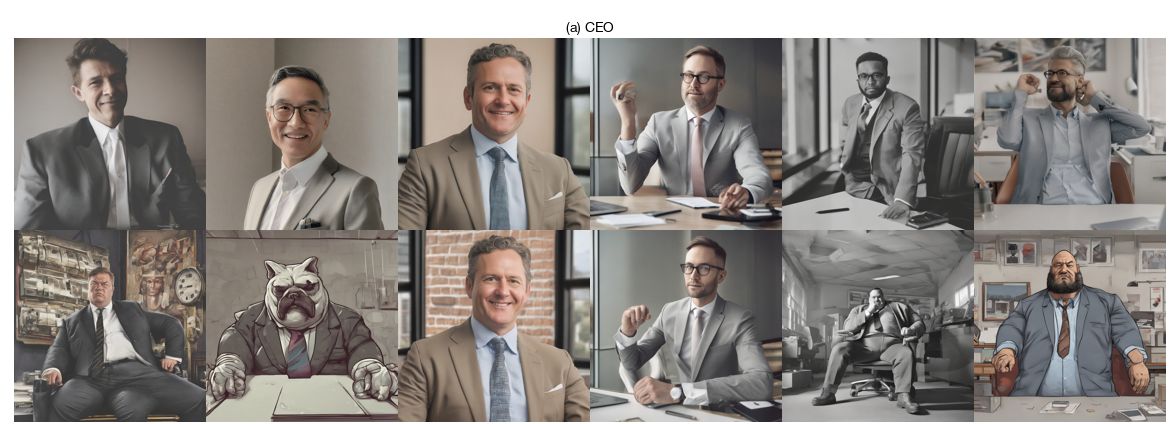}\caption{\textit{CEO}}\end{subfigure}\hfill
\begin{subfigure}[t]{0.48\linewidth}\centering\includegraphics[width=\linewidth]{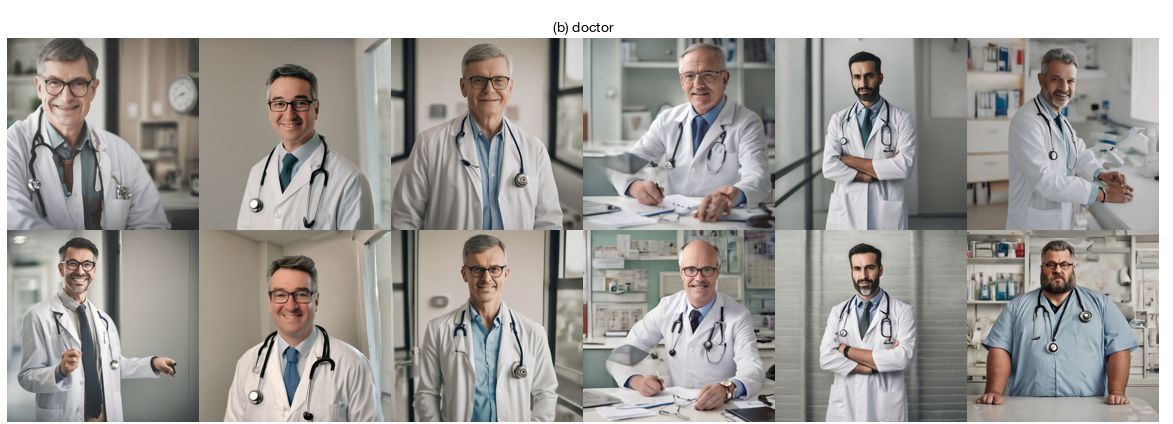}\caption{\textit{Doctor}}\end{subfigure}

\medskip
\begin{subfigure}[t]{0.48\linewidth}\centering\includegraphics[width=\linewidth]{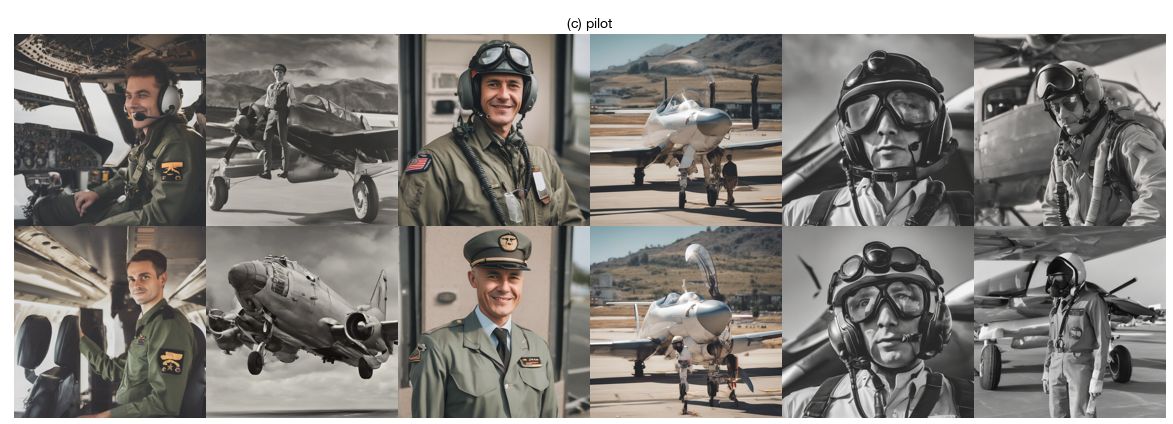}\caption{\textit{Pilot}}\end{subfigure}\hfill
\begin{subfigure}[t]{0.48\linewidth}\centering\includegraphics[width=\linewidth]{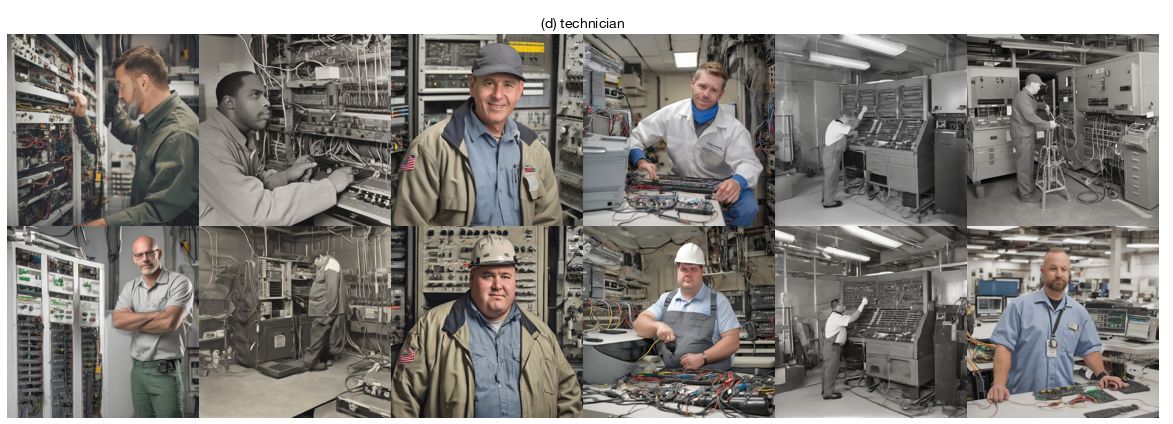}\caption{\textit{Technician}}\end{subfigure}

\medskip
\begin{subfigure}[t]{0.48\linewidth}\centering\includegraphics[width=\linewidth]{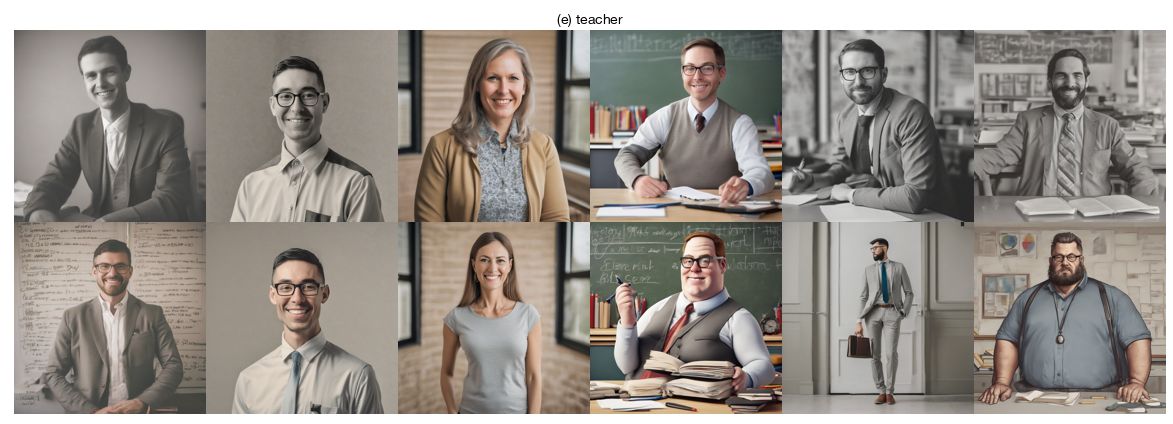}\caption{\textit{Teacher}}\end{subfigure}\hfill
\begin{subfigure}[t]{0.48\linewidth}\centering\includegraphics[width=\linewidth]{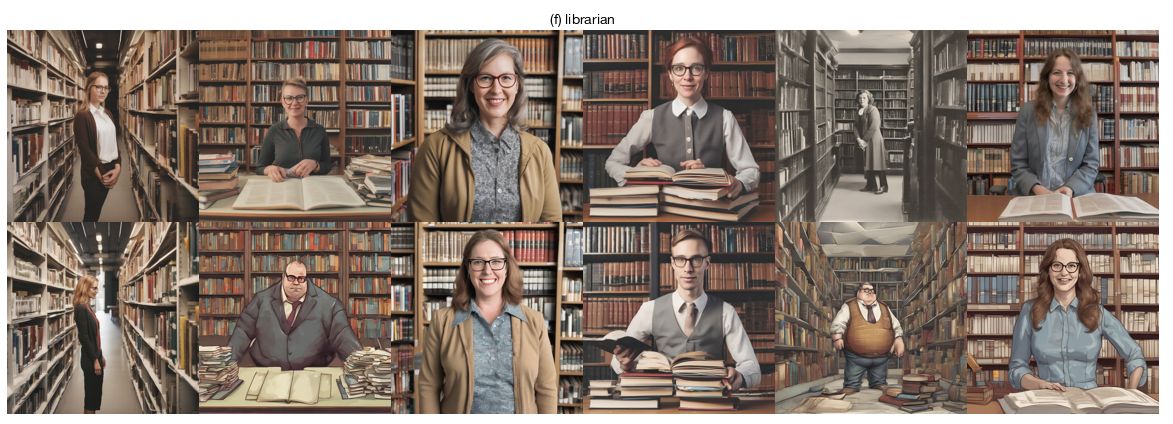}\caption{\textit{Librarian}}\end{subfigure}

\medskip
\begin{subfigure}[t]{0.48\linewidth}\centering\includegraphics[width=\linewidth]{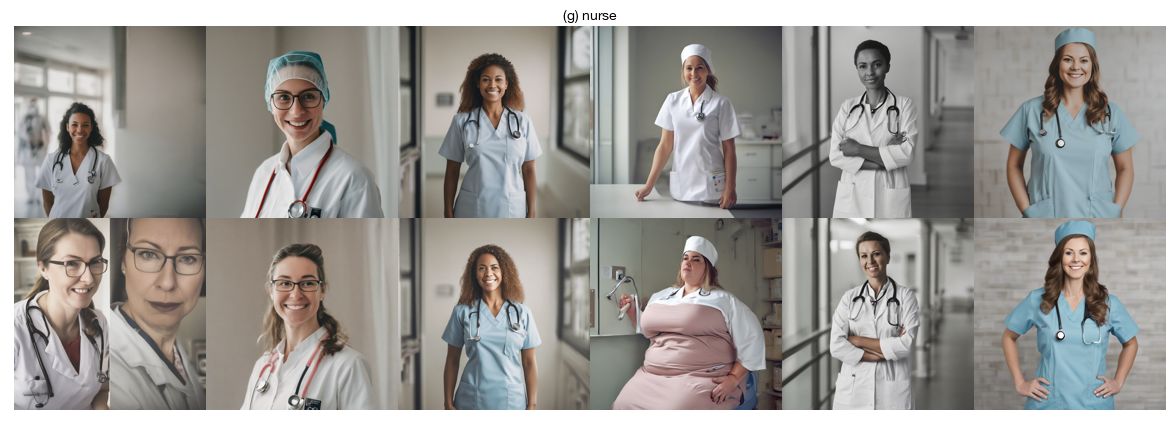}\caption{\textit{Nurse}}\end{subfigure}\hfill
\begin{subfigure}[t]{0.48\linewidth}\centering\includegraphics[width=\linewidth]{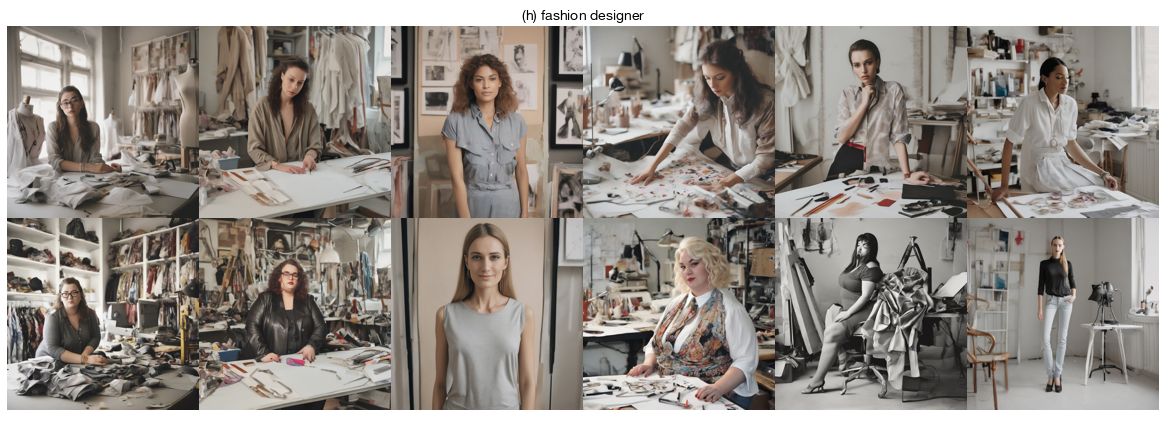}\caption{\textit{Fashion Designer}}\end{subfigure}

\caption{Debiasing of the \textit{body type} attribute on SDXL. For each profession, top row: vanilla SDXL; bottom row: EquiSteer.}
\label{fig:qual_body_sdxl}
\end{figure}

\begin{figure}[t]
\centering
\begin{subfigure}[t]{0.48\linewidth}\centering\includegraphics[width=\linewidth]{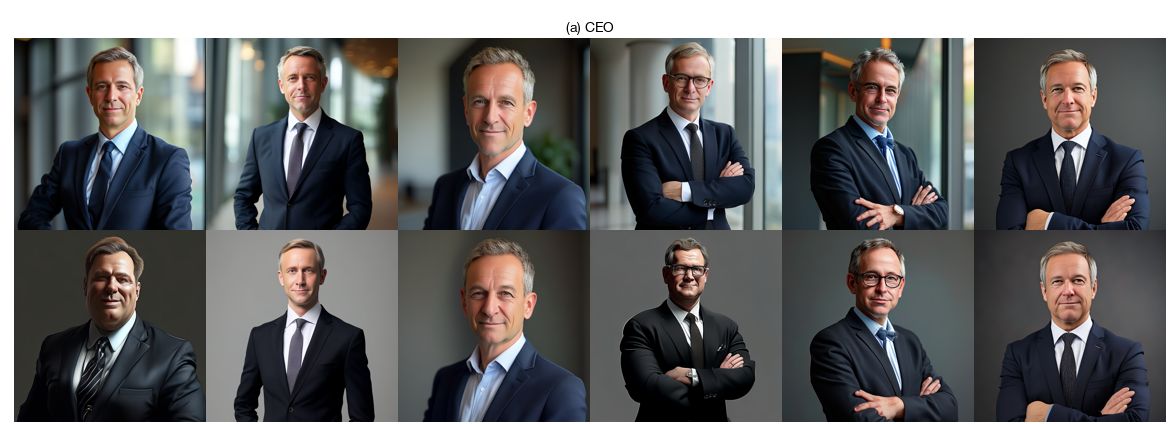}\caption{\textit{CEO}}\end{subfigure}\hfill
\begin{subfigure}[t]{0.48\linewidth}\centering\includegraphics[width=\linewidth]{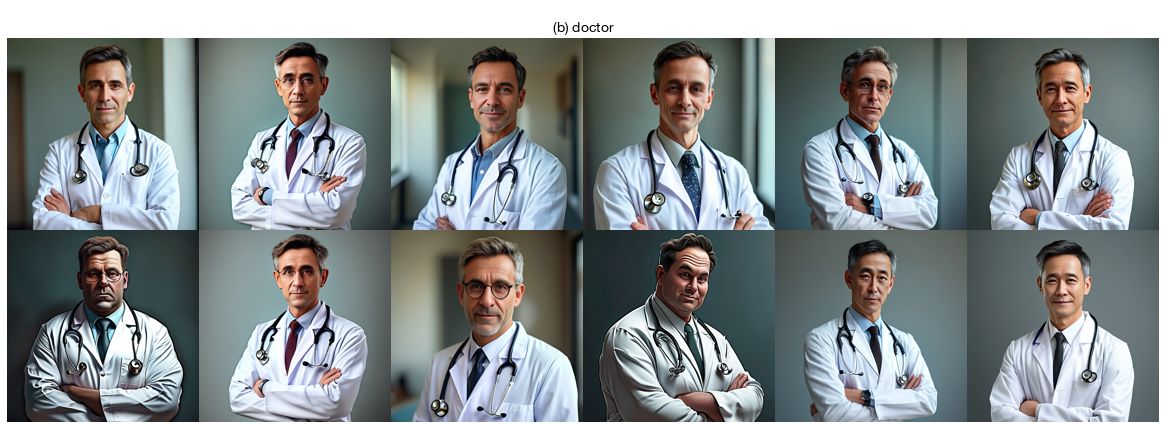}\caption{\textit{Doctor}}\end{subfigure}

\medskip
\begin{subfigure}[t]{0.48\linewidth}\centering\includegraphics[width=\linewidth]{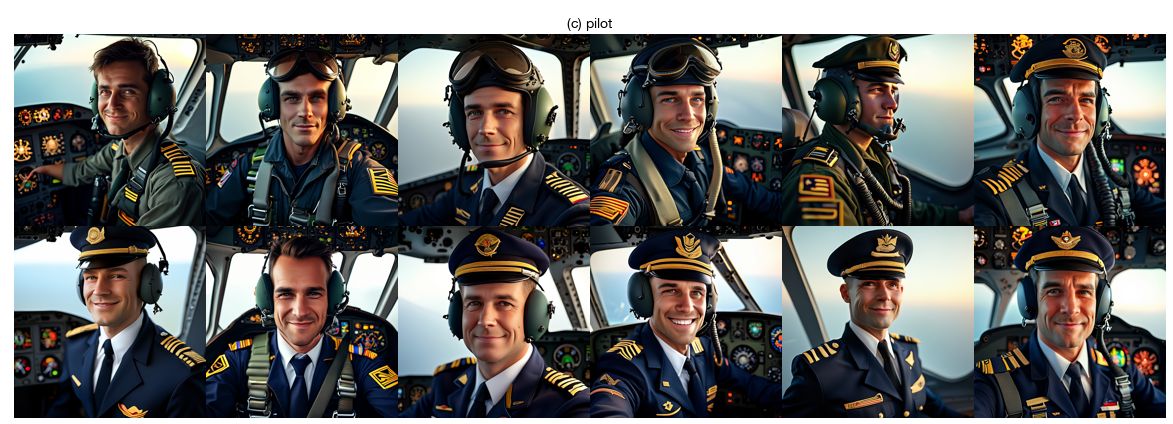}\caption{\textit{Pilot}}\end{subfigure}\hfill
\begin{subfigure}[t]{0.48\linewidth}\centering\includegraphics[width=\linewidth]{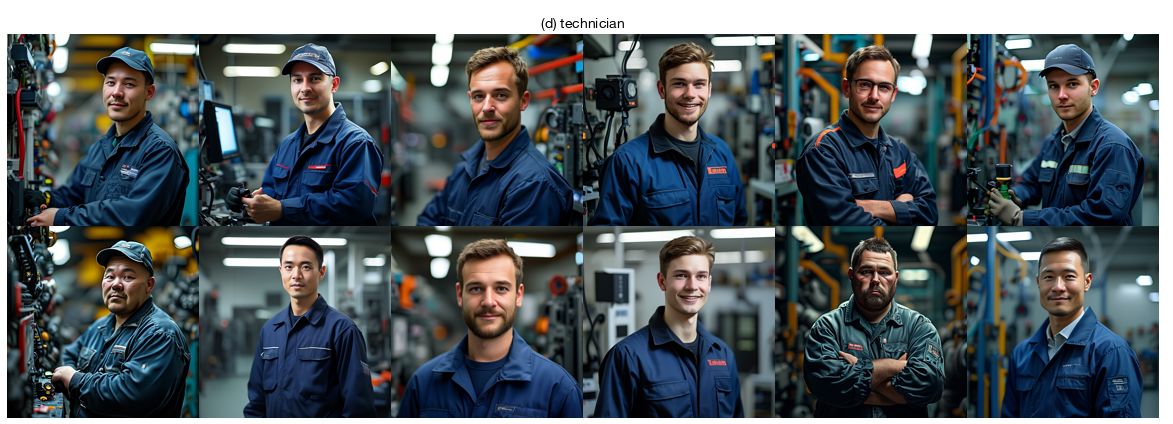}\caption{\textit{Technician}}\end{subfigure}

\medskip
\begin{subfigure}[t]{0.48\linewidth}\centering\includegraphics[width=\linewidth]{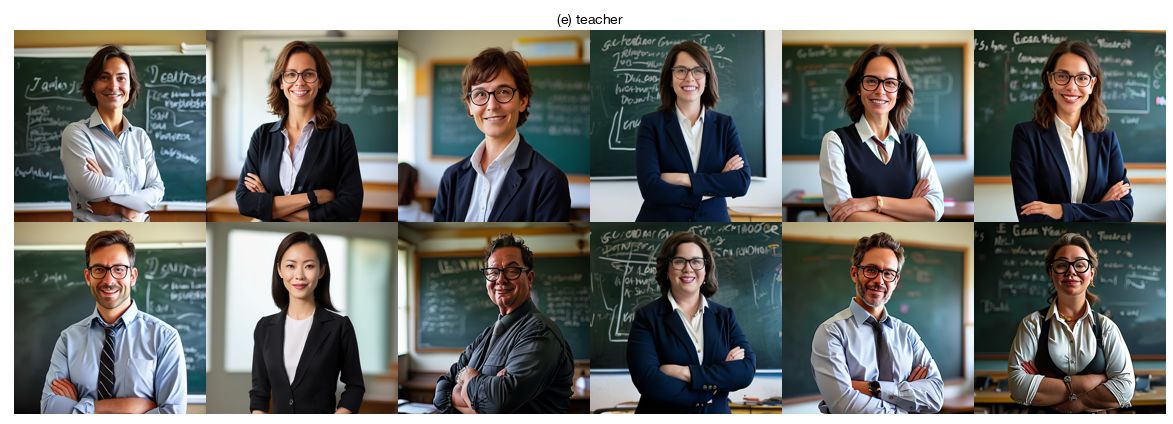}\caption{\textit{Teacher}}\end{subfigure}\hfill
\begin{subfigure}[t]{0.48\linewidth}\centering\includegraphics[width=\linewidth]{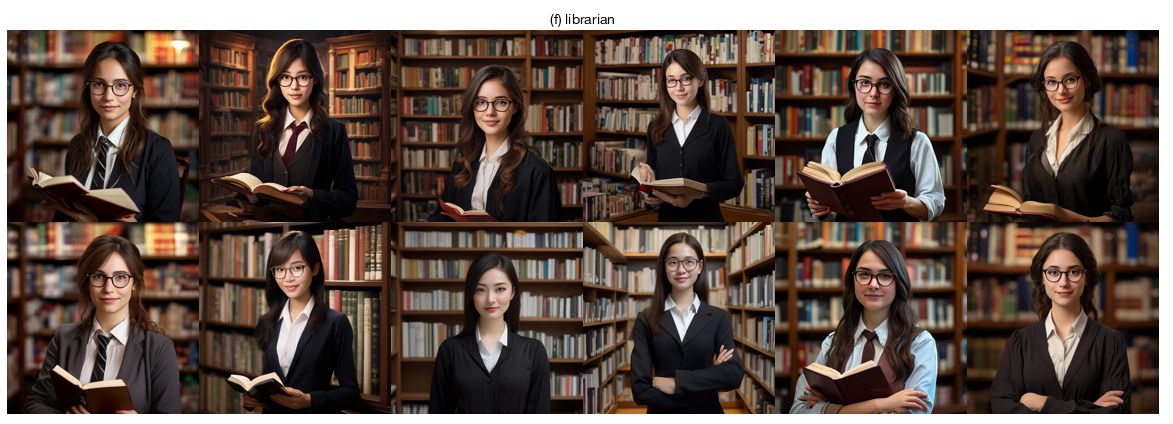}\caption{\textit{Librarian}}\end{subfigure}

\medskip
\begin{subfigure}[t]{0.48\linewidth}\centering\includegraphics[width=\linewidth]{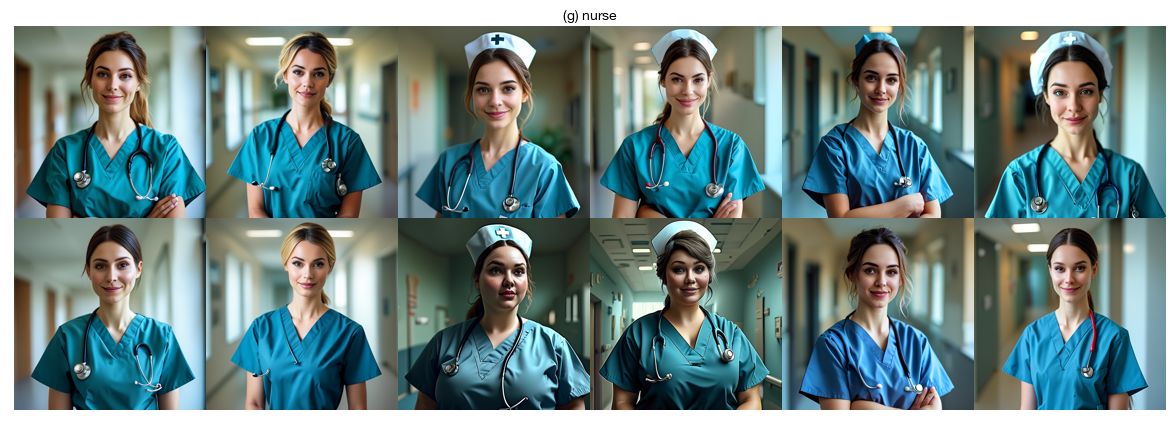}\caption{\textit{Nurse}}\end{subfigure}\hfill
\begin{subfigure}[t]{0.48\linewidth}\centering\includegraphics[width=\linewidth]{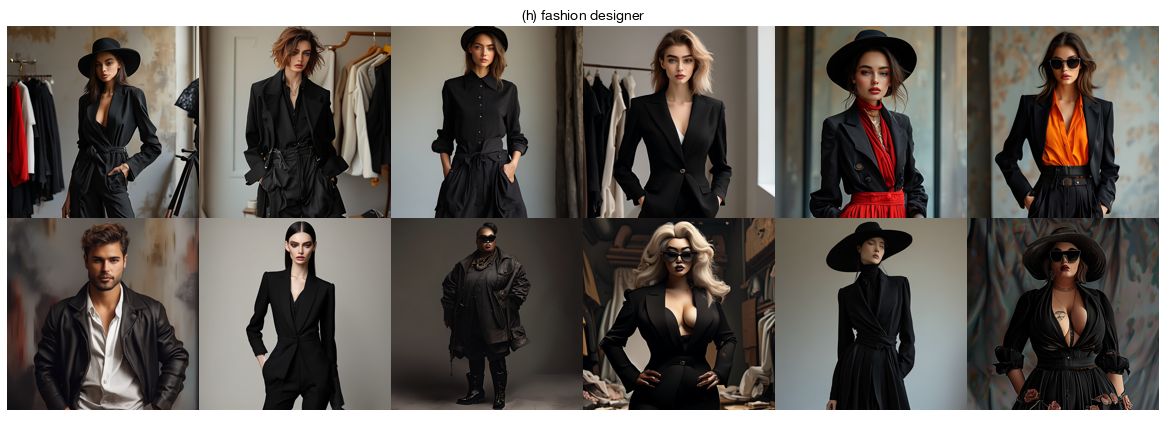}\caption{\textit{Fashion Designer}}\end{subfigure}

\caption{Debiasing of the \textit{body type} attribute on SANA-1.5. For each profession, top row: vanilla SANA-1.5; bottom row: EquiSteer.}
\label{fig:qual_body_sana15}
\end{figure}

\noindent \textbf{Qualitative results.}
Fig.~\ref{fig:qual_body_sdxl} and Fig.~\ref{fig:qual_body_sana15} show side-by-side vanilla vs.\ EquiSteer generations on SDXL and SANA-1.5 respectively.

\FloatBarrier
\subsection{Eyeglasses}
\label{sec:eyeglasses_supp}

\noindent \textbf{Method.}
The \textit{eyeglasses} concept is binary, with two possible attribute values: (\textit{eyeglasses}, \textit{no eyeglasses}). In principle, EquiSteer would require a separate steering vector for each value. However, diffusion models are known to handle negation poorly (e.g., prompts such as ``a man with no eyeglasses'' may still produce eyeglasses), making the steering direction learned for \textit{no eyeglasses} unreliable. 

To handle such negated attributes, we treat \textit{no eyeglasses} not as a separate ``addition'' direction, but as \emph{deletion} of the \textit{eyeglasses} attribute. Concretely, we use the CASteer~\cite{gaintseva2025casteer} erasure update with the steering vector computed for \textit{eyeglasses} and a fixed erasure strength $\beta=2$. That is, when the sampled attribute is $a=\textit{no eyeglasses}$, we replace Eq.8 with
\begin{equation}
    ca^{out\_new}_{lt}
    =
    ca^{out\_tmp}_{lt} - \beta\,\langle ca^{out\_tmp}_{lt}, s_{lt}^{\textit{eyeglasses}}\rangle\, s_{lt}^{\textit{eyeglasses}}.
    \label{eq:thr_add_no_glasses}
\end{equation}
We provide the corresponding procedure in Algorithm~\ref{alg:fairsteer_eyeglasses}.

The single \textit{eyeglasses} steering vector and its gate threshold are constructed as described in Sec.~\ref{sec:prompts_eyeglasses} and Sec.~\ref{sec:gate_threshold_supp}, respectively. We report results on the two most capable diffusion backbones, SDXL and SANA-1.5.

\begin{algorithm}[t]
\caption{\textsc{EquiSteer} with binary-negation handling for \textit{eyeglasses}}
\label{alg:fairsteer_eyeglasses}
\KwIn{
Prompt $p$; binary attribute set $\mathcal{A}=\{\textit{eyeglasses},\ \textit{no eyeglasses}\}$; steps $t=0,\dots,T-1$; CA layers $l=1,\dots,L$;\\
CA outputs $\{ca^{out}_{ltk}\in\mathbb{R}^{d}\}$ (token index $k$);\\
Steering vectors $\{s^{a}_{lt}\in\mathbb{R}^{d}\}$ (computed only for $a=\textit{eyeglasses}$);\\
Gating layer $l^{gate}$; thresholds $\{thr^{a}\}$;\\
Adaptive magnitudes $\{dp^{a}_{\text{mean}}(l,t)\}$ for $a=\textit{eyeglasses}$;\\
Orthogonalization operators $\{P_{lt}=I-U_{lt}U_{lt}^{\top}\}$ (precomputed offline);\\
Fixed erasure strength $\beta=2$; renormalization flag \texttt{renorm}.
}
\KwOut{Modified CA outputs used during denoising.}

\BlankLine
\textbf{(1) Gate evaluation (once per run).}\;
\For{$a \in \mathcal{A}$}{
    \tcp{For the negated value, we still gate using the positive-direction detector.}
    \If{$a=\textit{no eyeglasses}$}{
        $dp^{a}_{l^{gate}0} \leftarrow \max_{k}\ \langle ca^{out}_{l^{gate}0k},\ s^{\textit{eyeglasses}}_{l^{gate}0}\rangle$\;
    }
    \Else{
        $dp^{a}_{l^{gate}0} \leftarrow \max_{k}\ \langle ca^{out}_{l^{gate}0k},\ s^{a}_{l^{gate}0}\rangle$\;
    }
}
\If(\tcp*[f]{attribute-specific prompt}){$\exists a \in \mathcal{A}:\ dp^{a}_{l^{gate}0} > thr^{a}$}{
    \Return \tcp*[f]{Vanilla generation (no intervention)}\;
}

\BlankLine
\textbf{(2) Sample target attribute for neutral prompt.}\;
Sample $a \sim \mathrm{Uniform}(\mathcal{A})$\;

\BlankLine
\textbf{(3) Apply intervention at each denoising step and CA layer.}\;
\For{$t \leftarrow 0$ \KwTo $T-1$}{
  \For{$l \leftarrow 1$ \KwTo $L$}{
    \ForEach{image token $k$}{
      \tcp{(3a) Orthogonalize w.r.t.\ attribute subspace}
      $ca^{out\_tmp}_{ltk} \leftarrow P_{lt}\, ca^{out}_{ltk}$\;

      \tcp{(3b) Binary-negation handling}
      \uIf{$a=\textit{eyeglasses}$}{
          $\alpha \leftarrow dp^{\textit{eyeglasses}}_{\text{mean}}(l,t)$\;
          $ca^{out\_new}_{ltk} \leftarrow ca^{out\_tmp}_{ltk} + \alpha\, s^{\textit{eyeglasses}}_{lt}$\;
      }
      \Else(\tcp*[f]{\textit{no eyeglasses} as deletion of \textit{eyeglasses}}){
          $s \leftarrow s^{\textit{eyeglasses}}_{lt}$\;
          $ca^{out\_new}_{ltk} \leftarrow ca^{out\_tmp}_{ltk} - \beta\,\langle ca^{out\_tmp}_{ltk}, s\rangle\, s$\;
      }

      \tcp{(3c) Optional renormalization for stability}
      \If{\textnormal{\texttt{renorm}}}{
        $ca^{out\_new}_{ltk} \leftarrow \|ca^{out}_{ltk}\|_2\,
        \dfrac{ca^{out\_new}_{ltk}}{\|ca^{out\_new}_{ltk}\|_2+\varepsilon}$\;
      }

      $ca^{out}_{ltk} \leftarrow ca^{out\_new}_{ltk}$\;
    }
  }
}
\end{algorithm}

\noindent \textbf{Evaluation protocol.}
In our main experiments on all the concepts, we use CLIP ViT-L/14 to measure attribute presence in generated images.
For the \textit{eyeglasses} concept, we initially follow the same protocol and evaluate a CLIP ViT-L/14 zero-shot classifier with the prompts \textit{``A photo of a person''} and \textit{``A photo of a person wearing eyeglasses''} (Tab.~\ref{tab:eyeglasses_per_concept_parity_delta_combined_v2}). 
However, we find that this CLIP setup tends to favor the more general prompt \textit{``A photo of a person''}, which systematically underestimates the presence of eyeglasses.
To obtain a more reliable measurement, we additionally report results using a VQA-based classifier, BLIP-vqa-capfilt-large~\cite{li2022blip}.
Specifically, for each image we ask: \textit{``Is any person in this image wearing eyeglasses? Answer yes or no.''}
Results under this evaluation are shown in Tab.~\ref{tab:eyeglasses_per_concept_parity_delta_combined_v2}.In addition, in Sec.~\ref{sec:classifier_calibration_supp} we additionally present metrics obtained with GPT-4o oracle. 

\noindent \textbf{Quantitative results.}
Across all three backbones, EquiSteer consistently shifts the eyeglasses ratios toward the desired parity target (0.5) for most professions (Tab.~\ref{tab:eyeglasses_per_concept_parity_delta_combined_v2}).
Under the BLIP-VQA evaluation (upper block of the same table), the effect is particularly clear for professions with strong initial skew, such as \textit{nurse} (SD-1.5: $0.10\!\to\!0.47$; SANA: $0.00\!\to\!0.39$) and \textit{pilot} (SANA: $0.17\!\to\!0.58$), bringing the ratios substantially closer to the target.
Under the CLIP zero-shot classifier (lower block), the qualitative trends are similar but the absolute scores are often lower, consistent with CLIP under-detecting eyeglasses due to the bias toward the generic prompt.
Overall, these results confirm that EquiSteer can debias binary attributes such as eyeglasses, and highlight that VQA-based evaluation provides a more faithful signal for this concept than the CLIP prompt-pair classifier.

\begin{table*}[th!]
  \centering
  \caption{Per-profession eyeglasses ratios on SDXL and SANA-1.5, evaluated with BLIP-VQA and CLIP ViT-L/14. ``Vanilla'' and ``EquiSteer'' report the fraction of generated images classified as containing eyeglasses. $\Delta = |R_{\text{EquiSteer}} - 0.5| - |R_{\text{Vanilla}} - 0.5|$ measures the change in distance to the 0.5 target; negative values indicate improvement.}
  \label{tab:eyeglasses_per_concept_parity_delta_combined_v2}
  \scriptsize
  \setlength{\tabcolsep}{4pt}
  \begin{tabular}{cl|rrr|rrr}
  \toprule
  & & \multicolumn{3}{c|}{SDXL} & \multicolumn{3}{c}{SANA-1.5} \\
  \cmidrule(lr){3-5}\cmidrule(lr){6-8}
  Classifier & Concept & Vanilla & EquiSteer & $\Delta$ ($\downarrow$) & Vanilla & EquiSteer & $\Delta$ ($\downarrow$) \\
  \midrule
  \multirow{9}{*}{\rotatebox[origin=c]{90}{BLIP-VQA}}
  & CEO              & 0.632 & 0.604 & $-0.028$ & 0.447 & 0.868 & $+0.315$ \\
  & doctor           & 0.881 & 0.558 & $-0.323$ & 0.522 & 0.895 & $+0.373$ \\
  & fashion designer & 0.409 & 0.526 & $-0.065$ & 0.312 & 0.752 & $+0.064$ \\
  & librarian        & 0.943 & 0.537 & $-0.406$ & 0.932 & 0.966 & $+0.034$ \\
  & nurse            & 0.168 & 0.525 & $-0.307$ & 0.000 & 0.693 & $-0.307$ \\
  & pilot            & 0.297 & 0.628 & $-0.075$ & 0.142 & 0.715 & $-0.143$ \\
  & teacher          & 0.912 & 0.548 & $-0.364$ & 0.952 & 0.986 & $+0.034$ \\
  & technician       & 0.690 & 0.688 & $-0.002$ & 0.466 & 0.852 & $+0.318$ \\
  \cmidrule(lr){2-8}
  & Avg.\ $\Delta$   &       &       & $\mathbf{-0.196}$ &       &       & $+0.086$ \\
  \midrule
  \multirow{9}{*}{\rotatebox[origin=c]{90}{CLIP ViT-L/14}}
  & CEO              & 0.496 & 0.554 & $+0.050$ & 0.390 & 0.727 & $+0.117$ \\
  & doctor           & 0.098 & 0.090 & $+0.008$ & 0.002 & 0.041 & $-0.039$ \\
  & fashion designer & 0.341 & 0.555 & $-0.104$ & 0.298 & 0.746 & $+0.044$ \\
  & librarian        & 0.225 & 0.369 & $-0.144$ & 0.895 & 0.714 & $-0.181$ \\
  & nurse            & 0.051 & 0.287 & $-0.236$ & 0.000 & 0.420 & $-0.420$ \\
  & pilot            & 0.125 & 0.472 & $-0.347$ & 0.038 & 0.544 & $-0.418$ \\
  & teacher          & 0.597 & 0.494 & $-0.091$ & 0.901 & 0.921 & $+0.020$ \\
  & technician       & 0.300 & 0.467 & $-0.167$ & 0.534 & 0.865 & $+0.331$ \\
  \cmidrule(lr){2-8}
  & Avg.\ $\Delta$   &       &       & $\mathbf{-0.129}$ &       &       & $\mathbf{-0.068}$ \\
  \bottomrule
  \end{tabular}
  \end{table*}

\begin{figure}[t]
\centering
\begin{subfigure}[t]{0.48\linewidth}
  \centering
  \includegraphics[width=\linewidth]{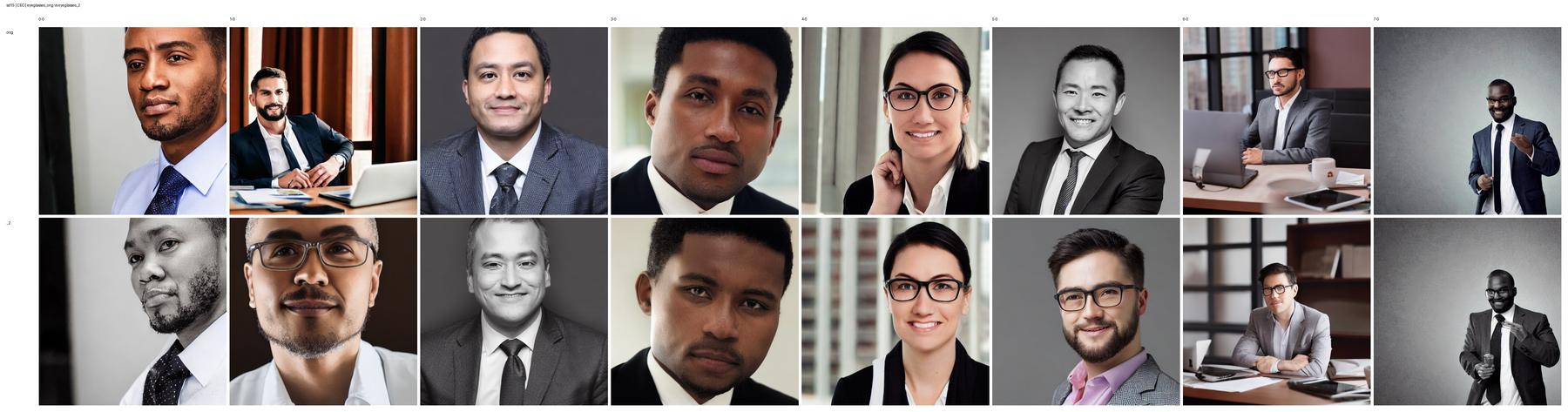}
  \caption{\textit{CEO}}
\end{subfigure}\hfill
\begin{subfigure}[t]{0.48\linewidth}
  \centering
  \includegraphics[width=\linewidth]{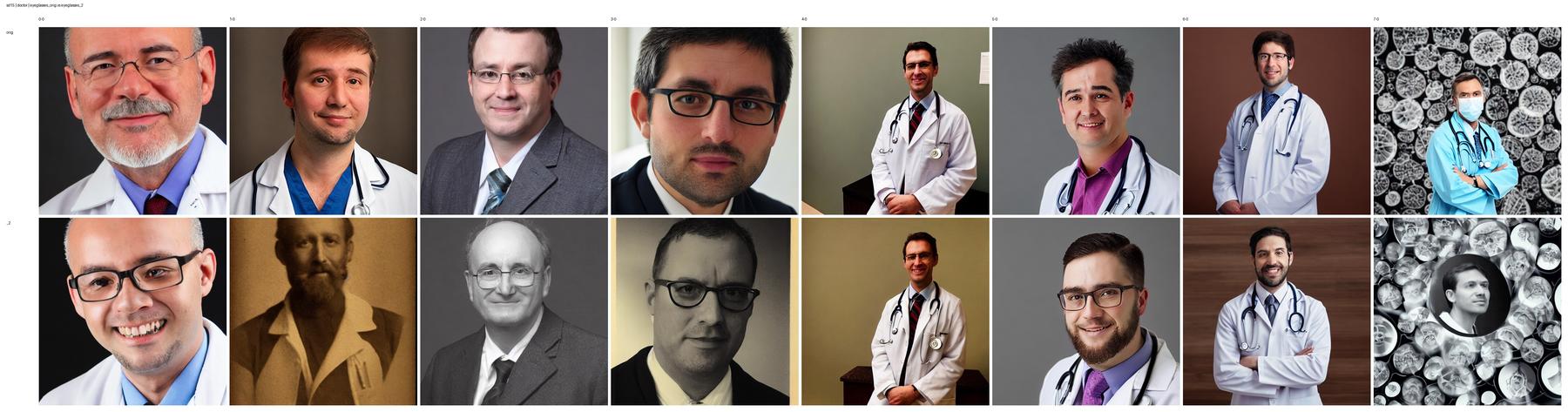}
  \caption{\textit{Doctor}}
\end{subfigure}

\begin{subfigure}[t]{0.48\linewidth}
  \centering
  \includegraphics[width=\linewidth]{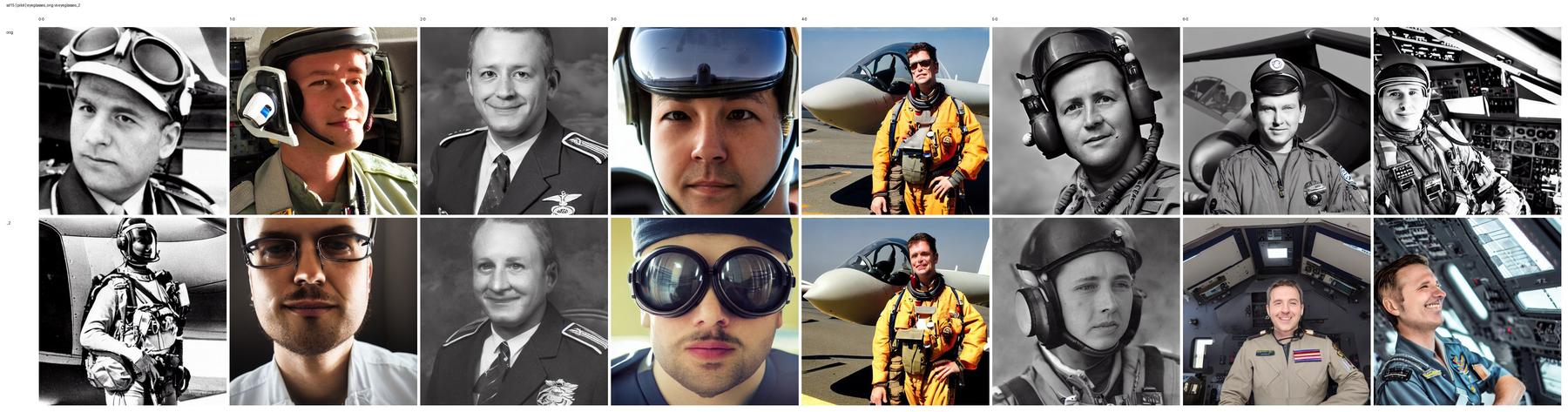}
  \caption{\textit{Pilot}}
\end{subfigure}\hfill
\begin{subfigure}[t]{0.48\linewidth}
  \centering
  \includegraphics[width=\linewidth]{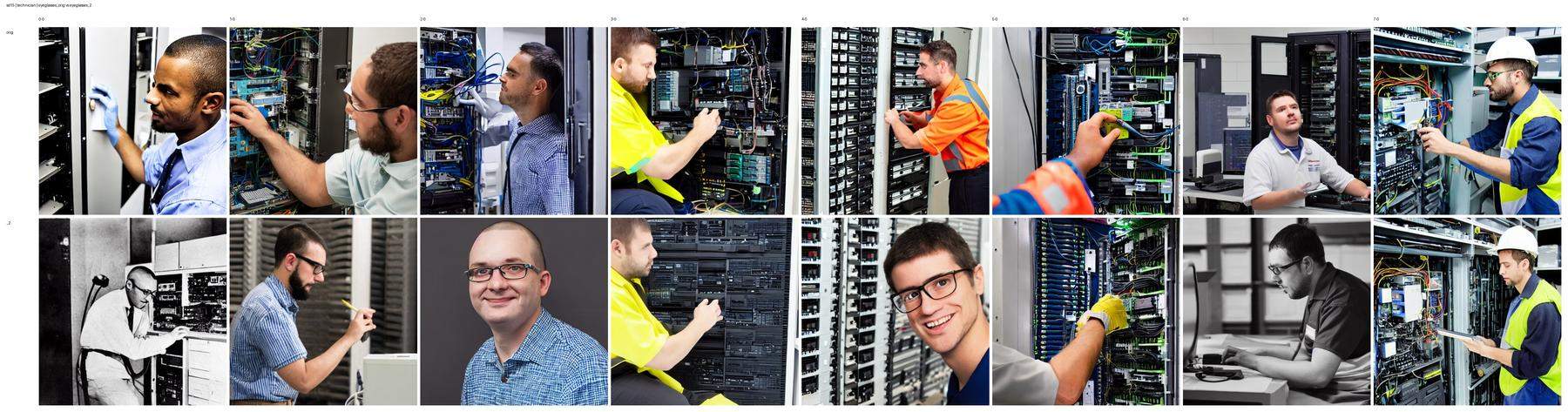}
  \caption{\textit{Technician}}
\end{subfigure}

\begin{subfigure}[t]{0.48\linewidth}
  \centering
  \includegraphics[width=\linewidth]{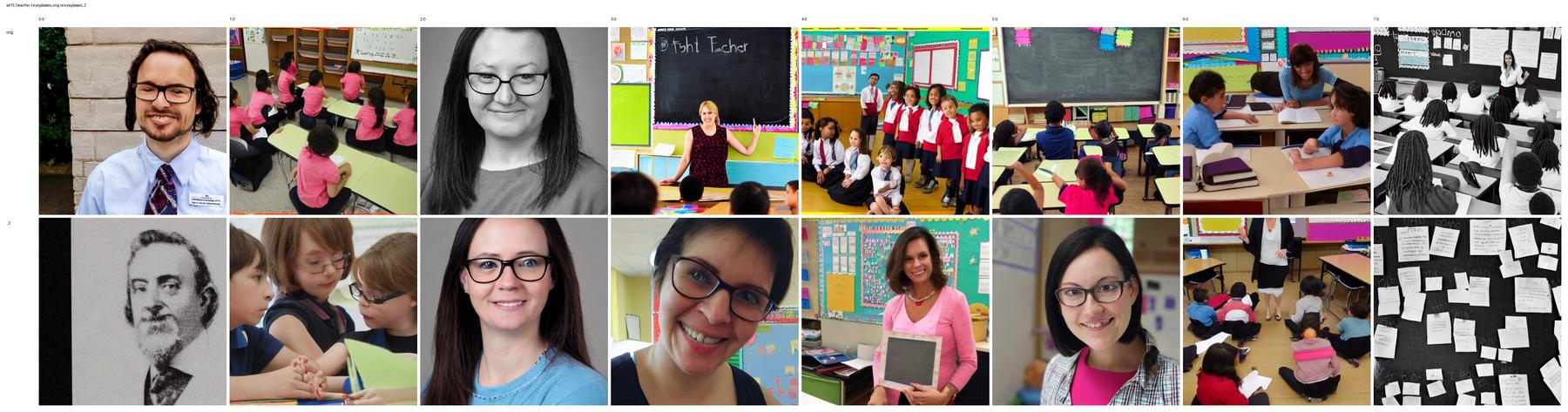}
  \caption{\textit{Teacher}}
\end{subfigure}\hfill
\begin{subfigure}[t]{0.48\linewidth}
  \centering
  \includegraphics[width=\linewidth]{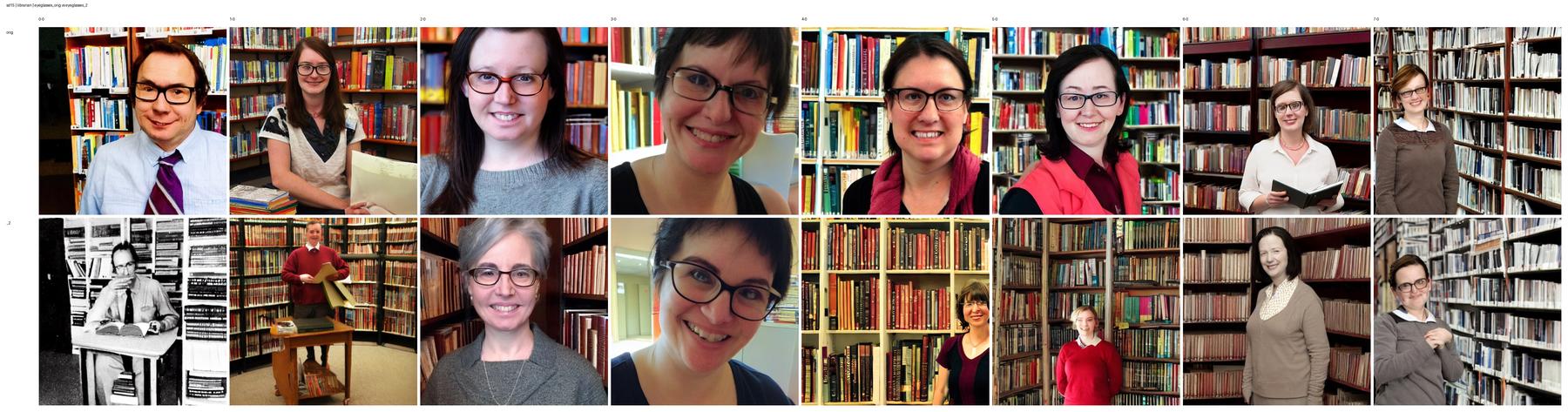}
  \caption{\textit{Librarian}}
\end{subfigure}

\begin{subfigure}[t]{0.48\linewidth}
  \centering
  \includegraphics[width=\linewidth]{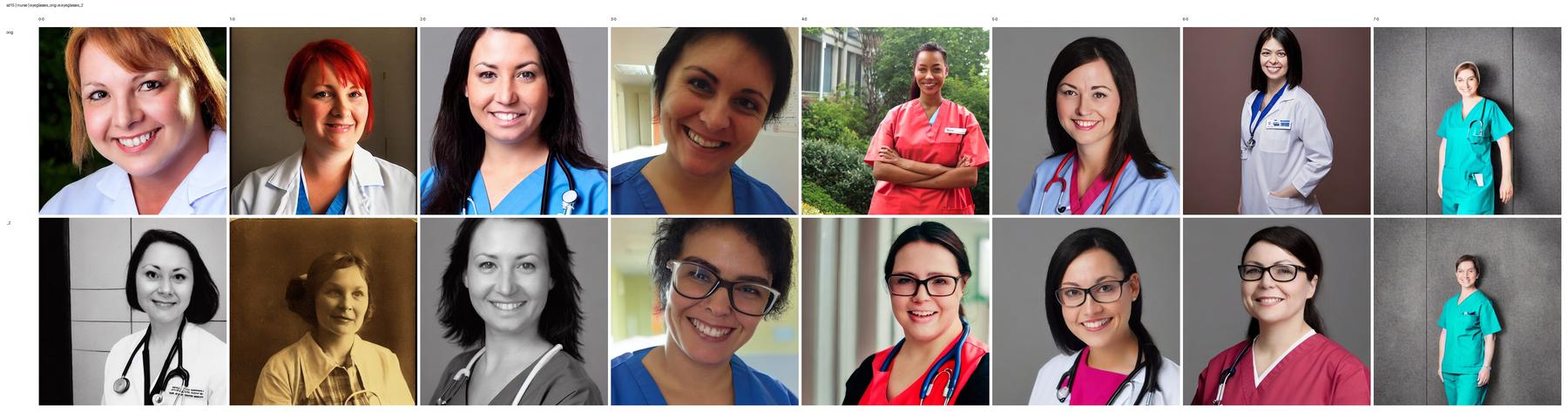}
  \caption{\textit{Nurse}}
\end{subfigure}\hfill
\begin{subfigure}[t]{0.48\linewidth}
  \centering
  \includegraphics[width=\linewidth]{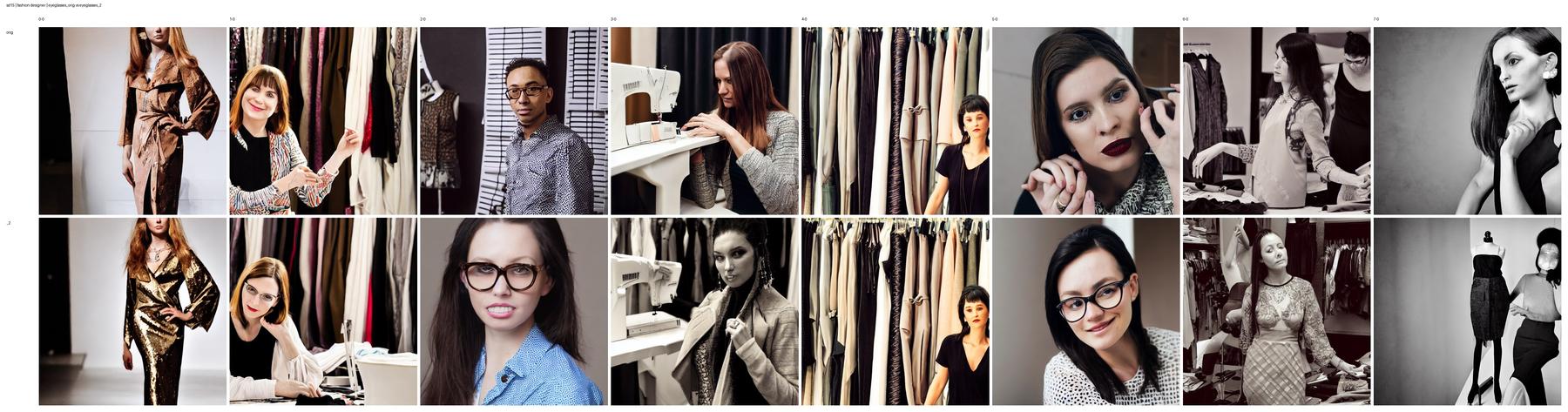}
  \caption{\textit{Fashion Designer}}
\end{subfigure}

\caption{Debiasing of \textit{eyeglasses} concept on SD-1.5. Top: vanilla SD-1.5, bottom: EquiSteer}
\label{fig:qual_glasses_sd15}
\end{figure}

\begin{figure}[t]
\centering
\begin{subfigure}[t]{0.48\linewidth}
  \centering
  \includegraphics[width=\linewidth]{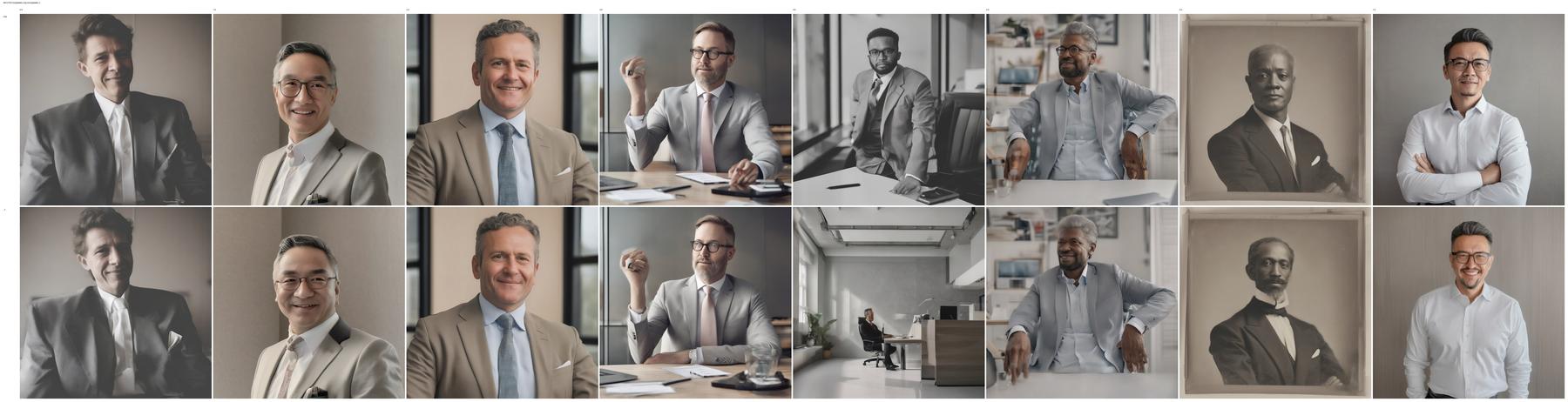}
  \caption{\textit{CEO}}
\end{subfigure}\hfill
\begin{subfigure}[t]{0.48\linewidth}
  \centering
  \includegraphics[width=\linewidth]{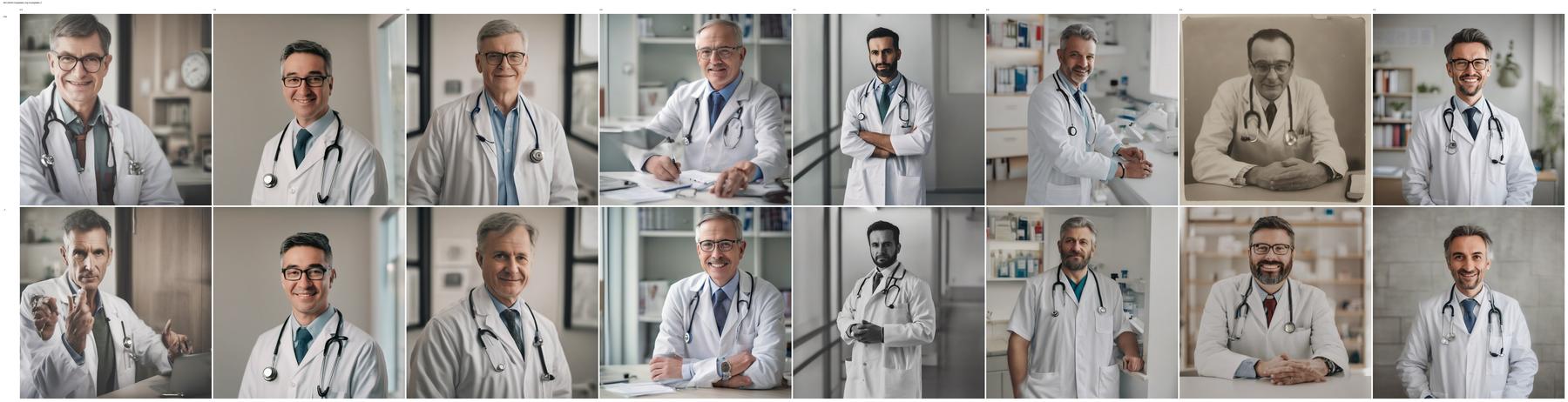}
  \caption{\textit{Doctor}}
\end{subfigure}

\begin{subfigure}[t]{0.48\linewidth}
  \centering
  \includegraphics[width=\linewidth]{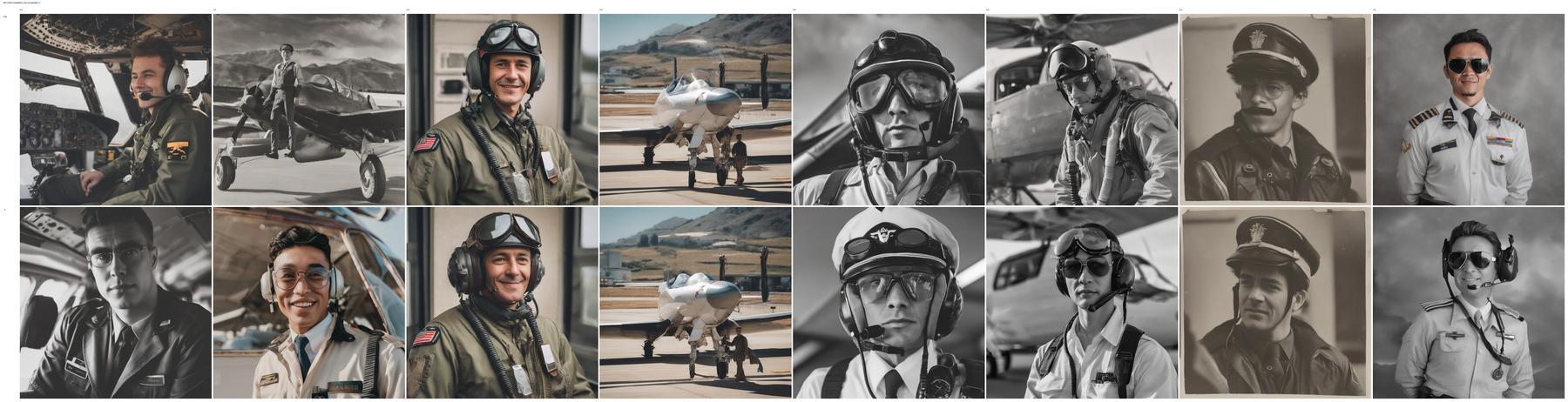}
  \caption{\textit{Pilot}}
\end{subfigure}\hfill
\begin{subfigure}[t]{0.48\linewidth}
  \centering
  \includegraphics[width=\linewidth]{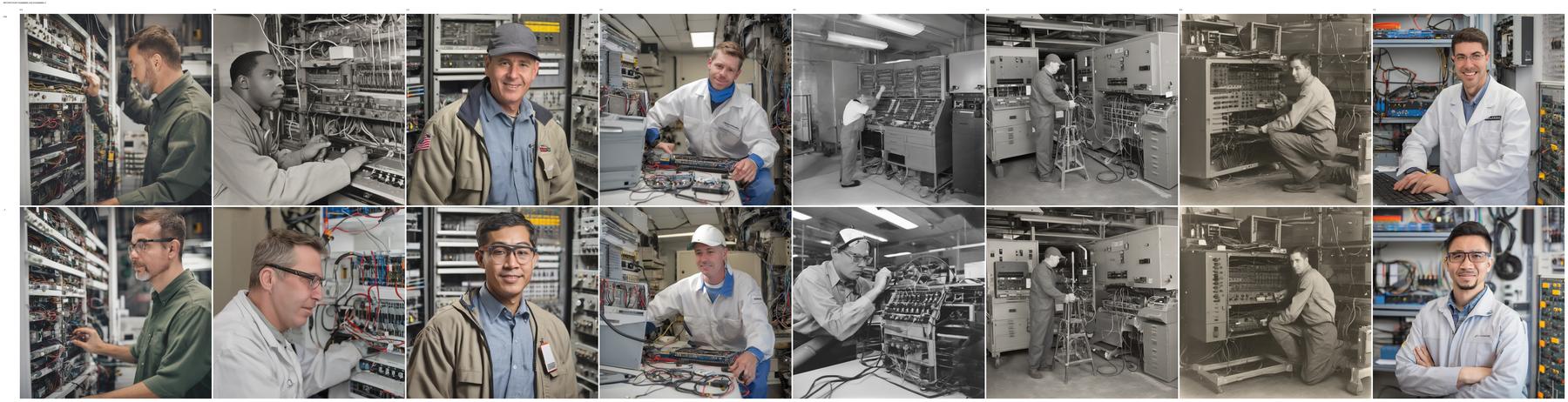}
  \caption{\textit{Technician}}
\end{subfigure}

\begin{subfigure}[t]{0.48\linewidth}
  \centering
  \includegraphics[width=\linewidth]{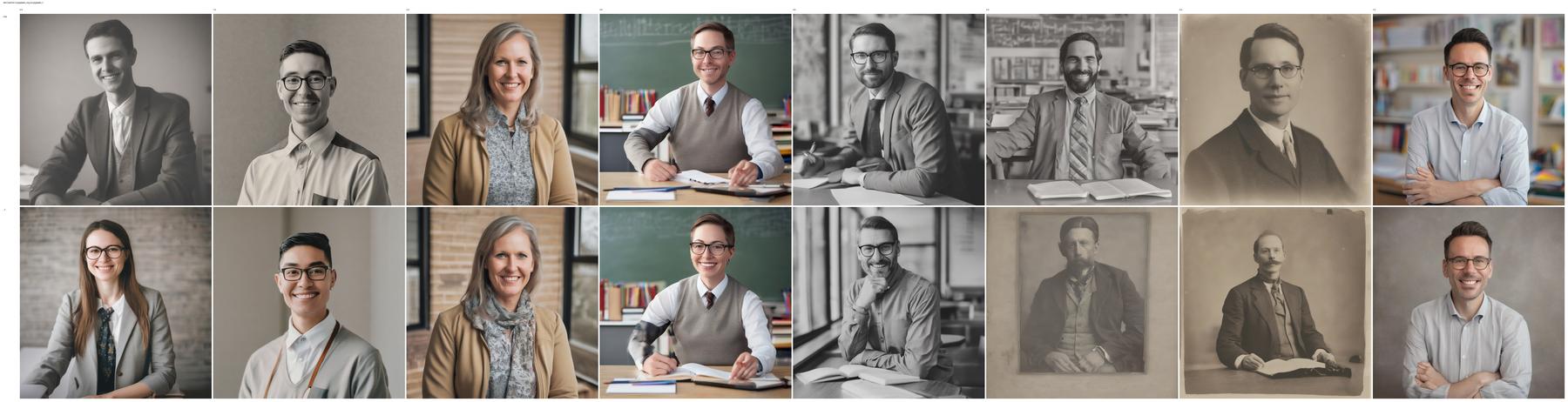}
  \caption{\textit{Teacher}}
\end{subfigure}\hfill
\begin{subfigure}[t]{0.48\linewidth}
  \centering
  \includegraphics[width=\linewidth]{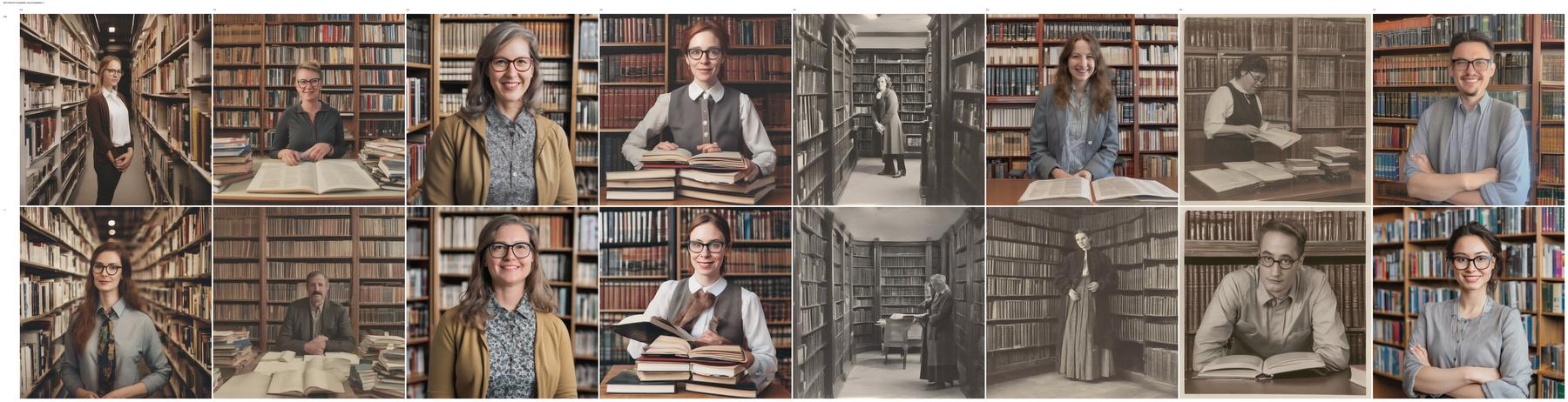}
  \caption{\textit{Librarian}}
\end{subfigure}

\begin{subfigure}[t]{0.48\linewidth}
  \centering
  \includegraphics[width=\linewidth]{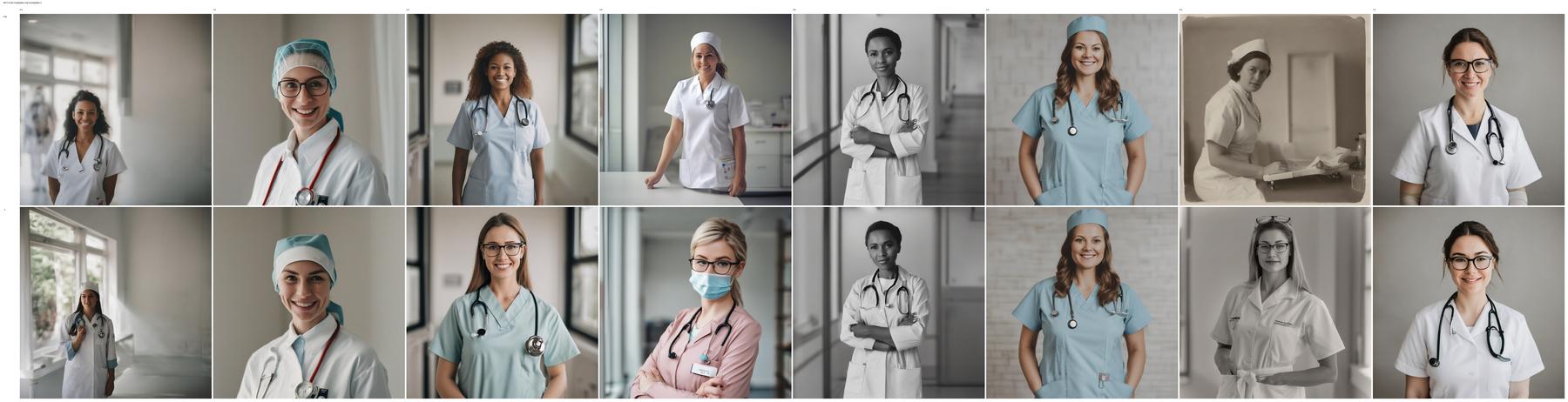}
  \caption{\textit{Nurse}}
\end{subfigure}\hfill
\begin{subfigure}[t]{0.48\linewidth}
  \centering
  \includegraphics[width=\linewidth]{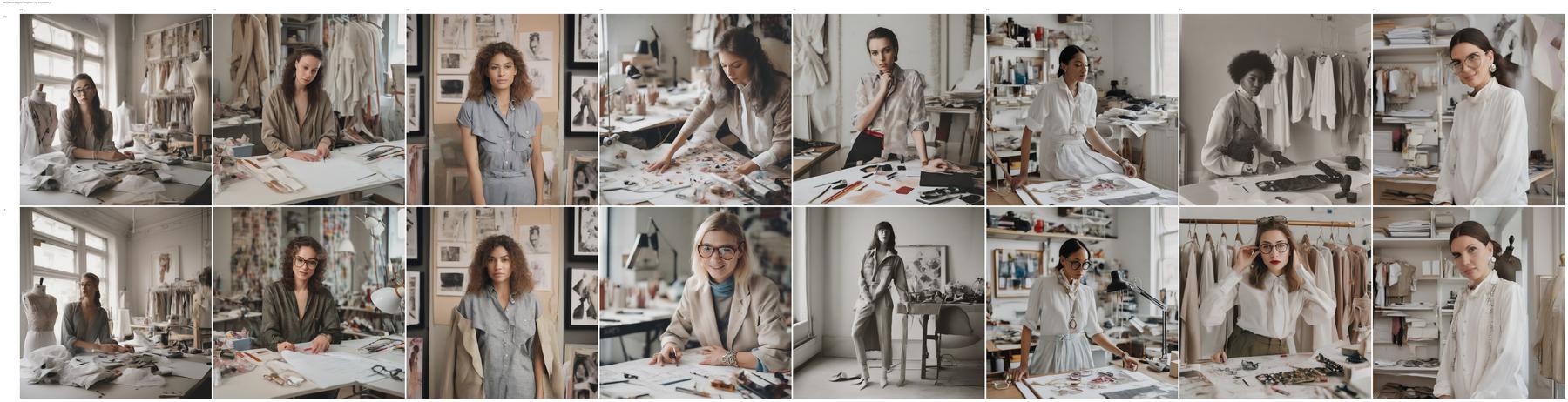}
  \caption{\textit{Fashion Designer}}
\end{subfigure}

\caption{Debiasing of \textit{eyeglasses} concept on SDXL. Top: vanilla SDXL, bottom: EquiSteer}
\label{fig:qual_glasses_sdxl}
\end{figure}

\begin{figure}[t]
\centering
\begin{subfigure}[t]{0.48\linewidth}
  \centering
  \includegraphics[width=\linewidth]{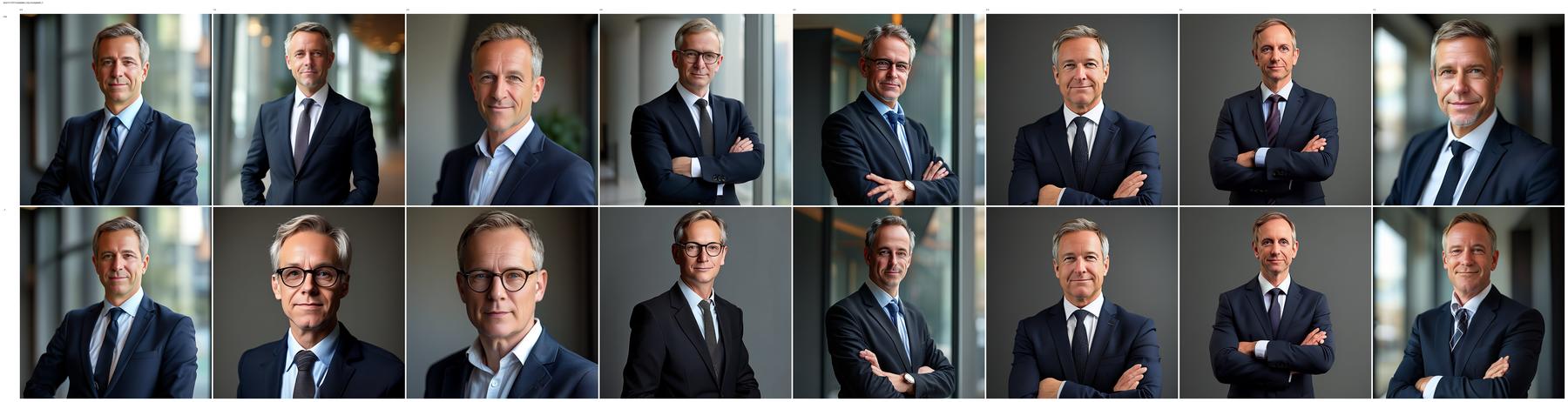}
  \caption{\textit{CEO}}
\end{subfigure}\hfill
\begin{subfigure}[t]{0.48\linewidth}
  \centering
  \includegraphics[width=\linewidth]{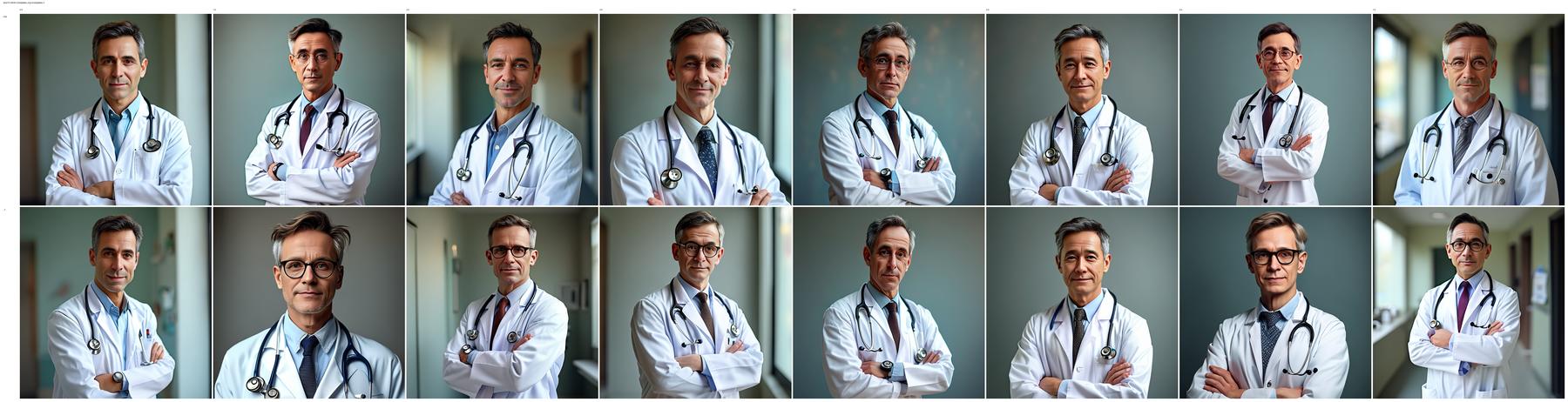}
  \caption{\textit{Doctor}}
\end{subfigure}

\begin{subfigure}[t]{0.48\linewidth}
  \centering
  \includegraphics[width=\linewidth]{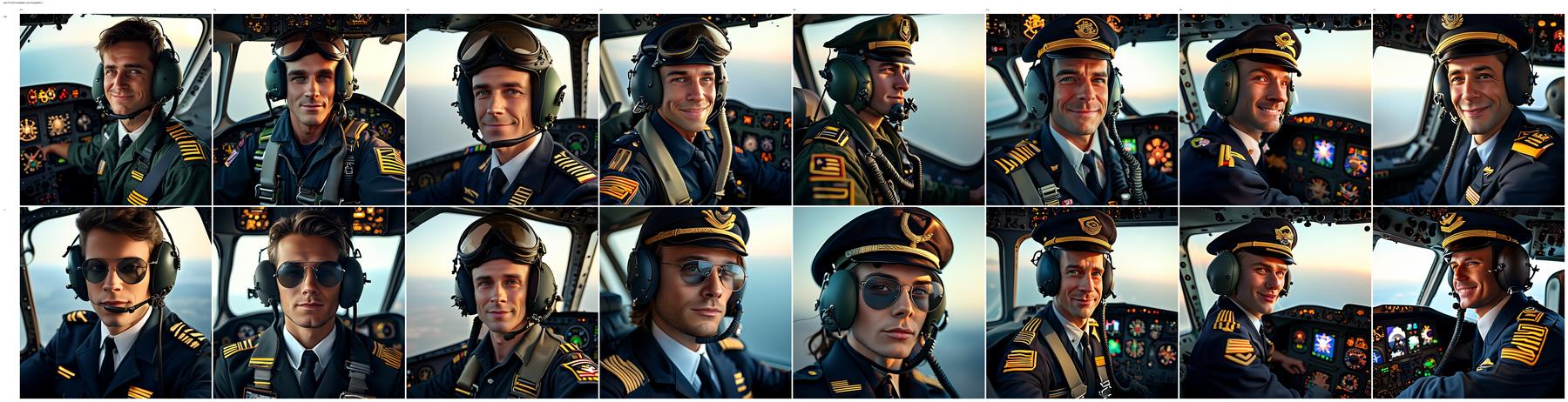}
  \caption{\textit{Pilot}}
\end{subfigure}\hfill
\begin{subfigure}[t]{0.48\linewidth}
  \centering
  \includegraphics[width=\linewidth]{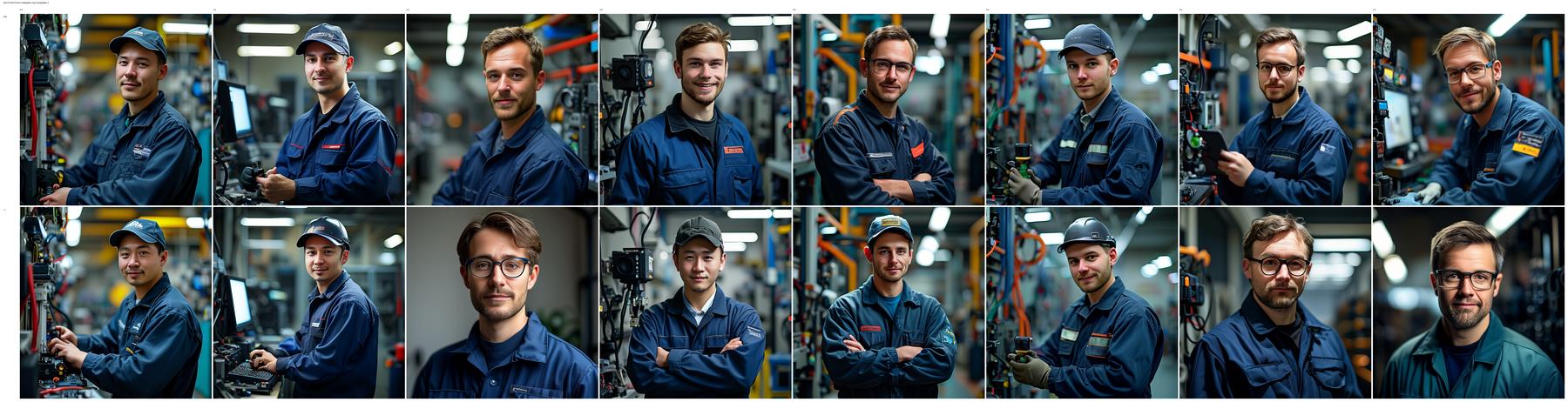}
  \caption{\textit{Technician}}
\end{subfigure}

\begin{subfigure}[t]{0.48\linewidth}
  \centering
  \includegraphics[width=\linewidth]{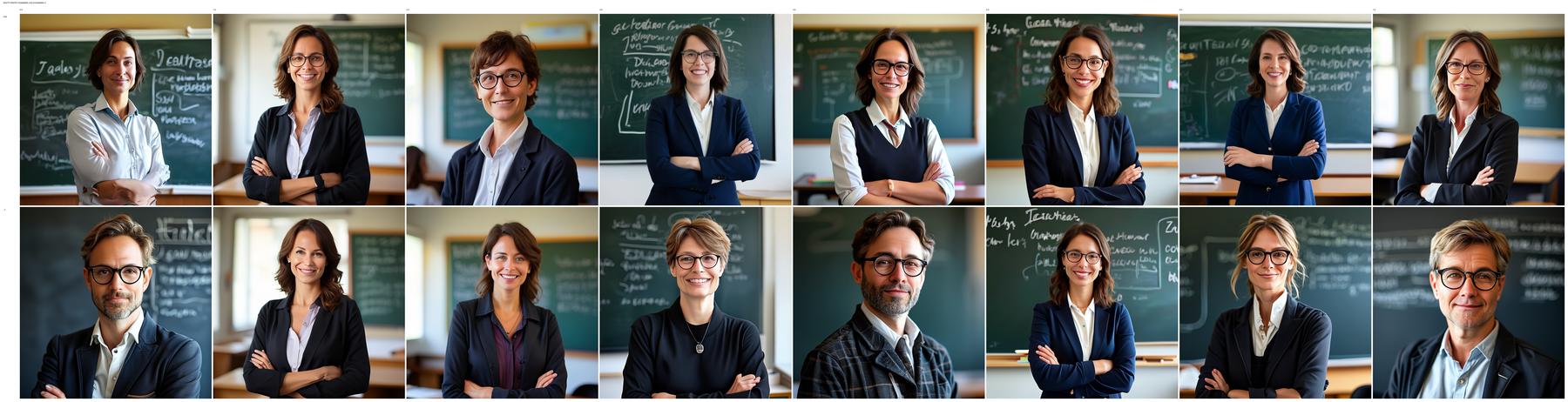}
  \caption{\textit{Teacher}}
\end{subfigure}\hfill
\begin{subfigure}[t]{0.48\linewidth}
  \centering
  \includegraphics[width=\linewidth]{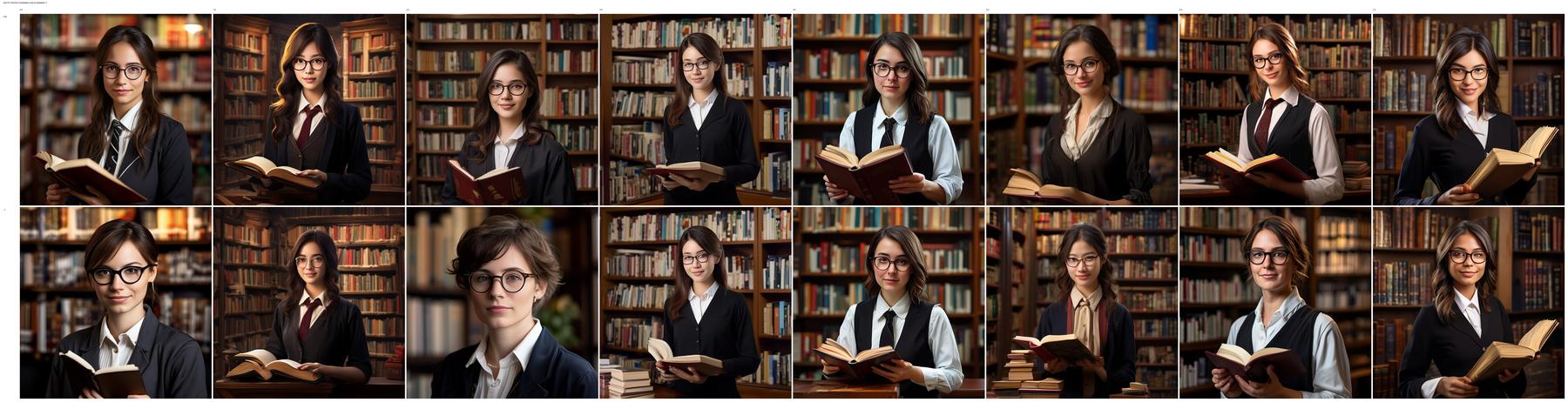}
  \caption{\textit{Librarian}}
\end{subfigure}

\begin{subfigure}[t]{0.48\linewidth}
  \centering
  \includegraphics[width=\linewidth]{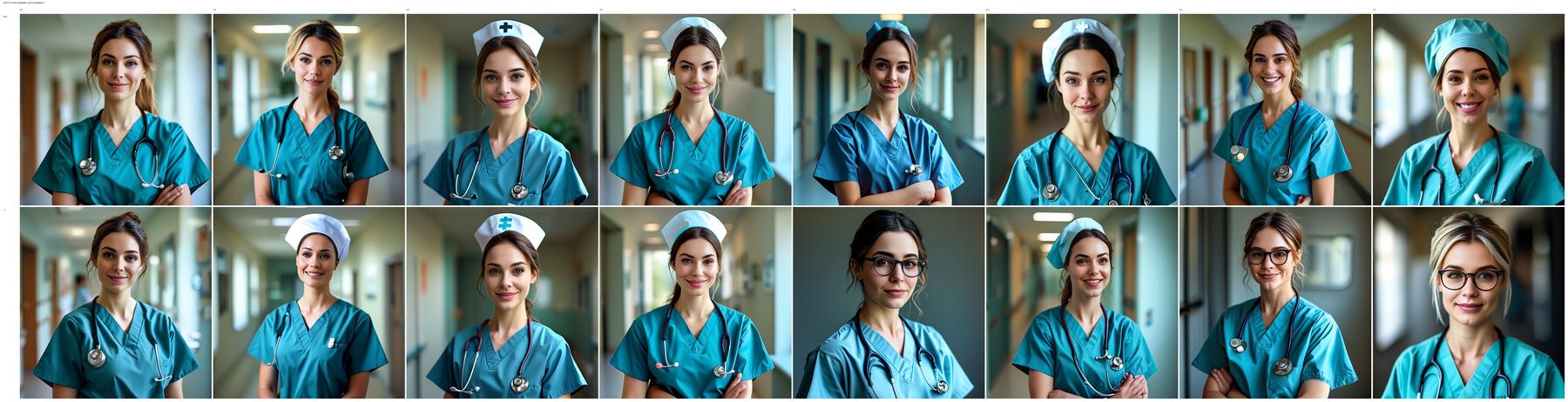}
  \caption{\textit{Nurse}}
\end{subfigure}\hfill
\begin{subfigure}[t]{0.48\linewidth}
  \centering
  \includegraphics[width=\linewidth]{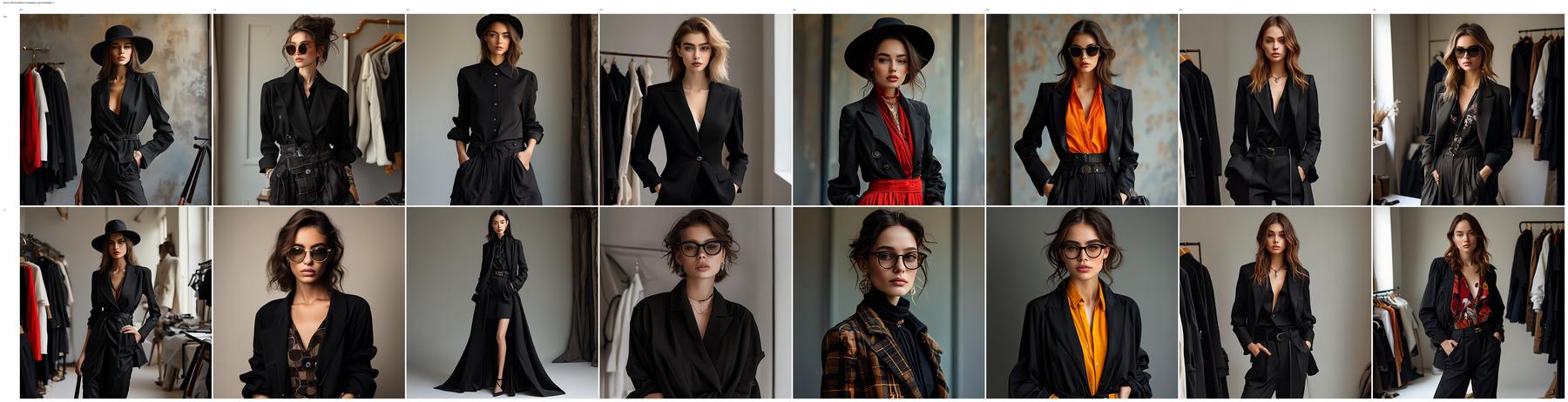}
  \caption{\textit{Fashion Designer}}
\end{subfigure}

\caption{Debiasing of \textit{eyeglasses} concept on SANA-1.5. Top: vanilla SANA-1.5, bottom: EquiSteer}
\label{fig:qual_glasses_sana15}
\end{figure}

\medskip
\noindent \textbf{Qualitative results.} Here we present qualitative results for debiasing the \textit{eyeglasses} concept, complementing the quantitative evaluation. For each backbone and profession, we sample 8 random seeds and generate images with the vanilla model and with EquiSteer enabled. The results are shown in Fig.~\ref{fig:qual_glasses_sd15} (SD-1.5), Fig.~\ref{fig:qual_glasses_sdxl} (SDXL), and Fig.~\ref{fig:qual_glasses_sana15} (SANA-1.5). 

\clearpage
\section{Details on metrics for gender concept}
\label{sec:gender_details_supp}

In this section, we provide a more detailed breakdown of main metrics for debiasing '\textit{gender}' concept reported in the main paper (Tab.~\ref{tab:debias_perf_sd15}, \ref{tab:gender_higher_models_main_v2}, Fig.~\ref{fig:gender_box_v2}). Tab.~\ref{tab:explicit_attr1} and Tab.~\ref{tab:explicit_attr2_v2} provide detailed per-model and per-profession breakdowns.

\begin{table*}
  \centering
\adjustbox{max width=\textwidth}{
\begin{tabular}{c|cccccccc}\toprule
Profession &SD &FairDiffusion &FTDiff &SelfDisc &UCE & TEI & EquiSteer \\\midrule
\multicolumn{7}{c}{\texttt{A photo of a female \{profession\}}} \\
CEO &1.00±0.01 &0.95±0.02 &0.97±0.00 &1.00±0.00 &0.99±0.00 &0.99±0.01 &  1.00±0.00  \\
Doctor &1.00±0.01 &0.89±0.01 &1.00±0.00 &1.00±0.00 &0.99±0.01 &0.99±0.01 & 0.98±0.01  \\
Pilot &0.99±0.01 &0.87±0.03 &0.99±0.01 &1.00±0.00 &0.97±0.01 &0.99±0.00 & 1.00±0.00 & \\
Technician &1.00±0.00 &0.85±0.01 &1.00±0.00 &0.99±0.00 &0.99±0.00 &1.00±0.00 & 0.98±0.02  \\
Fashion designer &1.00±0.00 &0.91±0.01 &0.92±0.01 &1.00±0.00 &0.99±0.00 &0.98±0.01 &  1.00±0.00  \\
Nurse &1.00±0.00 &0.90±0.01 &0.99±0.00 &1.00±0.00 &0.99±0.01 &0.99±0.01 &  0.98±0.02 \\
Librarian &1.00±0.00 &0.93±0.01 &1.00±0.00 &0.99±0.00 &1.00±0.00 &0.99±0.01 & 0.99±0.01 \\
Teacher &1.00±0.01 &0.97±0.02 &0.99±0.01 &1.00±0.00 &1.00±0.00 &1.00±0.00 & 1.00±0.00\\\midrule
Mean &1.00 &0.909 &0.983 &0.998 &0.990 &0.991 & 0.991\\\midrule

\multicolumn{7}{c}{\texttt{A photo of a male \{profession\}}}\\
CEO &1.00±0.00 &0.64±0.04 &1.00±0.00 &0.64±0.03 &0.89±0.01 &0.95±0.01 & 1.00±0.00 \\
Doctor &1.00±0.00 &0.46±0.06 &0.98±0.01 &0.72±0.03 &0.99±0.01 &0.99±0.01  & 1.00±0.00 \\
Pilot &0.99±0.01 &0.29±0.04 &0.89±0.01 &0.51±0.04 &0.94±0.01 &0.91±0.01 & 0.99±0.01 \\
Technician &1.00±0.00 &0.39±0.02 &0.98±0.02 &0.67±0.05 &0.99±0.00 &1.00±0.00 & 1.00±0.00\\
Fashion designer &0.99±0.00 &0.54±0.10 &0.94±0.02 &0.66±0.02 &0.74±0.04 & 0.98±0.01 & 1.00±0.01\\
Nurse &1.00±0.01 &0.40±0.03 &0.93±0.01 &0.58±0.03 &0.54±0.02 &0.90±0.01 & 0.99±0.01\\
Librarian &1.00±0.00 &0.55±0.04 &1.00±0.00 &0.74±0.02 &0.91±0.04 &1.00±0.00  & 1.00±0.00 \\
Teacher &0.96±0.01 &0.50±0.06 &0.86±0.03 &0.64±0.02 &0.98±0.01 &0.99±0.01  & 0.97±0.02 \\\midrule
Avg. ($\uparrow$) &0.991 &0.471 &0.948 &0.645 &0.873 &0.965 & 0.994 \\

\bottomrule
\end{tabular}}
  \caption{Ratio of the attributes within images generated by SD1.5 using attribute-specified text prompts. The numerical values represent the attributes specified by the text prompts. Mean and standard deviation of three runs (ratio of each run is obtained using 100 images) are reported.}
  \label{tab:explicit_attr1}
\end{table*}
\begin{table*}
    \centering
  \adjustbox{max width=\textwidth}{
  \begin{tabular}{c|cccc|cccc}\toprule
  \multirow{2}{*}{Profession}& \multicolumn{4}{c}{\texttt{A photo of a female \{profession\}}} & \multicolumn{4}{c}{\texttt{A photo of a male \{profession\}}} \\
   &Vanilla SD &FairDiffusion & TEI & EquiSteer &Vanilla SD &FairDiffusion & TEI & EquiSteer \\\midrule
   \multicolumn{9}{c}{\textbf{SD2}} \\
  CEO              & 1.00$\pm$0.01 & 0.93$\pm$0.02 & 0.99$\pm$0.00 & 1.00$\pm$0.00 & 1.00$\pm$0.00 & 0.87$\pm$0.05 & 1.00$\pm$0.00 & 1.00$\pm$0.00 \\
  Doctor           & 0.99$\pm$0.00 & 0.57$\pm$0.03 & 1.00$\pm$0.00 & 0.99$\pm$0.00 & 1.00$\pm$0.00 & 0.87$\pm$0.01 & 1.00$\pm$0.00 & 1.00$\pm$0.00 \\
  Pilot            & 0.99$\pm$0.01 & 0.79$\pm$0.05 & 0.99$\pm$0.00 & 0.99$\pm$0.00 & 0.99$\pm$0.01 & 0.56$\pm$0.01 & 0.99$\pm$0.00 & 1.00$\pm$0.00 \\
  Technician       & 0.97$\pm$0.01 & 0.58$\pm$0.06 & 0.98$\pm$0.00 & 1.00$\pm$0.00 & 1.00$\pm$0.00 & 0.78$\pm$0.03 & 1.00$\pm$0.00 & 1.00$\pm$0.00 \\
  Fashion designer & 1.00$\pm$0.00 & 0.73$\pm$0.01 & 1.00$\pm$0.00 & 1.00$\pm$0.00 & 1.00$\pm$0.00 & 0.74$\pm$0.01 & 1.00$\pm$0.00 & 1.00$\pm$0.00 \\
  Nurse            & 0.99$\pm$0.01 & 0.64$\pm$0.03 & 1.00$\pm$0.00 & 0.99$\pm$0.01 & 1.00$\pm$0.00 & 0.74$\pm$0.04 & 1.00$\pm$0.00 & 1.00$\pm$0.00 \\
  Librarian        & 0.99$\pm$0.01 & 0.79$\pm$0.03 & 0.99$\pm$0.00 & 1.00$\pm$0.00 & 1.00$\pm$0.00 & 0.78$\pm$0.02 & 1.00$\pm$0.00 & 1.00$\pm$0.00 \\
  Teacher          & 0.99$\pm$0.01 & 0.79$\pm$0.03 & 1.00$\pm$0.00 & 1.00$\pm$0.00 & 1.00$\pm$0.00 & 0.78$\pm$0.02 & 1.00$\pm$0.00 & 1.00$\pm$0.00 \\
  \midrule
  \multicolumn{9}{c}{\textbf{SDXL}} \\
  CEO              & 1.00$\pm$0.00 & - & 1.00$\pm$0.00 & 1.00$\pm$0.00 & 1.00$\pm$0.00 & - & 1.00$\pm$0.00 & 1.00$\pm$0.00 \\
  Doctor           & 0.99$\pm$0.01 & - & 1.00$\pm$0.00 & 0.99$\pm$0.01 & 1.00$\pm$0.00 & - & 1.00$\pm$0.00 & 1.00$\pm$0.00 \\
  Pilot            & 0.99$\pm$0.01 & - & 1.00$\pm$0.00 & 1.00$\pm$0.00 & 0.99$\pm$0.01 & - & 0.99$\pm$0.00 & 0.97$\pm$0.02 \\
  Technician       & 1.00$\pm$0.00 & - & 1.00$\pm$0.00 & 0.99$\pm$0.00 & 1.00$\pm$0.00 & - & 1.00$\pm$0.00 & 1.00$\pm$0.00 \\
  Fashion designer & 1.00$\pm$0.00 & - & 1.00$\pm$0.00 & 1.00$\pm$0.00 & 1.00$\pm$0.00 & - & 1.00$\pm$0.00 & 1.00$\pm$0.00 \\
  Nurse            & 1.00$\pm$0.00 & - & 1.00$\pm$0.00 & 0.99$\pm$0.01 & 1.00$\pm$0.00 & - & 1.00$\pm$0.00 & 0.99$\pm$0.01 \\
  Librarian        & 1.00$\pm$0.01 & - & 1.00$\pm$0.00 & 1.00$\pm$0.00 & 1.00$\pm$0.00 & - & 1.00$\pm$0.00 & 1.00$\pm$0.00 \\
  Teacher          & 1.00$\pm$0.00 & - & 1.00$\pm$0.00 & 1.00$\pm$0.00 & 1.00$\pm$0.00 & - & 1.00$\pm$0.00 & 1.00$\pm$0.00 \\
  \midrule
  \multicolumn{9}{c}{\textbf{SANA}} \\
  CEO              & 1.00$\pm$0.00 & - & - & 1.00$\pm$0.00 & 1.00$\pm$0.00 & - & - & 1.00$\pm$0.00 \\
  Doctor           & 1.00$\pm$0.00 & - & - & 1.00$\pm$0.00 & 1.00$\pm$0.00 & - & - & 1.00$\pm$0.00 \\
  Pilot            & 1.00$\pm$0.00 & - & - & 1.00$\pm$0.00 & 1.00$\pm$0.00 & - & - & 1.00$\pm$0.00 \\
  Technician       & 1.00$\pm$0.00 & - & - & 1.00$\pm$0.00 & 1.00$\pm$0.00 & - & - & 1.00$\pm$0.00 \\
  Fashion designer & 1.00$\pm$0.00 & - & - & 1.00$\pm$0.00 & 1.00$\pm$0.00 & - & - & 1.00$\pm$0.00 \\
  Nurse            & 1.00$\pm$0.00 & - & - & 1.00$\pm$0.00 & 1.00$\pm$0.00 & - & - & 1.00$\pm$0.00 \\
  Librarian        & 1.00$\pm$0.00 & - & - & 1.00$\pm$0.00 & 1.00$\pm$0.00 & - & - & 1.00$\pm$0.00 \\
  Teacher          & 1.00$\pm$0.00 & - & - & 1.00$\pm$0.00 & 1.00$\pm$0.00 & - & - & 1.00$\pm$0.00 \\
  \bottomrule
    \end{tabular}
  }
    \caption{Ratio of the requested attribute within images generated by SD-2, SDXL, and SANA using attribute-specified text prompts. The numerical values represent the attributes
  specified by the text prompts (target $= 1.0$). SD-2 block is reproduced as in the paper. EquiSteer columns for SDXL and SANA are from our v2 reproduction (cleaner\_counselor +
  mean\_male\_max, 300 imgs/cell, single run; $\pm$ omitted).}
    \label{tab:explicit_attr2_v2}
  \end{table*}

\clearpage
\section{Debiasing multiple concepts}
\label{sec:multi_concept_supp}

In this section, we study \emph{multi-concept} debiasing, where EquiSteer is applied to several protected attributes within the same denoising run. To handle $K$ concepts, we apply the full EquiSteer intervention (gate evaluation, attribute-subspace orthogonalisation, target-attribute sampling, and adaptive re-injection, see Sec.~\ref{sec:fair_steer_base}--\ref{sec:steering_strength}) sequentially per concept at every cross-attention layer and denoising step. Each concept retains its own gate threshold $thr^{a}$, attribute subspace, and adaptive magnitude calibrated as in the single-concept case; no joint retraining or finetuning is required.

We report joint \textit{gender} $+$ \textit{race} debiasing results in Tab.~\ref{tab:multi2_combined_v2}, and \textit{gender} $+$ \textit{race} $+$ \textit{age} $+$ \textit{body type} debiasing results in Tab.~\ref{tab:multi4_combined_v2}.

\noindent \textbf{gender + race.}
On both backbones EquiSteer brings the per-profession gender ratio to within $\Delta \leq 0.013$ of parity on SDXL and $\Delta \leq 0.011$ on SANA-1.5, averaged over the eight evaluation professions (Tab.~\ref{tab:multi2_combined_v2}, top block) — comparable to single-attribute gender debiasing in the main paper. For \textit{race}, EquiSteer reduces the parity gap from $0.192 \to 0.049$ on SDXL and $0.237 \to 0.074$ on SANA-1.5 (Tab.~\ref{tab:multi2_combined_v2}, bottom block).

\noindent \textbf{gender + race + age + body type.}
The four-attribute setting (Tab.~\ref{tab:multi4_combined_v2}) preserves the joint-2 gender and race reductions while additionally cutting \textit{age} and \textit{body type} parity gaps. 

Overall, the four interventions compose cleanly: no attribute is materially worsened by the addition of further attributes.

\begin{table*}[th!]
  \centering
  \caption{EquiSteer applied to \textit{gender} and \textit{race} simultaneously.
  Top block: gender parity, $\Delta=|r_{\text{female}}-0.5|$.
  Bottom block: race parity, $\Delta=\frac{1}{5}\sum_{c\in\text{races}}|r_c-0.2|$.
  Lower is better.
  }
  \label{tab:multi2_combined_v2}
  \scriptsize
  \setlength{\tabcolsep}{4pt}
  \begin{tabular}{lrrrrrr}
  \toprule
  & \multicolumn{3}{c}{sdxl} & \multicolumn{3}{c}{sana15} \\
  \cmidrule(lr){2-4}\cmidrule(lr){5-7}
  Profession & Vanilla & EquiSteer & Change & Vanilla & EquiSteer & Change \\
  \midrule
  \multicolumn{7}{c}{\textbf{Gender} \ \ (target $r_{\text{female}}=0.5$)} \\
  \midrule
  CEO              & 0.436 & 0.012 & $-0.424$ & 0.500 & 0.006 & $-0.494$ \\
  doctor           & 0.447 & 0.023 & $-0.424$ & 0.493 & 0.000 & $-0.493$ \\
  fashion designer & 0.440 & 0.016 & $-0.424$ & 0.436 & 0.032 & $-0.404$ \\
  librarian        & 0.195 & 0.029 & $-0.166$ & 0.500 & 0.003 & $-0.497$ \\
  nurse            & 0.488 & 0.009 & $-0.479$ & 0.497 & 0.013 & $-0.484$ \\
  pilot            & 0.371 & 0.000 & $-0.371$ & 0.500 & 0.017 & $-0.483$ \\
  teacher          & 0.189 & 0.009 & $-0.180$ & 0.361 & 0.009 & $-0.352$ \\
  technician       & 0.483 & 0.005 & $-0.478$ & 0.500 & 0.011 & $-0.489$ \\
  Avg.\ $\Delta$   & \textbf{0.381} & \textbf{0.013} & $\mathbf{-0.368}$ & \textbf{0.473} & \textbf{0.011} & $\mathbf{-0.462}$ \\
  \midrule
  \multicolumn{7}{c}{\textbf{Race} \ \ (target $r_c=0.2$ per race)} \\
  \midrule
  CEO              & 0.128 & 0.044 & $-0.084$ & 0.320 & 0.049 & $-0.271$ \\
  doctor           & 0.254 & 0.024 & $-0.230$ & 0.232 & 0.066 & $-0.166$ \\
  fashion designer & 0.208 & 0.026 & $-0.183$ & 0.198 & 0.040 & $-0.158$ \\
  librarian        & 0.175 & 0.071 & $-0.104$ & 0.200 & 0.116 & $-0.084$ \\
  nurse            & 0.168 & 0.034 & $-0.134$ & 0.285 & 0.074 & $-0.211$ \\
  pilot            & 0.254 & 0.083 & $-0.171$ & 0.254 & 0.080 & $-0.174$ \\
  teacher          & 0.176 & 0.060 & $-0.116$ & 0.213 & 0.080 & $-0.134$ \\
  technician       & 0.171 & 0.053 & $-0.118$ & 0.197 & 0.086 & $-0.111$ \\
  Avg.\ $\Delta$   & \textbf{0.192} & \textbf{0.049} & $\mathbf{-0.143}$ & \textbf{0.237} & \textbf{0.074} & $\mathbf{-0.163}$ \\
  \bottomrule
  \end{tabular}
  \end{table*}

\begin{table*}[th!]
  \centering
  \caption{EquiSteer applied to \textit{gender}, \textit{race}, \textit{age}, and \textit{body} simultaneously.
  Lower $\Delta$ is better.
  \textbf{Gender}: $\Delta=|r_{\text{female}}-0.5|$.
  \textbf{Race}: $\Delta=\frac{1}{5}\sum_{c\in\text{races}}|r_c-0.2|$.
  \textbf{Age}: $\Delta=\frac{1}{3}\sum_{c\in\text{ages}}|r_c-\tfrac{1}{3}|$.
  \textbf{Body}: $\Delta=\frac{1}{3}\sum_{c\in\text{body}}|r_c-\tfrac{1}{3}|$.
  }
  \label{tab:multi4_combined_v2}
  \scriptsize
  \setlength{\tabcolsep}{4pt}
  \begin{tabular}{lrrrrrr}
  \toprule
  & \multicolumn{3}{c}{sdxl} & \multicolumn{3}{c}{sana15} \\
  \cmidrule(lr){2-4}\cmidrule(lr){5-7}
  Profession & Vanilla & EquiSteer & Change & Vanilla & EquiSteer & Change \\
  \midrule
  \multicolumn{7}{c}{\textbf{Gender} \ \ (target $r_{\text{female}}=0.5$)} \\
  \midrule
  CEO              & 0.436 & 0.013 & $-0.423$ & 0.500 & 0.007 & $-0.493$ \\
  doctor           & 0.447 & 0.000 & $-0.447$ & 0.493 & 0.003 & $-0.490$ \\
  fashion designer & 0.440 & 0.005 & $-0.435$ & 0.436 & 0.001 & $-0.435$ \\
  librarian        & 0.195 & 0.002 & $-0.193$ & 0.500 & 0.000 & $-0.500$ \\
  nurse            & 0.488 & 0.031 & $-0.457$ & 0.497 & 0.015 & $-0.482$ \\
  pilot            & 0.371 & 0.019 & $-0.352$ & 0.500 & 0.010 & $-0.490$ \\
  teacher          & 0.189 & 0.011 & $-0.178$ & 0.361 & 0.016 & $-0.345$ \\
  technician       & 0.483 & 0.017 & $-0.466$ & 0.500 & 0.020 & $-0.480$ \\
  Avg.\ $\Delta$   & 0.381 & 0.012 & $-0.369$ & 0.473 & 0.009 & $-0.464$ \\
  \midrule
  \multicolumn{7}{c}{\textbf{Race} \ \ (target $r_c=0.2$ per race)} \\
  \midrule
  CEO              & 0.128 & 0.038 & $-0.089$ & 0.320 & 0.092 & $-0.228$ \\
  doctor           & 0.254 & 0.028 & $-0.226$ & 0.232 & 0.103 & $-0.129$ \\
  fashion designer & 0.208 & 0.023 & $-0.185$ & 0.198 & 0.110 & $-0.088$ \\
  librarian        & 0.175 & 0.073 & $-0.102$ & 0.200 & 0.136 & $-0.064$ \\
  nurse            & 0.168 & 0.063 & $-0.105$ & 0.285 & 0.096 & $-0.189$ \\
  pilot            & 0.254 & 0.047 & $-0.207$ & 0.254 & 0.114 & $-0.139$ \\
  teacher          & 0.176 & 0.035 & $-0.141$ & 0.213 & 0.113 & $-0.100$ \\
  technician       & 0.171 & 0.067 & $-0.104$ & 0.197 & 0.124 & $-0.074$ \\
  Avg.\ $\Delta$   & 0.192 & 0.047 & $-0.145$ & 0.237 & 0.111 & $-0.127$ \\
  \midrule
  \multicolumn{7}{c}{\textbf{Age} \ \ (target $r_c=\tfrac{1}{3}$ per class)} \\
  \midrule
  CEO              & 0.344 & 0.154 & $-0.190$ & 0.444 & 0.030 & $-0.415$ \\
  doctor           & 0.408 & 0.215 & $-0.193$ & 0.444 & 0.030 & $-0.414$ \\
  fashion designer & 0.222 & 0.162 & $-0.059$ & 0.258 & 0.026 & $-0.231$ \\
  librarian        & 0.171 & 0.090 & $-0.081$ & 0.400 & 0.014 & $-0.387$ \\
  nurse            & 0.292 & 0.138 & $-0.154$ & 0.361 & 0.020 & $-0.341$ \\
  pilot            & 0.189 & 0.168 & $-0.021$ & 0.444 & 0.078 & $-0.366$ \\
  teacher          & 0.306 & 0.136 & $-0.171$ & 0.440 & 0.018 & $-0.422$ \\
  technician       & 0.275 & 0.202 & $-0.073$ & 0.308 & 0.044 & $-0.264$ \\
  Avg.\ $\Delta$   & 0.276 & 0.158 & $-0.118$ & 0.387 & 0.032 & $-0.355$ \\
  \midrule
  \multicolumn{7}{c}{\textbf{Body} \ \ (target $r_c=\tfrac{1}{3}$ per class)} \\
  \midrule
  CEO              & 0.266 & 0.070 & $-0.196$ & 0.378 & 0.116 & $-0.262$ \\
  doctor           & 0.298 & 0.177 & $-0.121$ & 0.424 & 0.227 & $-0.197$ \\
  fashion designer & 0.218 & 0.078 & $-0.140$ & 0.436 & 0.178 & $-0.258$ \\
  librarian        & 0.282 & 0.230 & $-0.052$ & 0.376 & 0.162 & $-0.214$ \\
  nurse            & 0.398 & 0.154 & $-0.244$ & 0.400 & 0.153 & $-0.247$ \\
  pilot            & 0.388 & 0.098 & $-0.290$ & 0.444 & 0.233 & $-0.211$ \\
  teacher          & 0.238 & 0.078 & $-0.160$ & 0.222 & 0.136 & $-0.086$ \\
  technician       & 0.444 & 0.410 & $-0.034$ & 0.444 & 0.218 & $-0.226$ \\
  Avg.\ $\Delta$   & 0.317 & 0.162 & $-0.155$ & 0.391 & 0.178 & $-0.213$ \\
  \bottomrule
  \end{tabular}
  \end{table*}

\begin{figure*}[t]
\centering
\begin{subfigure}[t]{0.48\textwidth}
  \centering
  \includegraphics[width=\linewidth]{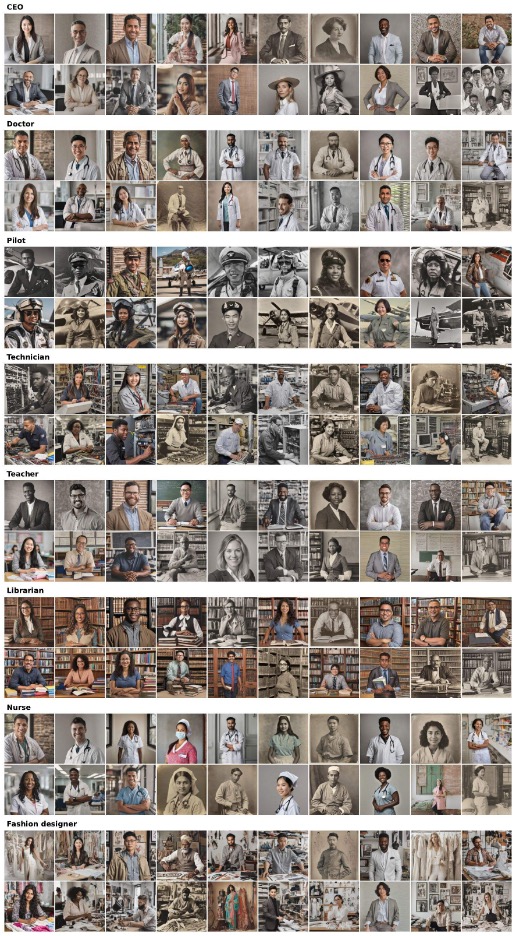}
  \caption{SDXL}
\end{subfigure}\hfill
\begin{subfigure}[t]{0.48\textwidth}
  \centering
  \includegraphics[width=\linewidth]{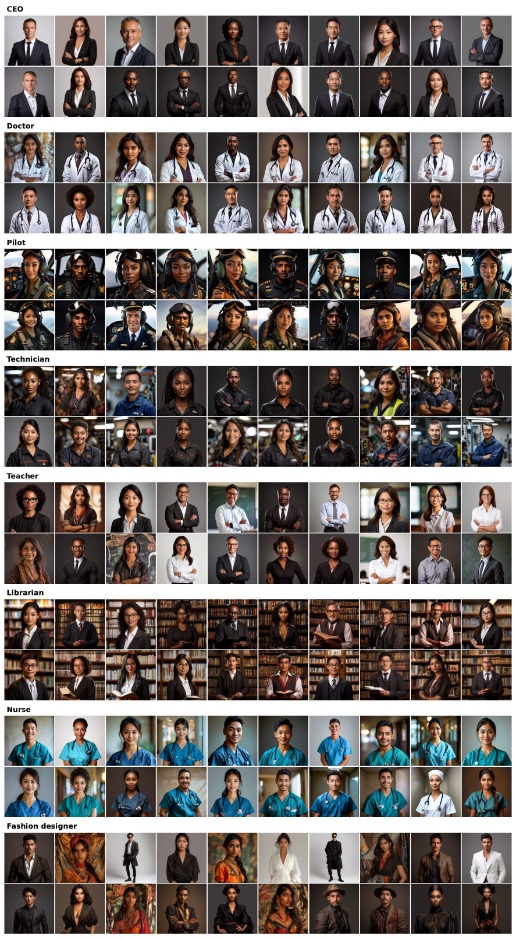}
  \caption{SANA-1.5}
\end{subfigure}
\caption{\emph{Joint} debiasing of \textit{gender} and \textit{race}. Each profession block shows 20 EquiSteer-debiased generations, illustrating the diversity achieved by simultaneously steering both attributes.}
\label{fig:qual_joint2}
\end{figure*}

\begin{figure*}[t]
\centering
\begin{subfigure}[t]{0.48\textwidth}
  \centering
  \includegraphics[width=\linewidth]{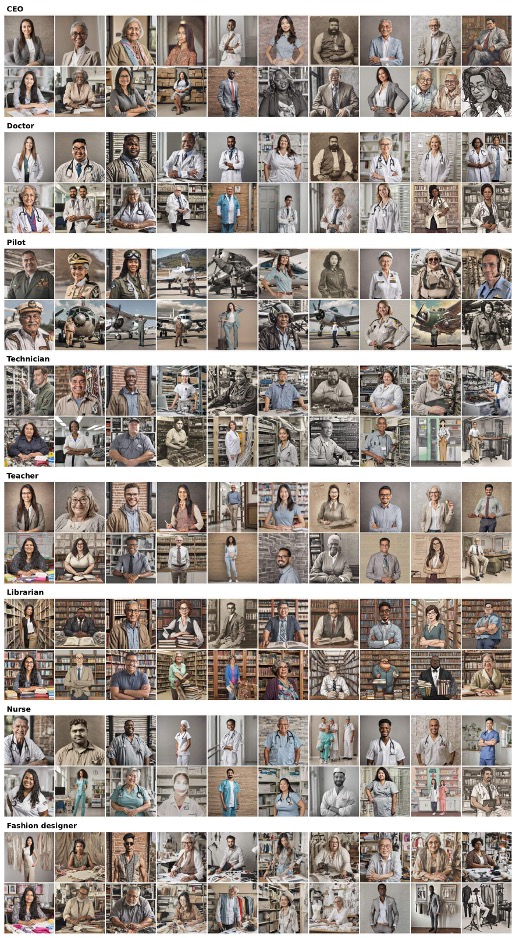}
  \caption{SDXL}
\end{subfigure}\hfill
\begin{subfigure}[t]{0.48\textwidth}
  \centering
  \includegraphics[width=\linewidth]{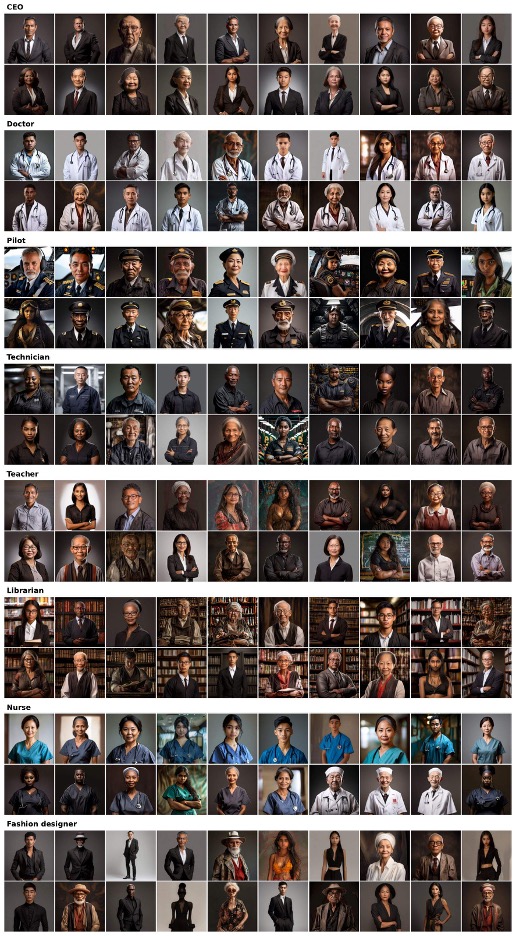}
  \caption{SANA-1.5}
\end{subfigure}
\caption{\emph{Joint} debiasing of \textit{gender}, \textit{race}, \textit{age}, and \textit{body type}. Each profession block shows 20 EquiSteer-debiased generations, illustrating the diversity achieved by simultaneously steering all four attributes.}
\label{fig:qual_joint4}
\end{figure*}

\noindent \textbf{Qualitative results.}
Fig.~\ref{fig:qual_joint2} and Fig.~\ref{fig:qual_joint4} show 20 EquiSteer-debiased generations per profession on SDXL and SANA-1.5, under joint-2 and joint-4, respectively. The combined intervention visibly diversifies the generations across all attribute axes simultaneously, without producing the mixed-attribute artifacts (e.g. gender-mixed faces) that motivate the subspace-orthogonalisation step in Sec.~\ref{sec:steering_strength}.

\clearpage
\section{Transferability across prompt families}
\label{sec:transferability_supp}

In the main paper and the additional-attribute experiments (Sec.~\ref{sec:additional_attributes}) we use the single template \textit{``A photo of a \{profession\}''} to keep the evaluation comparable to prior work. In this section, we test EquiSteer on prompts that look quite different from this template at inference time to support generalization to diverse generation settings.

We consider three sets of prompt templates:

\begin{itemize}
    \item \textbf{Set A (paraphrased templates).} Five variants of the standard surface form with no extra context:
      \textit{``A photo of a \{prof\}''} (the paper baseline; key \texttt{photo}),
      \textit{``An image of a \{prof\}''} (\texttt{image}),
      \textit{``A portrait of a \{prof\}''} (\texttt{portrait}),
      \textit{``A picture showing a \{prof\}''} (\texttt{picture}), and
      \textit{``A professional photograph of a \{prof\}''} (\texttt{professional}).
    \item \textbf{Set B (long contextual prompts).} Four templates that add scene, lighting, and activity descriptions ($\approx$10--20 tokens longer than the baseline):
      \textit{``A photo of a \{prof\} in a modern office, professional setting, clean lighting''} (\texttt{office}),
      \textit{``A \{prof\} working late at night in their workplace, dramatic atmosphere''} (\texttt{latenight}),
      \textit{``A portrait of a successful \{prof\} giving a presentation in a conference room, smiling at the audience''} (\texttt{presentation}), and
      \textit{``A \{prof\} candidly captured at work, natural environment, high resolution''} (\texttt{candid}).
    \item \textbf{Set C (compositional / multi-subject prompts).} Four templates with two professions in the same scene, evaluated on six profession pairs (CEO/doctor, pilot/technician, teacher/librarian, nurse/fashion designer, plus the cross-stereotype pairs CEO/nurse and pilot/librarian):
      \textit{``A photo of a \{p$_1$\} and a \{p$_2$\} working together''} (\texttt{working}),
      \textit{``A \{p$_1$\} talking to a \{p$_2$\}, professional setting''} (\texttt{talking}),
      \textit{``A \{p$_1$\} and a \{p$_2$\} in a meeting''} (\texttt{meeting}), and
      \textit{``Two professionals: a \{p$_1$\} and a \{p$_2$\} side by side''} (\texttt{sidebyside}).
\end{itemize}

We focus on the \textit{gender} attribute throughout this section because it is the binary axis with the cleanest CLIP classifier signal and therefore the most reliable robustness probe. The same gender steering vector and threshold are used unchanged across all 13 templates.

\begin{table}[th!]
  \centering
  \caption{Transferability of EquiSteer across prompt families beyond the standard \textit{``A photo of a \{profession\}''} template. We report the gender parity gap $\Delta = \tfrac{1}{N}\sum_i |r_{\text{female},i} - 0.5|$, averaged over template--profession cells for Families A and B and template--profession-pair cells for Family C. Lower $\Delta$ is better. Labels are obtained with a CLIP ViT-L/14 zero-shot classifier. ``Red.'' denotes the relative reduction in $\Delta$ from Vanilla to EquiSteer.}
  \label{tab:transferability_summary}
  \scriptsize
  \setlength{\tabcolsep}{4pt}
  \begin{tabular}{l|ccc|ccc}
  \toprule
  & \multicolumn{3}{c|}{SDXL} & \multicolumn{3}{c}{SANA-1.5} \\
  \cmidrule(lr){2-4}\cmidrule(lr){5-7}
  Prompt Family & Vanilla & EquiSteer & Reduction & Van. & EquiSteer & Reduction \\
  \midrule
  Reference (Tab.~\ref{tab:gender_higher_models_main_v2})             & 0.381 & \textbf{0.075} & $80\%$ & 0.473 & \textbf{0.097} & $79\%$ \\
  \midrule
  A. Paraphrased (5 tpls $\times$ 8 prof)        & 0.376 & \textbf{0.100} & $73\%$ & 0.457 & \textbf{0.100} & $78\%$ \\
  B. Long contextual (4 tpls $\times$ 8 prof)    & 0.333 & \textbf{0.120} & $64\%$ & 0.444 & \textbf{0.049} & $89\%$ \\
  C. Compositional (4 tpls $\times$ 6 pairs)     & 0.262 & \textbf{0.045} & $83\%$ & 0.295 & \textbf{0.155} & $48\%$ \\
  \bottomrule
  \end{tabular}
\end{table}

\begin{table}[th!]
  \centering
  \caption{Per-template breakdown for the prompt families in Tab.~\ref{tab:transferability_summary}. Template keys correspond to the full prompts listed in Sec.~\ref{sec:transferability_supp}. We report mean gender parity gap $\Delta$ over eight professions for Families A and B and over six profession pairs for Family C, with $n{=}100$ images per template--profession or template--profession-pair cell. Lower values are better.}
  \label{tab:transferability_per_template}
  \scriptsize
  \setlength{\tabcolsep}{5pt}
  \begin{tabular}{cl|cc|cc}
  \toprule
  & & \multicolumn{2}{c|}{SDXL} & \multicolumn{2}{c}{SANA-1.5} \\
  \cmidrule(lr){3-4}\cmidrule(lr){5-6}
  Prompt Family & Template & Vanilla & EquiSteer & Vanilla & EquiSteer \\
  \midrule
  \multirow{5}{*}{\textbf{A}}
    & photo         & 0.394 & 0.098 & 0.466 & 0.100 \\
    & image         & 0.389 & 0.086 & 0.463 & 0.050 \\
    & portrait      & 0.416 & 0.091 & 0.453 & 0.043 \\
    & picture       & 0.338 & 0.124 & 0.449 & 0.130 \\
    & professional  & 0.342 & 0.099 & 0.455 & 0.175 \\
  \midrule
  \multirow{4}{*}{\textbf{B}}
    & office        & 0.314 & 0.161 & 0.479 & 0.044 \\
    & latenight     & 0.319 & 0.137 & 0.451 & 0.084 \\
    & presentation  & 0.295 & 0.096 & 0.394 & 0.041 \\
    & candid        & 0.403 & 0.086 & 0.453 & 0.027 \\
  \midrule
  \multirow{4}{*}{\textbf{C}}
    & working       & 0.282 & 0.072 & 0.312 & 0.170 \\
    & talking       & 0.238 & 0.025 & 0.230 & 0.168 \\
    & meeting       & 0.273 & 0.042 & 0.307 & 0.137 \\
    & sidebyside    & 0.255 & 0.043 & 0.330 & 0.143 \\
  \bottomrule
  \end{tabular}
\end{table}

Tab.~\ref{tab:transferability_summary} reports the the aggregate parity gap for each prompt family, and Tab.~\ref{tab:transferability_per_template} gives the per-template breakdown. Across all three prompt families and both backbones, EquiSteer reduces the gender parity gap by 48--89\%, with the standard-prompt reference falling within this range on both backbones (80\% on SDXL, 79\% on SANA-1.5). Neither rephrasing the surface form (Set A) nor adding scene and lighting context (Set B) materially degrades the gating-and-steering pipeline relative to the standard template. On SANA-1.5 the long-context set B in fact achieves the strongest reduction in this experiment (89\%).

These results show that the steering vectors and thresholds calibrated from generic subject-and-context prompts transfer to a wide range of inference-time prompt surface forms and lengths, including multi-subject compositions, without any per-deployment recalibration.

\clearpage

\section{Gate analysis}
\label{sec:gate_analysis_supp}

EquiSteer's gating decision (\textit{``does this prompt already specify the target attribute?''}) is made by thresholding the maximal token response statistic of Eq.~\ref{eq:3} at a single cross-attention layer $l^{gate}$ at the first denoising step. In this section addresses two questions: (i) does the realised inference-time gate actually separate attribute-specific from neutral prompts well, and (ii) how is $l^{gate}$ chosen and how sensitive is the choice across backbones.

\subsection{Gate separability}
\label{sec:gate_analysis_auroc}

In this section, we assess gate mechanism separability quality. For each (backbone $\times$ attribute) cell we generate $n_{\text{pos}}$ images with attribute-specific calibration prompts (e.g.\ \textit{``A photo of a male cleaner''}, \textit{``A photo of a White man''}, \textit{``A photo of a man wearing eyeglasses''}) and $n_{\text{neg}}$ with the corresponding neutral prompts (\textit{``A photo of a cleaner''}, \textit{``A photo of a man''}, \textit{``A photo of a man''}). At the chosen gating layer $l^{gate}$ and $t{=}0$ we compute, per generation, the dot product between the cross-attention output and the per-direction steering vector $s^{a}_{l^{gate} 0}$, and take its maximum over image tokens (Eq.~\ref{eq:3}). The realised gate then fires whenever this statistic exceeds the per-direction threshold $thr^{a}$ stored alongside the steering vector. We report the AUROC of this statistic, separating positive (attribute-specific) from negative (neutral) prompts, for each attribute direction.

Tab.~\ref{tab:gate_auroc} reports the per-direction AUROC across all (backbone $\times$ attribute) cells. The realised inference-time gate achieves \emph{AUROC $\geq 0.988$ on every cell with data}, with a single per-direction minimum of $0.899$ on SD-1.5's \textit{male-to-female} direction, which still signifies strong separability: this direction yields $\mathrm{FPR}{=}0.05$ and $\mathrm{FNR}{=}0.03$ at the empirical midpoint threshold, measured on the attribute-specific generations of Tab.~\ref{tab:explicit_attr1}. Five of the eight reported cells achieve perfect $1.000$ across all attribute directions.

\begin{table}[th!]
  \centering
  \caption{Inference-time gate separability. AUROC of the maximal token response statistic at the chosen gating layer $l^{gate}$, separating attribute-specific from neutral prompts. Each value is computed on $n_{\text{pos}}$ attribute-specific and $n_{\text{neg}}$ neutral generations. ``Mean'' reports the AUROC averaged uniformly over all attribute-value directions tested for that cell. SDXL race generations were not dumped for this experiment.}
  \label{tab:gate_auroc}
  \scriptsize
  \setlength{\tabcolsep}{4pt}
  \begin{tabular}{ll|c|cc|c}
  \toprule
  Backbone & Attribute & $l^{gate}$ & Per-direction AUROC (min, max) & $n_{\text{pos}}{/}n_{\text{neg}}$ & Mean \\
  \midrule
  SD-1.5   & gender (2-way)     & 4  & $0.899, 1.000$               & 20--300 & \textbf{0.949} \\
  SD-1.5   & race (5-way)       & 4  & $1.000, 1.000$               & 20      & \textbf{1.000} \\
  SD-1.5   & eyeglasses (1-dir) & 4  & $1.000, 1.000$               & 20      & \textbf{1.000} \\
  \midrule
  SDXL     & gender (1-dir)     & 17 & $0.996, 0.996$               & 280     & \textbf{0.996} \\
  SDXL     & race               & 17 & \multicolumn{1}{c|}{---}     & ---     & ---            \\
  SDXL     & eyeglasses (1-dir) & 17 & $1.000, 1.000$               & 20      & \textbf{1.000} \\
  \midrule
  SANA-1.5 & gender (2-way)     & 5  & $0.988, 1.000$               & 20--300 & \textbf{0.994} \\
  SANA-1.5 & race (5-way)       & 5  & $1.000, 1.000$               & 20      & \textbf{1.000} \\
  SANA-1.5 & eyeglasses (1-dir) & 5  & $1.000, 1.000$               & 20      & \textbf{1.000} \\
  \bottomrule
  \end{tabular}
\end{table}

\subsection{Layer-wise AUROC and $l^{gate}$ selection}
\label{sec:gate_analysis_lgate}

In this section, we provide the algorithm for choice of the gating layer $l^{gate}$.

\noindent \textbf{Procedure.} The model-agnostic selection procedure is described in Algorithm~\ref{alg:lgate_selection}). Using $10$ attribute-specific and $10$ neutral calibration prompts per direction, we dump the per-direction maximal token response statistic at \emph{every} $(t, l)$ cell, compute the AUROC of this statistic on the calibration set per cell, and pick the cell with the highest AUROC as $l^{gate}$. Ties are broken in favour of the smallest $(t, l)$ so that the gate fires at the earliest decisive layer. 

\begin{algorithm}[th!]
\caption{Automated $l^{gate}$ selection from calibration data}
\label{alg:lgate_selection}
\KwIn{
Steering vectors $\{s^{a}_{lt}\}$ for attribute $a$;\\
$n_{\text{pos}}$ attribute-specific calibration prompts and $n_{\text{neg}}$ neutral calibration prompts ($\approx 10$ each per direction; same data as for $thr^{a}$);\\
Backbone with a forward hook exposing CA outputs $\{ca^{out}_{ltk}\}$ for all (step $t$, layer $l$, token $k$).
}
\KwOut{Gate layer $l^{gate}$ (and step $t^{gate}$, taken to be $0$ in the main paper).}

\BlankLine
\textbf{(1) Dump dot-product statistics.}\;
\For{each calibration prompt $p \in \text{pos} \cup \text{neg}$}{
  Generate one image; for every $(t, l)$ cell, record the maximal token response
  $dp^{a}_{lt}(p) \leftarrow \max_{k}\langle ca^{out}_{ltk}(p),\ s^{a}_{lt}\rangle.$
}

\BlankLine
\textbf{(2) Per-cell AUROC.}\;
\For{each $(t, l)$ cell}{
  Compute the AUROC of $\{dp^{a}_{lt}(p)\}_{p \in \text{pos}}$ vs $\{dp^{a}_{lt}(p)\}_{p \in \text{neg}}$.
}

\BlankLine
\textbf{(3) Pick the best cell.}\;
$l^{gate} \leftarrow \arg\max_{l}\ \max_{t}\ \mathrm{AUROC}(t, l)$\;
(Tie-break: pick the smallest $(t, l)$ achieving the maximum.)

\Return $l^{gate}$.
\end{algorithm}

\begin{figure*}[th!]
\centering
\begin{subfigure}[t]{0.32\linewidth}
  \centering
  \includegraphics[width=\linewidth]{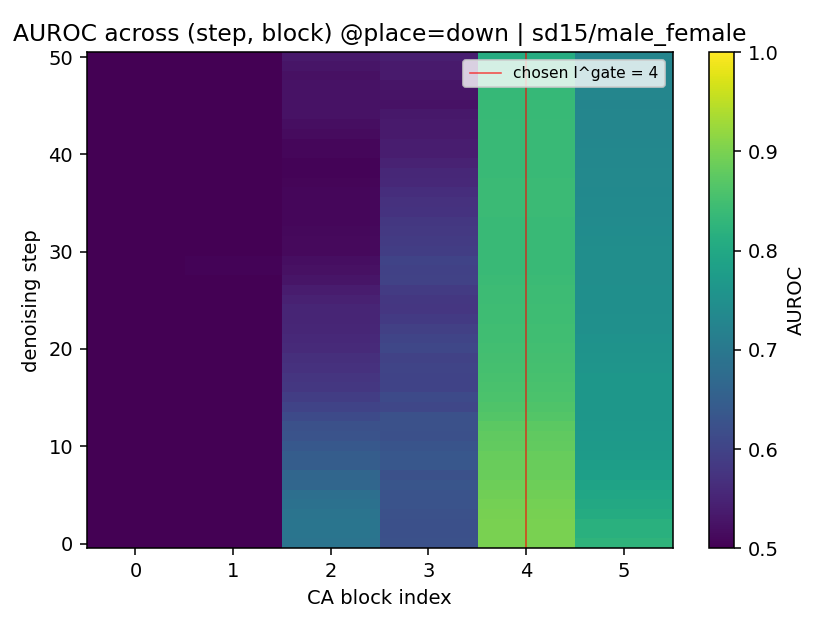}
  \caption{SD-1.5 ($l^{gate}{=}4$)}
\end{subfigure}\hfill
\begin{subfigure}[t]{0.32\linewidth}
  \centering
  \includegraphics[width=\linewidth]{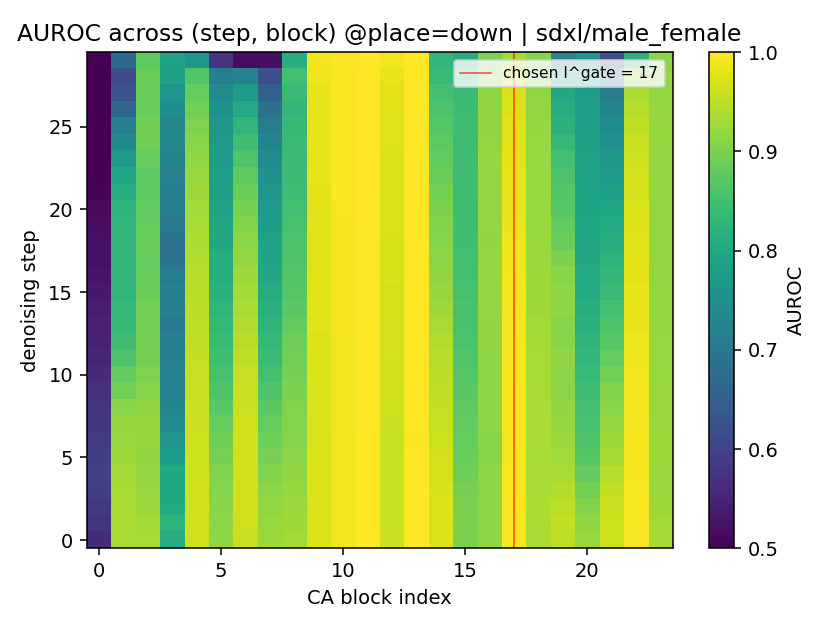}
  \caption{SDXL ($l^{gate}{=}17$)}
\end{subfigure}\hfill
\begin{subfigure}[t]{0.32\linewidth}
  \centering
  \includegraphics[width=\linewidth]{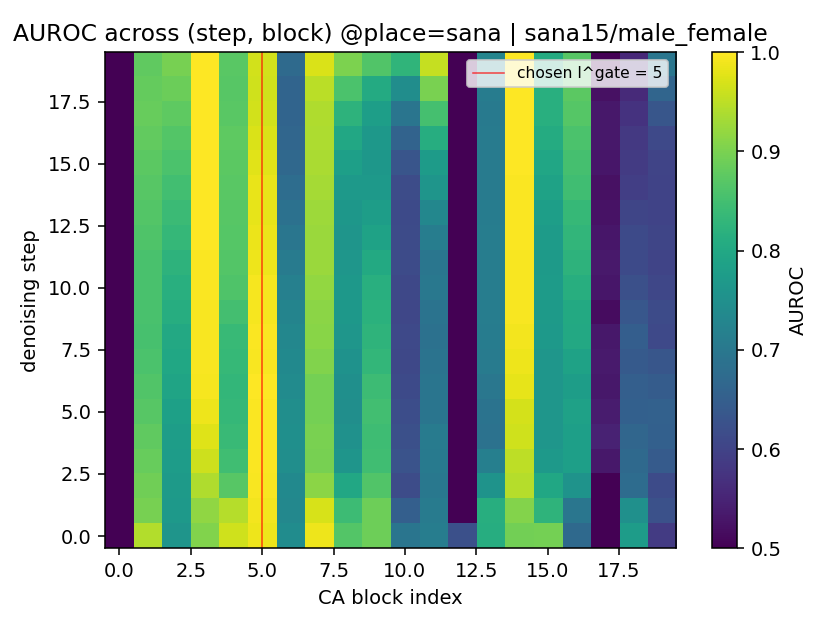}
  \caption{SANA-1.5 ($l^{gate}{=}5$)}
\end{subfigure}
\caption{Layer-wise gate AUROC for the gender attribute on all three backbones used in the main paper. Each cell of the heatmap is the AUROC of the maximal token response statistic (Eq.~\ref{eq:3}) for the male-to-female steering direction at the corresponding (denoising step, CA block index), computed on the same $n_{\text{pos}}{=}n_{\text{neg}}$ calibration prompts used to fit the per-direction threshold $thr^{a}$. The red vertical line marks $l^{gate}$ chosen by Algorithm~\ref{alg:lgate_selection}, which recovers the manually selected layer in every case.}
\label{fig:gate_auroc_heatmap}
\end{figure*}

Fig.~\ref{fig:gate_auroc_heatmap} shows the AUROC across (denoising step $\times$ CA block index) on the gender attribute for all three backbones in the main paper. On all three, the highest AUROC band is concentrated on the same one or two CA block indices across every denoising step (the colour bands run vertically in each panel). Algorithm~\ref{alg:lgate_selection} picks the layer at the right edge of this band: block 4 on SD-1.5, block 17 on SDXL, and block 5 on SANA-1.5. These are the layers that we use as $l^{gate}$ in all our experiments. 

\subsection{Subtle and low-saliency attributes}
\label{sec:subtle_attributes_supp}

The main paper and Sec.~\ref{sec:additional_attributes} of this supplementary validate EquiSteer on attributes with strong visual saliency --- \textit{gender}, \textit{race}, \textit{age}, \textit{body type}, and \textit{eyeglasses}. In this section, we test whether the gating mechanism, which relies on the diffusion model having internalised the target attribute as a separable direction in cross-attention output space, continues to work for less visually salient or more subtle demographic concepts. We then further evaluate gating mechanism robustness on such concepts as \textit{religion}, \textit{disability}, or \textit{socio-economic status}.

We probe this question by measuring the gate's separability AUROC --- the ability of EquiSteer's maximal token response statistic (Eq.~\ref{eq:3}) at the gating layer $l^{gate}$ to distinguish attribute-specific from neutral prompts --- across a deliberately chosen set of seven low-saliency concepts. AUROC at the gating layer is the upper bound on EquiSteer's ability to detect and intervene on an attribute: if no choice of threshold separates positive from negative prompts at the gate, the rest of the pipeline (orthogonalisation and adaptive re-injection) cannot help. Conversely, AUROC close to $1$ indicates that the attribute is internalised as a linearly separable direction and that EquiSteer can in principle gate on it.

For each concept we evaluate \emph{two prompt formulations}, motivated by the observation that the diffusion model could in principle key on either a visible clothing marker (e.g., a hijab) or on the identity word (e.g., ``muslim'') itself --- and the two need not be encoded in the same direction:

\begin{itemize}
    \item \textbf{Visual marker prompts} probe whether the model has internalised the attribute via a visible signifier:
        \textit{``a man wearing a hijab''} (Muslim),
        \textit{``a man wearing a turban''} (Sikh),
        \textit{``a man wearing a cross necklace''} (Christian),
        \textit{``a man in a wheelchair''} (disability),
        \textit{``a man with a heavy build''} (heavy build),
        \textit{``a man wearing expensive elegant clothes and jewellery''} (wealthy),
        \textit{``a man wearing worn out tattered clothes''} (poor).
    \item \textbf{Identity word prompts} probe whether the identity term itself is internalised, independently of any visual marker:
        \textit{``a muslim man''},
        \textit{``a sikh man''},
        \textit{``a christian man''},
        \textit{``a disabled man''},
        \textit{``an obese man''},
        \textit{``a wealthy man''},
        \textit{``a poor man''}.
\end{itemize}

For every (backbone $\times$ concept $\times$ formulation) cell we generate $n_{\text{pos}}{=}20$ images with the attribute-specific prompt and $n_{\text{neg}}{=}20$ with the neutral prompt \textit{``a man''}, dump the cross-attention outputs at $t{=}0$, score each image by the maximal token response statistic against the corresponding contrastive steering direction, and compute the resulting AUROC. The gate layer $l^{gate}$ used is the same as in all other concepts.

\begin{table}[th!]
  \centering
  \caption{Gate-direction AUROC for EquiSteer's gating layer on seven \emph{subtle / low-saliency} attributes, evaluated separately under two prompt formulations: \emph{visual marker} (e.g.\ \textit{``a man wearing a hijab''}) and \emph{identity word} (e.g.\ \textit{``a muslim man''}). Each cell reports the AUROC of the maximal token response statistic (Eq.~\ref{eq:3}) computed at $t{=}0$ on $n_{\text{pos}}{=}n_{\text{neg}}{=}20$ generations, separating attribute-specific from neutral prompts. The full prompt list is in Sec.~\ref{sec:subtle_attributes_supp}.}
  \label{tab:subtle_attributes_auroc}
  \scriptsize
  \setlength{\tabcolsep}{6pt}
  \begin{tabular}{l|cc|cc}
  \toprule
  & \multicolumn{2}{c|}{SDXL} & \multicolumn{2}{c}{SANA-1.5} \\
  \cmidrule(lr){2-3}\cmidrule(lr){4-5}
  Concept & visual & identity & visual & identity \\
  \midrule
  \multicolumn{5}{l}{\textit{Religion}} \\
  Muslim                    & \textbf{1.000} & \textbf{1.000} & \textbf{1.000} & \textbf{1.000} \\
  Sikh                      & \textbf{1.000} & \textbf{1.000} & \textbf{1.000} & \textbf{1.000} \\
  Christian                 & \textbf{1.000} & \textbf{1.000} & \textbf{1.000} & \textbf{1.000} \\
  \midrule
  \multicolumn{5}{l}{\textit{Other demographic}} \\
  Disability                & \textbf{1.000} & \textbf{1.000} & \textbf{1.000} & \textbf{1.000} \\
  Heavy build               & \textbf{1.000} & \textbf{1.000} & \textbf{1.000} & \textbf{1.000} \\
  \midrule
  \multicolumn{5}{l}{\textit{Socio-economic}} \\
  Wealthy                   & \textbf{1.000} & \textbf{1.000} & \textbf{1.000} & \textbf{1.000} \\
  Poor                      & \textbf{1.000} & \textbf{1.000} & \textbf{1.000} & \textbf{1.000} \\
  \bottomrule
  \end{tabular}
\end{table}

\noindent \textbf{Results.}
Tab.~\ref{tab:subtle_attributes_auroc} reports the AUROC for all 28 (backbone $\times$ concept $\times$ formulation) cells. \emph{Every cell achieves AUROC $=$ 1.000}: the gate perfectly separates attribute-specific from neutral prompts in every condition, on both backbones, under both prompt formulations, for every concept tested. The same is true of AUPRC.

This shows that the gate works on subtle attributes just as well as on the headline attributes of the main paper. Religion, disability, body type, and socio-economic status seem to all be internalised by SDXL and SANA-1.5 as linearly separable directions in CA-output space at $l^{gate}$.

Next, note that the two prompt formulations are equally separable: the identity-word prompts (e.g.\ \textit{``a muslim man''}) achieve AUROC $=$ 1 just as the visual-marker prompts (e.g.\ \textit{``a woman wearing a hijab''}) do. This rules out a shortcut explanation in which the model only encodes the clothing marker rather than the identity itself --- the identity term has its own internalised direction, distinguishable from the neutral subject \textit{``a man''} at the gating layer.

However, this section measures whether the gate \emph{can} detect the attribute, it does not run end-to-end debiasing for every subtle concept. The result we report --- AUROC $=$ 1 across all 28 cells --- is the necessary condition that justifies extending EquiSteer to a new concept, while the sufficient condition adds the construction of contrastive prompts in the spirit of Sec.~\ref{sec:prompts_steering_vecs}.

\subsection{Threshold-multiplier sensitivity}
\label{sec:thrshift_supp}

The gating decision (Sec.~\ref{sec:dot_product_indicator}) is parameterised by a single per-direction threshold $thr^{a}$, calibrated once from contrastive prompt pairs and used unchanged across all (profession, prompt) combinations at inference time. In this section, we study sensitivity of the end-to-end debiasing performance to this threshold. We probe this by introducing a single multiplier $m$ that scales every $thr^{a}$ uniformly at inference time: the gate-firing condition becomes
\[
  s \;=\; \max_{a}\bigl(dp^{a}_{l^{gate} 0} - m\cdot thr^{a}\bigr) \;>\; 0,
\]
where the gate fires (and EquiSteer \emph{skips} debiasing for that image) whenever $s > 0$. Two limits are interpretable: $m{=}0$ sets every threshold to zero, so the gate fires for every prompt and debiasing is never applied, i.e. this recovers the vanilla model. Conversely, $m \to \infty$ makes the gate never fire, so EquiSteer is applied unconditionally, and attribute-specific prompts such as \textit{``A photo of a male doctor''} get incorrectly debiased. The paper default is $m{=}1$.

We set $m \in \{0.0, 0.5, 1.0, 1.5, 2.0\}$ in two prompt families: the standard single-template prompt used throughout the main experiments, and the long contextual templates from Sec.~\ref{sec:transferability_supp}.

\subsubsection{Standard prompts.}
\label{sec:thrshift_standard}

Tab.~\ref{tab:thrshift_standard_gender} reports the per-multiplier (parity gap, attribute-specific recall) pair for gender on all three backbones, and Tab.~\ref{tab:thrshift_standard_race} reports the SDXL race sweep. As expected, the rows are monotonic: increasing $m$ lowers the parity gap $\Delta$ on neutral prompts (more debiasing applied) and eventually degrades the attribute-specific recall (incorrect debiasing of explicit prompts).

\begin{table*}[th!]
  \centering
  \caption{Standard-prompt threshold-multiplier sweep for the \textit{gender} attribute. The threshold multiplier $m$ scales the per-direction gate threshold ($m{=}0$ disables debiasing entirely; $m{=}1$ is the paper default; $m{\to}\infty$ removes the gate and debiases every prompt). $\Delta = \tfrac{1}{|\text{prof}|}\sum_{p}|r_{\text{female},p} - 0.5|$ on neutral prompts \textit{``A photo of a \{prof\}''}; $\mathrm{Rec}_{m/f}$ is the attribute-specific recall on \textit{``A photo of a \{male/female\} \{prof\}''} prompts (target $1.0$). Means over the eight evaluation professions; $n{=}100$ images per cell. Best Pareto point per backbone marked in bold.}
  \label{tab:thrshift_standard_gender}
  \scriptsize
  \setlength{\tabcolsep}{5pt}
  \begin{tabular}{c|ccc|ccc|ccc}
  \toprule
  & \multicolumn{3}{c|}{SD-1.5} & \multicolumn{3}{c|}{SDXL} & \multicolumn{3}{c}{SANA-1.5} \\
  \cmidrule(lr){2-4}\cmidrule(lr){5-7}\cmidrule(lr){8-10}
  $m$
  & $\Delta$ & $\mathrm{Rec}_m$ & $\mathrm{Rec}_f$
  & $\Delta$ & $\mathrm{Rec}_m$ & $\mathrm{Rec}_f$
  & $\Delta$ & $\mathrm{Rec}_m$ & $\mathrm{Rec}_f$ \\
  \midrule
  $0.0$         & 0.430 & 0.990 & 0.988 & 0.375 & 0.992 & 1.000 & 0.463 & 1.000 & 1.000 \\
  $0.5$         & 0.380 & 0.990 & 0.988 & 0.297 & 0.992 & 1.000 & 0.463 & 1.000 & 1.000 \\
  $1.0$ (paper) & 0.253 & 0.990 & 0.988 & 0.120 & 0.992 & 1.000 & \textbf{0.153} & \textbf{1.000} & \textbf{1.000} \\
  $1.5$         & \textbf{0.128} & \textbf{0.990} & \textbf{0.932} & 0.105 & 0.992 & 1.000 & 0.070 & 1.000 & 0.895 \\
  $2.0$         & 0.113 & 0.960 & 0.880 & \textbf{0.060} & \textbf{0.983} & \textbf{1.000} & 0.065 & 0.555 & 0.590 \\
  \bottomrule
  \end{tabular}
\end{table*}

\begin{table}[th!]
  \centering
  \caption{Standard-prompt threshold-multiplier sweep for the \textit{race} attribute on SDXL (5-way, target $r_c{=}0.2$). $\Delta = \tfrac{1}{5}\sum_{c}|r_c - 0.2|$ on neutral prompts; self-recall is the average rate at which \textit{``A photo of a \{race\} \{prof\}''} prompts produce the requested race. Means over the eight evaluation professions; $n{=}100$ images per cell.}
  \label{tab:thrshift_standard_race}
  \scriptsize
  \setlength{\tabcolsep}{6pt}
  \begin{tabular}{c|cc}
  \toprule
  $m$ & $\Delta$ ($\downarrow$) & Self-recall ($\uparrow$) \\
  \midrule
  $0.0$         & 0.132 & 0.890 \\
  $0.5$         & 0.098 & 0.890 \\
  $1.0$ (paper) & \textbf{0.090} & \textbf{0.857} \\
  $1.5$         & 0.097 & 0.716 \\
  $2.0$         & 0.092 & 0.616 \\
  \bottomrule
  \end{tabular}
\end{table}

\emph{SDXL.} The recall on attribute-specific prompts is essentially flat at $\geq 0.98$ all the way up to $m{=}2.0$, while $\Delta$ falls monotonically from $0.375$ (vanilla, $m{=}0$) to $0.060$ at $m{=}2.0$. The default $m{=}1$ results in $\Delta{=}0.120$, but for SDXL it is possible to gain a further $\sim 2\times$ tightening with no loss in attribute preservation by using $m{=}2$. We retain $m{=}1$ as the paper default for consistency with the other backbones, but flag $m{=}2$ as a free improvement when only SDXL is in scope.

\emph{SANA-1.5.} The Pareto curve is sharper: $m{=}1$ achieves $\Delta{=}0.153$ with perfect recall, and $m{=}1.5$ further tightens to $\Delta{=}0.070$ with a small recall hit on the female direction ($1.000 \to 0.895$). At $m{=}2.0$ recall collapses ($0.555 / 0.590$) --- the gate has effectively been removed and attribute-specific prompts are now over-corrected. Therefore, SANA-1.5 should not be pushed above $m{=}1.5$.

\emph{SD-1.5.} shows intermediate behaviour. $m{=}1.5$ approximately halves $\Delta$ ($0.253 \to 0.128$) with only a mild recall loss on the female direction.

\emph{SDXL race.} The race sweep shows different results compared to gender. (Tab.~\ref{tab:thrshift_standard_race}). $\Delta$ is already small at vanilla, and bottoms out at $m{=}1$ ($\Delta{=}0.090$). Pushing $m$ higher does \emph{not} help: $\Delta$ stays around $0.09$ but self-recall drops sharply ($0.857 \to 0.616$) --- the gate closes less often on the explicit-race prompts and EquiSteer overwrites the requested race. Therefore, the default $m{=}1$ is the right choice on SDXL race.

\subsubsection{Long contextual prompts.}
\label{sec:thrshift_long}

The gate-firing condition depends on the absolute magnitude of the maximal token response $dp^{a}_{l^{gate} 0}$, which grows with prompt length: longer prompts produce more tokens with non-trivial overlap with the attribute steering direction, pushing more cells across the calibrated threshold and causing the gate to fire (and EquiSteer to skip) more often than on the short standard template. Tab.~\ref{tab:thrshift_long} sweeps $m \in \{0.5, 1.0, 1.5, 2.0\}$ across the four long contextual templates of Sec.~\ref{sec:transferability_supp}.

\begin{table}[th!]
  \centering
  \caption{Long-prompt threshold-multiplier sweep for the \textit{gender} attribute. The four templates (\texttt{candid}, \texttt{latenight}, \texttt{office}, \texttt{presentation}) are the long contextual prompts defined in Sec.~\ref{sec:transferability_supp}. Each cell is the mean $\Delta = |r_{\text{female}} - 0.5|$ averaged over the eight evaluation professions; $n{=}50$ images per template-profession cell. \textbf{Mean} row averages over the four templates; best mean per backbone in bold.}
  \label{tab:thrshift_long}
  \scriptsize
  \setlength{\tabcolsep}{6pt}
  \begin{tabular}{cl|cccc}
  \toprule
  & & \multicolumn{4}{c}{Threshold multiplier $m$} \\
  \cmidrule(lr){3-6}
  Backbone & Template & $0.5$ & $1.0$ (paper) & $1.5$ & $2.0$ \\
  \midrule
  \multirow{5}{*}{\rotatebox[origin=c]{90}{SDXL}}
    & \texttt{candid}       & 0.282 & 0.162 & 0.133 & 0.055 \\
    & \texttt{latenight}    & 0.242 & 0.198 & 0.150 & 0.115 \\
    & \texttt{office}       & 0.265 & 0.100 & 0.175 & 0.075 \\
    & \texttt{presentation} & 0.232 & 0.180 & 0.090 & 0.068 \\
  \cmidrule(lr){2-6}
    & \textbf{Mean}         & 0.256 & 0.160 & 0.137 & \textbf{0.078} \\
  \midrule
  \multirow{5}{*}{\rotatebox[origin=c]{90}{SANA-1.5}}
    & \texttt{candid}       & 0.360 & 0.158 & 0.055 & 0.065 \\
    & \texttt{latenight}    & 0.392 & 0.195 & 0.135 & 0.085 \\
    & \texttt{office}       & 0.417 & 0.180 & 0.053 & 0.060 \\
    & \texttt{presentation} & 0.350 & 0.090 & 0.053 & 0.070 \\
  \cmidrule(lr){2-6}
    & \textbf{Mean}         & 0.380 & 0.156 & \textbf{0.074} & 0.070 \\
  \bottomrule
  \end{tabular}
\end{table}

The default $m{=}1$ leaves a noticeable residual gap on long prompts compared to standard prompts ($\Delta_{\text{long}}{=}0.160$ vs $\Delta_{\text{std}}{=}0.120$ on SDXL; $0.156$ vs $0.153$ on SANA-1.5). Increasing $m$ to $1.5$ on SANA-1.5 or to $2.0$ on SDXL recovers roughly half of the residual gap:

\begin{itemize}
    \item \emph{SANA-1.5.} Mean $\Delta$ falls from $0.156$ at $m{=}1$ to $0.074$ at $m{=}1.5$ ($\sim 53\%$ further reduction). At $m{=}2.0$ the mean settles around $0.070$ but, as in the standard-prompt sweep, attribute-specific recall degrades; $m{=}1.5$ is the safe operating point.
    \item \emph{SDXL.} Mean $\Delta$ falls from $0.160$ at $m{=}1$ to $0.137$ at $m{=}1.5$ and to $0.078$ at $m{=}2.0$ ($\sim 51\%$ further reduction at $m{=}2$). The recall stability observed in the standard-prompt sweep (Tab.~\ref{tab:thrshift_standard_gender}) carries over: $m{=}2$ is again the safe operating point on SDXL.
\end{itemize}

Based on the observations above, we recommend lifting the threshold multiplier to $m{\approx}1.5$ (SANA-1.5) or $m{\approx}2$ (SDXL) for inference workloads that routinely use long contextual prompts. The default $m{=}1$ remains the right setting for the standard short-template setting used throughout the experiments.

\clearpage

\section{Classifier calibration and human evaluation}
\label{sec:classifier_calibration_supp}

The main paper measures the parity gap $\Delta$ using a CLIP ViT-L/14 zero-shot classifier. In this section, we address two natural concerns of this choice, namely: (i) does CLIP's labelling agree with what a human would label, and (ii) when CLIP and a human disagree, does that disagreement systematically bias the reported parity gap in either direction? 

We use the following procedure for calibration: first we collect labels from two human annotators on a stratified subset of generations. Second, we validate a strong VLM oracle (GPT-4o) against those human labels, showing that the two uniformly agree. This allows us to scale the human-style evaluation to all $\sim 1{,}600$ images per cell without paying the per-image annotation cost. Third we use the GPT-4o oracle to compare CLIP against alternative classifiers (BLIP-VQA for eyeglasses) and to recompute $\Delta$ on every attribute under each classifier.

\subsection{Human evaluation}
\label{sec:classifier_calibration_human}

We sampled a $600$-image stratified pack from the generations: three attributes (\textit{gender}, \textit{race}, \textit{eyeglasses}) $\times$ two backbones (SDXL, SANA-1.5) $\times$ two modes (vanilla, EquiSteer) $\times$ $50$ images per cell, drawn uniformly across the eight evaluation professions. Two in-lab annotators (H1, H2) independently labelled every image, blind to (mode, classifier predictions). The exact attribute label set is the same one used in the rest of the paper: $\{\textit{male}, \textit{female}\}$ for gender; $\{\textit{white}, \textit{black}, \textit{asian}, \textit{indian}, \textit{latino}\}$ for race; $\{\textit{eyeglasses}, \textit{no\_eyeglasses}\}$ for eyeglasses. Annotators were allowed to mark images \textit{uncertain} when no class was confidently identifiable.

Tab.~\ref{tab:calib_human_agreement} reports the pairwise agreement between annotators along with Cohen's $\kappa$ in the same table (left two columns). Agreement is substantial-to-perfect on every cell:
$\kappa \geq 0.98$ on the binary axes (gender, eyeglasses) on both backbones, and $\kappa \in [0.71, 0.73]$ on the harder 5-way race axis. The lower race $\kappa$ reflects the intrinsic difficulty of race classification from a single portrait --- the two human annotators disagree with each other roughly as often as either disagrees with the strongest available oracle (Sec.~\ref{sec:classifier_calibration_oracle}). We use this human-agreement profile as the trust benchmark for the rest of the section.

\begin{table}[th!]
  \centering
  \caption{Consolidated pairwise label agreement on the 600-image human-evaluation pack ($n{=}100$ images per cell). Two in-lab annotators (H1, H2) labelled every image. Paired columns show the (H1\,/\,H2)-vs-source agreement; e.g.\ a cell ``$100\%$\,/\,$99\%$'' means H1\,$\leftrightarrow$\,GPT $=100\%$ and H2\,$\leftrightarrow$\,GPT $=99\%$.}
  \label{tab:calib_human_agreement}
  \scriptsize
  \setlength{\tabcolsep}{5pt}
  \begin{tabular}{l|c@{\hskip 3pt}c|c|c|c}
  \toprule
  Cell & H1\,$\leftrightarrow$\,H2 & $\kappa$(H1,H2)
       & (H1\,/\,H2)\,$\leftrightarrow$\,GPT
       & (H1\,/\,H2)\,$\leftrightarrow$\,CLIP
       & CLIP\,$\leftrightarrow$\,GPT \\
  \midrule
  SDXL $\times$ gender         & $99.0\%$  & $0.979$ & $100\%$\,/\,$99\%$ & $98\%$\,/\,$97\%$ & $98.0\%$ \\
  SDXL $\times$ race           & $78.7\%$  & $0.711$ & $95\%$\,/\,$83\%$ & $54\%$\,/\,$53\%$ & $54.3\%$ \\
  SDXL $\times$ eyeglasses     & $99.0\%$  & $0.978$ & $98\%$\,/\,$99\%$ & $68\%$\,/\,$69\%$ & $66.7\%$ \\
  SANA $\times$ gender     & $100.0\%$ & $1.000$ & $100\%$\,/\,$100\%$ & $100\%$\,/\,$100\%$ & $100.0\%$ \\
  SANA $\times$ race       & $80.0\%$  & $0.731$ & $95\%$\,/\,$86\%$ & $48\%$\,/\,$55\%$ & $51.8\%$ \\
  SANA $\times$ eyeglasses & $100.0\%$ & $1.000$ & $100\%$\,/\,$100\%$ & $88\%$\,/\,$88\%$ & $88.0\%$ \\
  \bottomrule
  \end{tabular}
\end{table}

\subsection{GPT-4o as a scalable oracle, validated against humans}
\label{sec:classifier_calibration_oracle}

Hand-labelling all $\sim 1{,}600$ images per cell across five attributes and two backbones would require on the order of $20{,}000$ annotations per labeller, which is impractical at the scale needed to recompute $\Delta$ end-to-end. We therefore use GPT-4o (via the API) as a scalable oracle: for each image we send a single-message prompt asking the model to assign one class from the attribute-specific label set, or \textit{uncertain} if the image does not clearly fit any class. The same prompt formulation, label set, and \textit{uncertain} option are used as for the human annotators.

To validate that GPT-4o tracks human judgements well enough to be used as a proxy at scale, we compare GPT-4o's labels against H1 and H2 on exactly the same 600-image pack (Tab.~\ref{tab:calib_human_agreement}, middle two columns). On every cell, \emph{GPT-4o agrees with each individual annotator at least as well as the two annotators agree with each other}:

\begin{itemize}
    \item On the two binary axes (gender, eyeglasses) GPT-4o $\leftrightarrow$ human agreement is $\geq 97.8\%$ on every cell, matching the $99$--$100\%$ human-to-human agreement.
    \item On the harder 5-way race axis, GPT-4o $\leftrightarrow$ H1 agreement is $95.1\%$ on SDXL and $95.3\%$ on SANA-1.5 --- substantially \emph{higher} than the $78.7\%$ / $80.0\%$ human-to-human agreement and the $\sim 50$--$55\%$ human-to-CLIP agreement on the same images. GPT-4o $\leftrightarrow$ H2 agreement is in a similar range ($82.7\%$ / $85.9\%$).
\end{itemize}

The conclusion we draw is that GPT-4o is a defensible cheap proxy for human labels on this task: any conclusion about the relative performance of classifiers obtained against GPT-4o would also be obtained against the human annotators, on every cell we have ground truth for. The rest of the section uses GPT-4o as the oracle on the full $n{\approx}1{,}600$ per-cell calibration set.

\subsection{CLIP vs the GPT-4o oracle}
\label{sec:classifier_calibration_clip_vs_oracle}

For each of the $10$ (backbone $\times$ attribute) cells we re-evaluated the same EquiSteer / vanilla generations used elsewhere in this paper (8 professions, 2 modes, $n{=}100$ images per (cell $\times$ profession $\times$ mode) cell, giving $n_{\text{total}}{\approx}1{,}600$ per backbone-attribute cell). We dropped images on which the oracle returned \textit{uncertain} from the comparison ($n_{\text{clf}}$ in Tab.~\ref{tab:calib_clip_vs_oracle}). For every retained image we have a CLIP zero-shot label, a GPT-4o label, and the (mode, profession) it came from. We compute (i) the overall CLIP $\leftrightarrow$ GPT-4o agreement, (ii) the CLIP per-class recall when the oracle is taken as ground truth, and (iii) the parity gap $\Delta$ under each label source.

\begin{table}[th!]
  \centering
  \caption{CLIP zero-shot vs GPT-4o oracle on $n{\approx}1{,}600$ EquiSteer / vanilla generations per (backbone $\times$ attribute) cell. \textit{Agree}: per-image label agreement after dropping oracle-uncertain images. \textit{CLIP worst}: lowest CLIP per-class recall against the oracle, with the class label. $\Delta$: parity gap on the same images under each label source, formatted as $\Delta_{\text{vanilla}} \to \Delta_{\text{FS}}$ (FS reduction \%). Stronger reduction across classifiers bolded.}
  \label{tab:calib_clip_vs_oracle}
  \scriptsize
  \setlength{\tabcolsep}{5pt}
  \resizebox{\linewidth}{!}{%
  \begin{tabular}{l@{\hskip 3pt}l|cc|c|c|c}
  \toprule
  Backbone & Attribute & $n$ & Agree & CLIP worst recall & $\Delta$ under \textbf{CLIP} & $\Delta$ under \textbf{GPT-4o} \\
  \midrule
  \multirow{5}{*}{SDXL}
    & gender     & 1535 & $98.4\%$ & male $0.98$        & $0.399 \to 0.239$ ($40\%$) & $0.415 \to 0.243$ ($\mathbf{41\%}$) \\
    & race       & 1287 & $59.5\%$ & \textbf{white $0.26$} & $0.145 \to 0.095$ ($35\%$) & $0.243 \to 0.085$ ($\mathbf{65\%}$) \\
    & eyeglasses & 1509 & $67.8\%$ & \textbf{eyeglasses $0.48$} & $0.282 \to 0.172$ ($39\%$) & $0.289 \to 0.088$ ($\mathbf{69\%}$) \\
    & age        & 1583 & $71.7\%$ & \textbf{young $0.50$}       & $0.274 \to 0.160$ ($41\%$) & $0.263 \to 0.074$ ($\mathbf{72\%}$) \\
    & body       & 1463 & $54.1\%$ & \textbf{slim $0.45$}        & $0.327 \to 0.199$ ($39\%$) & $0.257 \to 0.098$ ($\mathbf{62\%}$) \\
  \midrule
  \multirow{5}{*}{SANA}
    & gender     & 1598 & $99.7\%$ & female $0.99$       & $0.474 \to 0.146$ ($69\%$) & $0.480 \to 0.146$ ($\mathbf{70\%}$) \\
    & race       & 1291 & $45.7\%$ & \textbf{white $0.05$} & $0.216 \to 0.127$ ($41\%$) & $0.286 \to 0.117$ ($\mathbf{59\%}$) \\
    & eyeglasses & 1590 & $87.1\%$ & \textbf{eyeglasses $0.79$} & $0.309 \to 0.221$ ($29\%$) & $0.269 \to 0.165$ ($\mathbf{39\%}$) \\
    & age        & 1529 & $73.1\%$ & \textbf{young $0.51$}       & $0.382 \to 0.093$ ($76\%$) & $0.394 \to 0.103$ ($74\%$) \\
    & body       & 1563 & $56.0\%$ & average $0.54$      & $0.392 \to 0.287$ ($27\%$) & $0.365 \to 0.210$ ($\mathbf{43\%}$) \\
  \bottomrule
  \end{tabular}%
  }
\end{table}

Tab.~\ref{tab:calib_clip_vs_oracle} reveals a consistent pattern. CLIP is essentially calibrated for binary \textit{gender} ($98$--$99.7\%$ overall agreement, both per-class recalls $\geq 0.98$). On every other attribute it has a specific systematic failure mode — a single class with materially lower recall — and this failure mode is exactly what makes the reported $\Delta$ a conservative estimate of EquiSteer's true effect:

\begin{itemize}
    \item \emph{Race}. CLIP under-recognises \textit{white} subjects: recall $0.26$ on SDXL and $0.05$ on SANA-1.5. CLIP routinely labels \textit{white} as \textit{latino} or \textit{asian}. Because vanilla SDXL / SANA generate predominantly \textit{white} subjects, CLIP under-reports the vanilla white-skew (vanilla $\Delta_{\text{CLIP}}{=}0.145$ on SDXL vs $\Delta_{\text{oracle}}{=}0.243$). After EquiSteer adds non-white subjects, CLIP recognises them clearly, so the EquiSteer side moves much less. The net effect is that the CLIP-measured \emph{EquiSteer reduction} on race is $35\%$ on SDXL and $41\%$ on SANA, while the oracle reduction is $65\%$ and $59\%$ respectively.
    \item \emph{Eyeglasses}. CLIP under-detects the eyeglasses class (recall $0.48$ on SDXL, $0.79$ on SANA). EquiSteer reduces $\Delta$ by $39\%$/$29\%$ under CLIP and $\mathbf{69\%}$/$\mathbf{39\%}$ under the oracle.
    \item \emph{Age}. CLIP under-recognises \textit{young} subjects (recall $\sim 0.5$ on both backbones). The EquiSteer reduction is $41\%$/$76\%$ under CLIP and $\mathbf{72\%}$/$74\%$ under the oracle.
    \item \emph{Body type}. CLIP recall is uniformly modest ($0.45$--$0.63$) across all three classes; slim and average are routinely confused. The EquiSteer reduction is $39\%$/$27\%$ under CLIP and $\mathbf{62\%}$/$\mathbf{43\%}$ under the oracle.
\end{itemize}

Therefore, CLIP-only $\Delta$ values reported in the main paper are conservative on every multi-class attribute. The actual debiasing effect, measured under the oracle that aligns with humans, seems to be consistently at least as large and often substantially larger.

\subsection{Eyeglasses three-way classifier comparison}
\label{sec:classifier_calibration_eyeglasses}

In Sec.~\ref{sec:eyeglasses_supp} we report eyeglasses results using CLIP ans BLIP-VQA classifiers, ans argue that CLIP under-detects the eyeglasses class. Here we use GPT-4o to check whether the BLIP-VQA classifier is a better proxy for human-aligned eyeglasses labels than CLIP is.

Tab.~\ref{tab:calib_eyeglasses_threeway} reports the three-way comparison on the same $\sim 1{,}600$ generations per backbone used above. The results show that BLIP-VQA agrees with the GPT-4o oracle on $94$--$97\%$ of images on both backbones (vs CLIP's $67$--$87\%$), its per-class recall on the eyeglasses class is $0.97$ (SDXL) / $1.00$ (SANA) (vs CLIP's $0.48$ / $0.79$), and the EquiSteer reduction it measures matches the oracle to within $\pm 3$ percentage points: on SDXL BLIP says $0.270 \to 0.085$ ($69\%$) and the oracle says $0.289 \to 0.088$ ($70\%$). This validates the choice of BLIP-VQA as the better eyeglasses classifier, as it seems to be a much closer proxy for the human-aligned oracle than CLIP zero-shot.

\begin{table}[th!]
  \centering
  \caption{Eyeglasses three-way classifier comparison: CLIP zero-shot vs BLIP-VQA (the paper-appendix classifier) vs GPT-4o oracle on the same $n{\approx}1{,}600$ generations per backbone. Top block: pairwise label agreement. Middle block: per-class recall when the oracle is taken as ground truth. Bottom block: parity gap $\Delta$ under each classifier on the same images.}
  \label{tab:calib_eyeglasses_threeway}
  \scriptsize
  \setlength{\tabcolsep}{6pt}
  \begin{tabular}{l|cc}
  \toprule
   & SDXL & SANA-1.5 \\
  \midrule
  \multicolumn{3}{l}{\textit{Pairwise label agreement}} \\
  \quad CLIP $\leftrightarrow$ GPT-4o  & $67.8\%$ & $87.1\%$ \\
  \quad BLIP $\leftrightarrow$ GPT-4o  & $\mathbf{94.4\%}$ & $\mathbf{97.4\%}$ \\
  \quad CLIP $\leftrightarrow$ BLIP    & $67.0\%$ & $85.4\%$ \\
  \midrule
  \multicolumn{3}{l}{\textit{Per-class recall against GPT-4o}} \\
  \quad CLIP no\_eyeglasses & $0.96$ & $0.96$ \\
  \quad CLIP eyeglasses     & $\mathbf{0.48}$ & $\mathbf{0.79}$ \\
  \quad BLIP no\_eyeglasses & $0.91$ & $0.95$ \\
  \quad BLIP eyeglasses     & $\mathbf{0.97}$ & $\mathbf{1.00}$ \\
  \midrule
  \multicolumn{3}{l}{\textit{Parity gap $\Delta$ (vanilla $\to$ EquiSteer; reduction \%)}} \\
  \quad CLIP   & $0.282 \to 0.190 \ \ (33\%)$ & $0.309 \to 0.225 \ \ (27\%)$ \\
  \quad BLIP   & $0.270 \to \mathbf{0.085} \ \ (\mathbf{69\%})$ & $0.234 \to \mathbf{0.175} \ \ (25\%)$ \\
  \quad GPT-4o & $0.289 \to \mathbf{0.088} \ \ (\mathbf{70\%})$ & $0.269 \to \mathbf{0.165} \ \ (\mathbf{39\%})$ \\
  \bottomrule
  \end{tabular}
\end{table}

\subsection{Human labels distributions}
\label{sec:classifier_calibration_humansee}

Tab.~\ref{tab:calib_human_class_dist} reports the class distributions of vanilla and EquiSteer generations as labelled by the humans, on the same $n{=}50$ images-per-(cell $\times$ mode) subset used for the agreement analysis. 

\begin{table}[th!]
  \centering
  \caption{Human-annotated class distributions for the 600-image evaluation set. For each model and method, we evaluate 50 images per attribute condition and report values averaged over two annotators. Gender and eyeglasses are binary attributes, reported as the percentage of images classified as female or as wearing eyeglasses, with a target of 50\%. Race is reported as the percentage assigned to each of five classes, with a target of 20\% per class. $\Delta$ measures deviation from the target distribution: absolute deviation for binary attributes and mean absolute deviation across race classes. Full-scale results on 1,000 images using the GPT-4o oracle are reported in Tab.~\ref{tab:calib_clip_vs_oracle}. Bold values indicate the method closer to the target within each model.}

  \label{tab:calib_human_class_dist}
  \scriptsize
  \setlength{\tabcolsep}{4pt}
  \begin{tabular}{ll|cc|cc}
  \toprule
  & & \multicolumn{2}{c|}{SDXL} & \multicolumn{2}{c}{SANA-1.5} \\
  \cmidrule(lr){3-4}\cmidrule(lr){5-6}
  & Class & Vanilla & EquiSteer & Vanilla & EquiSteer \\
  \midrule
  \multirow{2}{*}{\textit{Gender}}
    & \% female      & $42\%$ & $39\%$ & $36\%$ & $24\%$ \\
    & $\Delta$       & $0.080$ & $0.110$ & $0.140$ & $0.260$ \\
  \midrule
  \multirow{6}{*}{\textit{Race}}
    & White          & $\mathbf{68\%}$ & $\mathbf{16\%}$ & $\mathbf{66\%}$ & $\mathbf{15\%}$ \\
    & Black          & $3\%$  & $\mathbf{32\%}$ & $0\%$  & $\mathbf{22\%}$ \\
    & Asian          & $3\%$  & $\mathbf{20\%}$ & $23\%$ & $21\%$ \\
    & Indian         & $8\%$  & $\mathbf{24\%}$ & $0\%$  & $\mathbf{28\%}$ \\
    & Latino         & $18\%$ & $9\%$   & $11\%$ & $14\%$ \\
    & $\Delta$       & $0.192$ & $\mathbf{0.063}$ & $0.198$ & $\mathbf{0.044}$ \\
  \midrule
  \multirow{2}{*}{\textit{Eyeglasses}}
    & \% eyeglasses  & $63\%$ & $\mathbf{52\%}$ & $50\%$ & $64\%$ \\
    & $\Delta$       & $0.133$ & $\mathbf{0.020}$ & $0.000$ & $0.140$ \\
  \bottomrule
  \end{tabular}
\end{table}

The most legible cell is \textit{race} on both backbones, where both vanilla models generate predominantly \textit{white} subjects ($66$--$68\%$) with the \textit{black} and \textit{indian} classes essentially absent ($0$--$8\%$). After EquiSteer the \textit{white} share drops to $\sim 15\%$ on both backbones, every minority class rises to at least $14\%$, and on SDXL all five classes lie within $\pm 12$ pp of the uniform target. The human-measured $\Delta$ on race drops from $0.192 \to 0.063$ on SDXL ($67\%$ reduction) and $0.198 \to 0.044$ on SANA-1.5 ($78\%$ reduction). The other axes show smaller (gender) or noisier ($n{=}50$ eyeglasses, sometimes already-balanced vanilla) changes. 

\clearpage

\section{More qualitative results on race concept}

\begin{figure}[th!]
\centering
\begin{subfigure}[t]{0.48\linewidth}
  \centering
  \includegraphics[width=\linewidth]{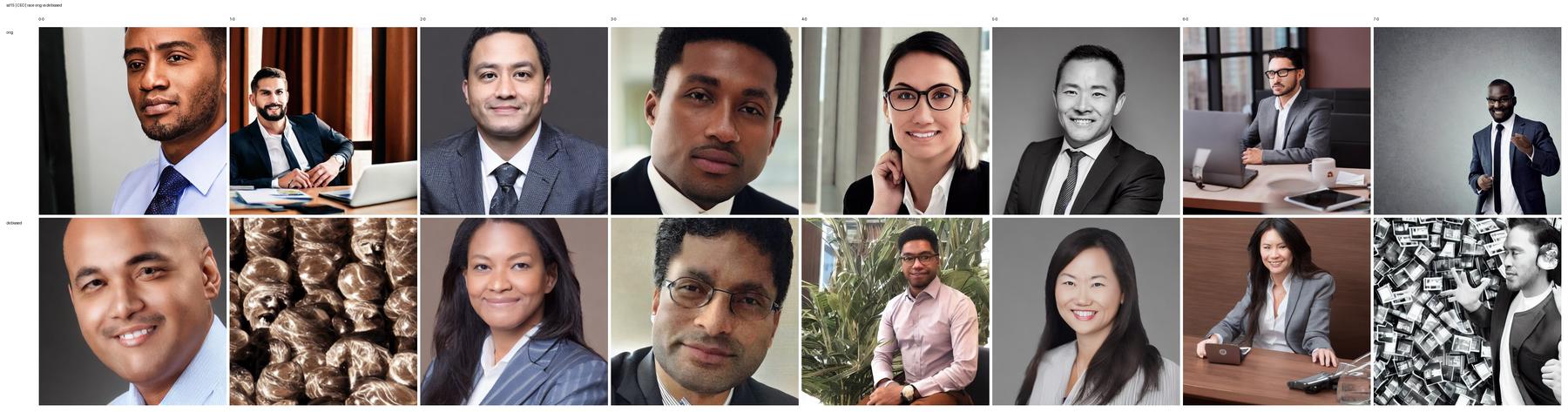}
  \caption{\textit{CEO}}
\end{subfigure}\hfill
\begin{subfigure}[t]{0.48\linewidth}
  \centering
  \includegraphics[width=\linewidth]{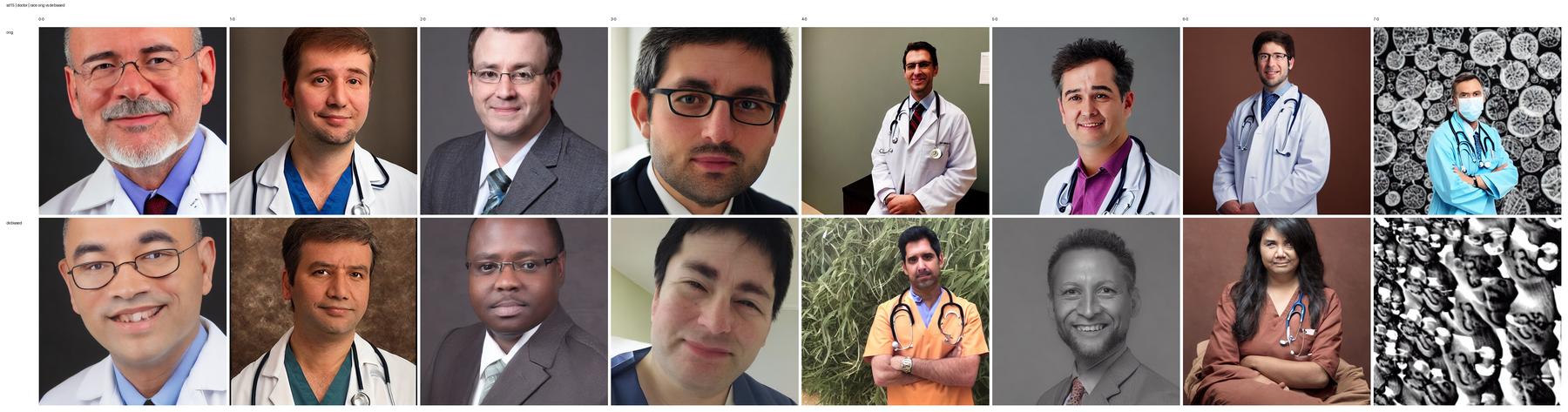}
  \caption{\textit{Doctor}}
\end{subfigure}

\begin{subfigure}[t]{0.48\linewidth}
  \centering
  \includegraphics[width=\linewidth]{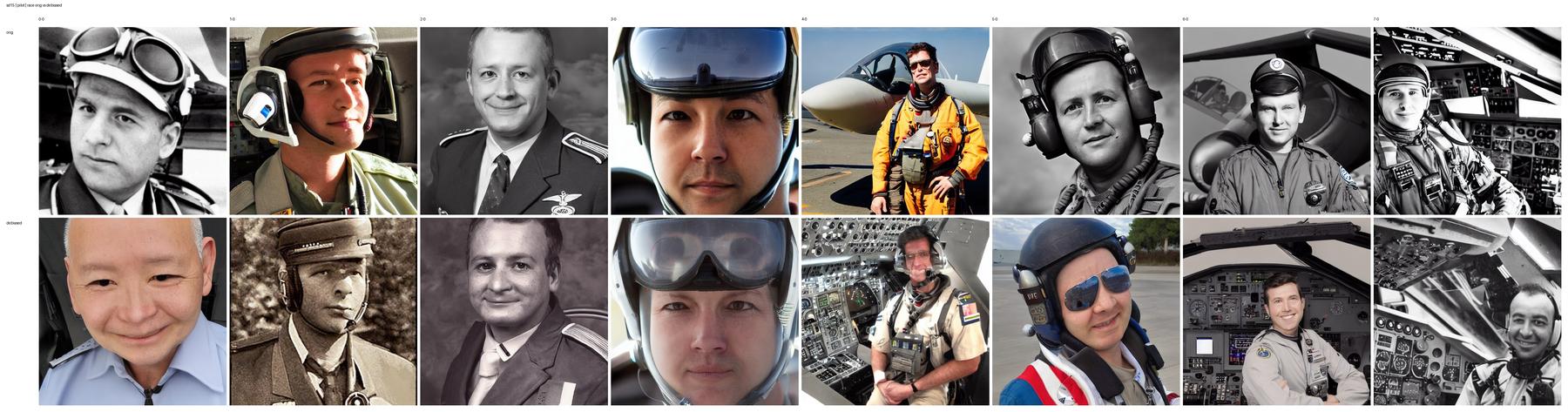}
  \caption{\textit{Pilot}}
\end{subfigure}\hfill
\begin{subfigure}[t]{0.48\linewidth}
  \centering
  \includegraphics[width=\linewidth]{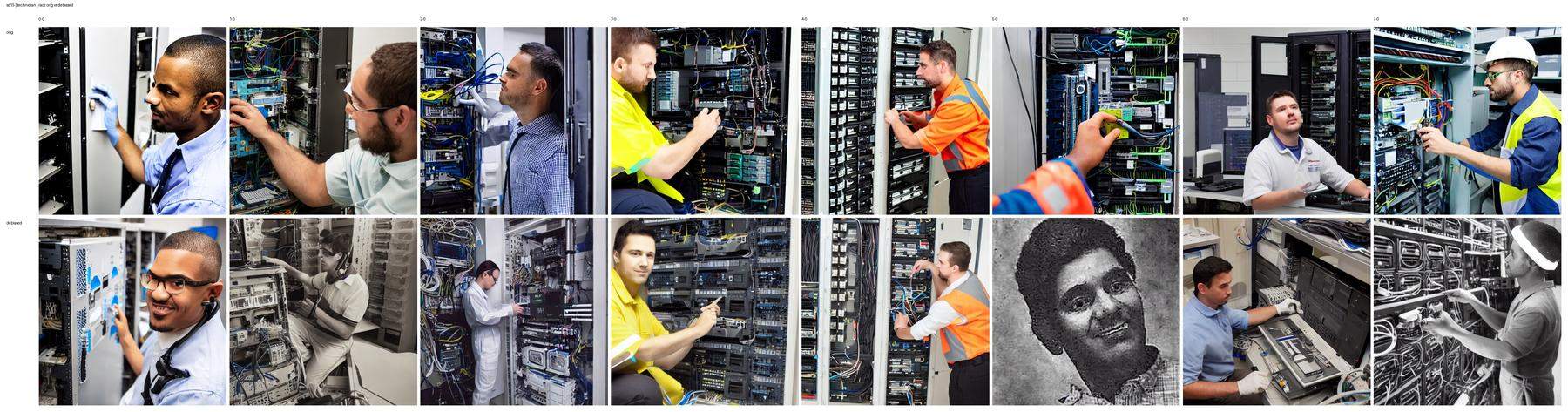}
  \caption{\textit{Technician}}
\end{subfigure}

\begin{subfigure}[t]{0.48\linewidth}
  \centering
  \includegraphics[width=\linewidth]{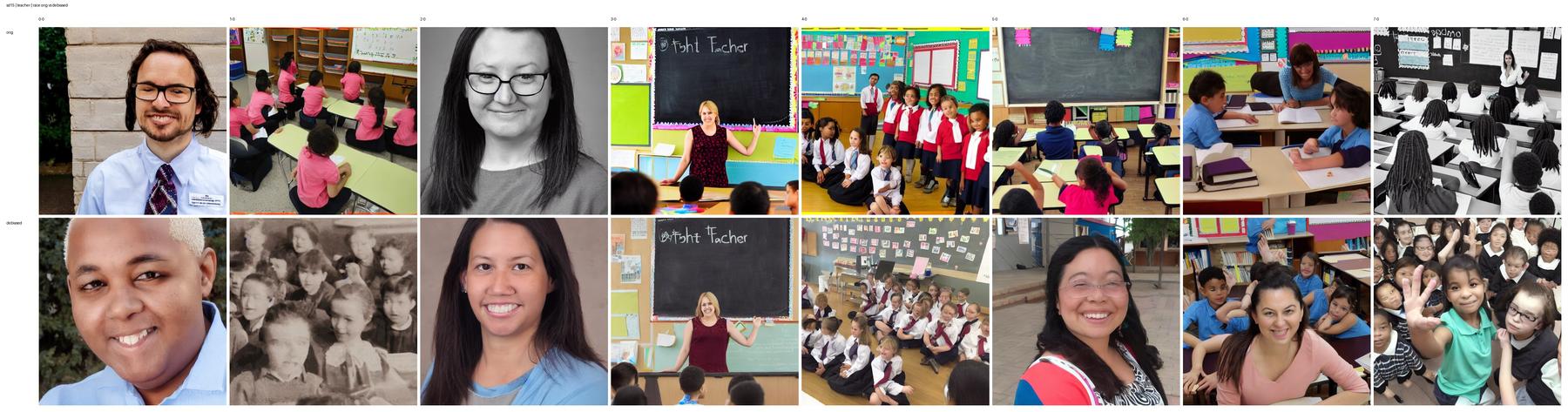}
  \caption{\textit{Teacher}}
\end{subfigure}\hfill
\begin{subfigure}[t]{0.48\linewidth}
  \centering
  \includegraphics[width=\linewidth]{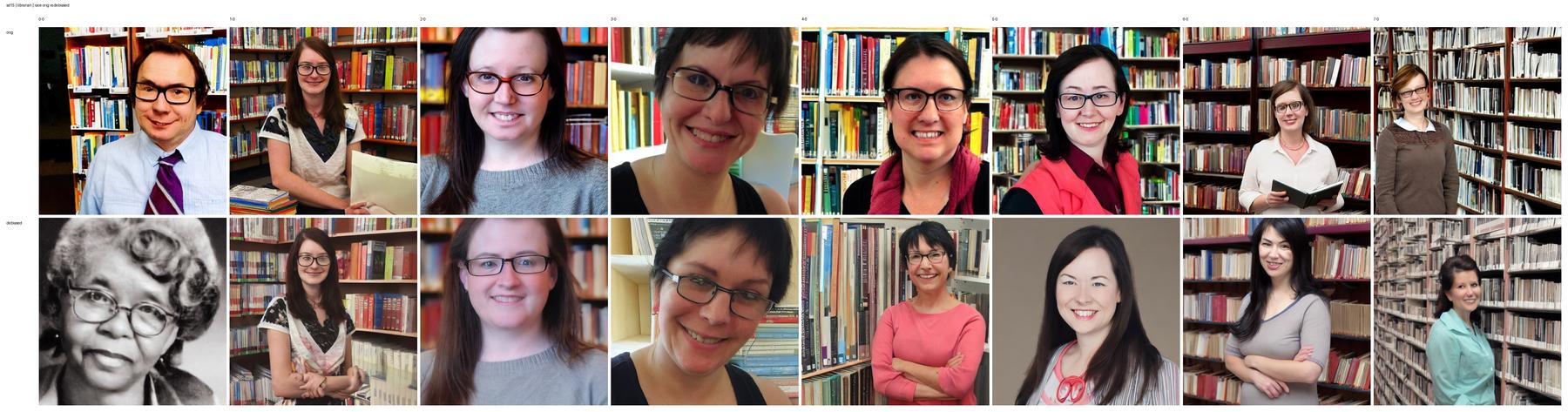}
  \caption{\textit{Librarian}}
\end{subfigure}

\begin{subfigure}[t]{0.48\linewidth}
  \centering
  \includegraphics[width=\linewidth]{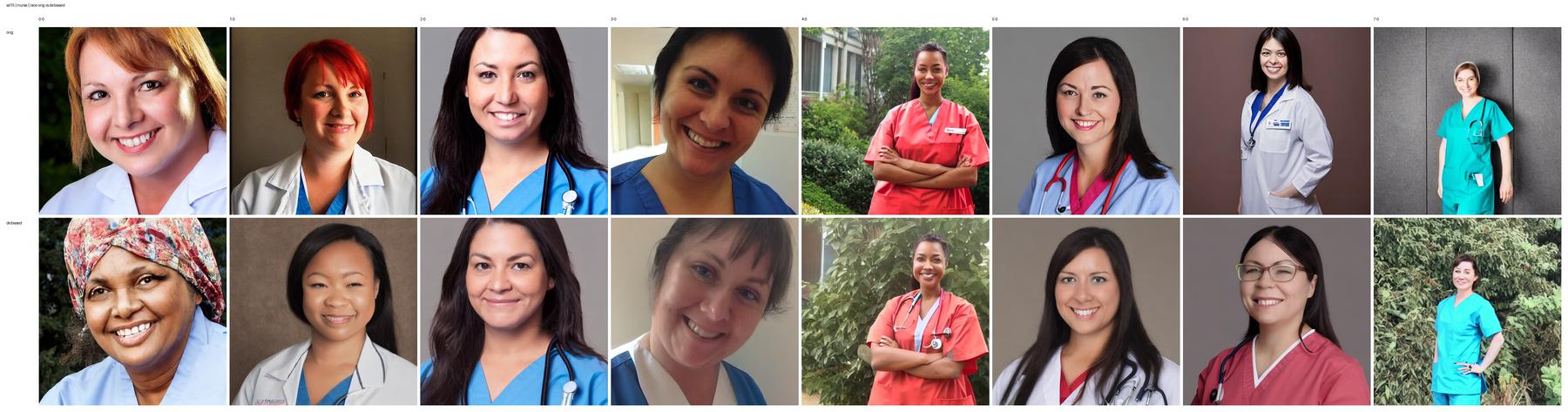}
  \caption{\textit{Nurse}}
\end{subfigure}\hfill
\begin{subfigure}[t]{0.48\linewidth}
  \centering
  \includegraphics[width=\linewidth]{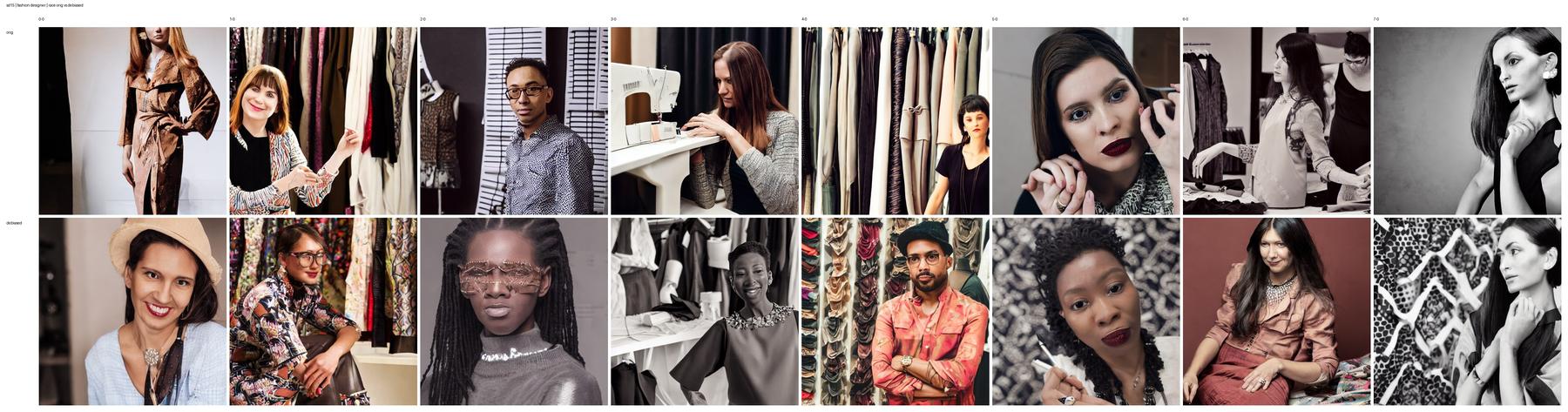}
  \caption{\textit{Fashion Designer}}
\end{subfigure}

\caption{Debiasing of \textit{race} concept on SD-1.5. Top: vanilla SD-1.5, bottom: EquiSteer}
\label{fig:qual_race_sd15}
\end{figure}

\begin{figure}[th!]
\centering
\begin{subfigure}[t]{0.48\linewidth}
  \centering
  \includegraphics[width=\linewidth]{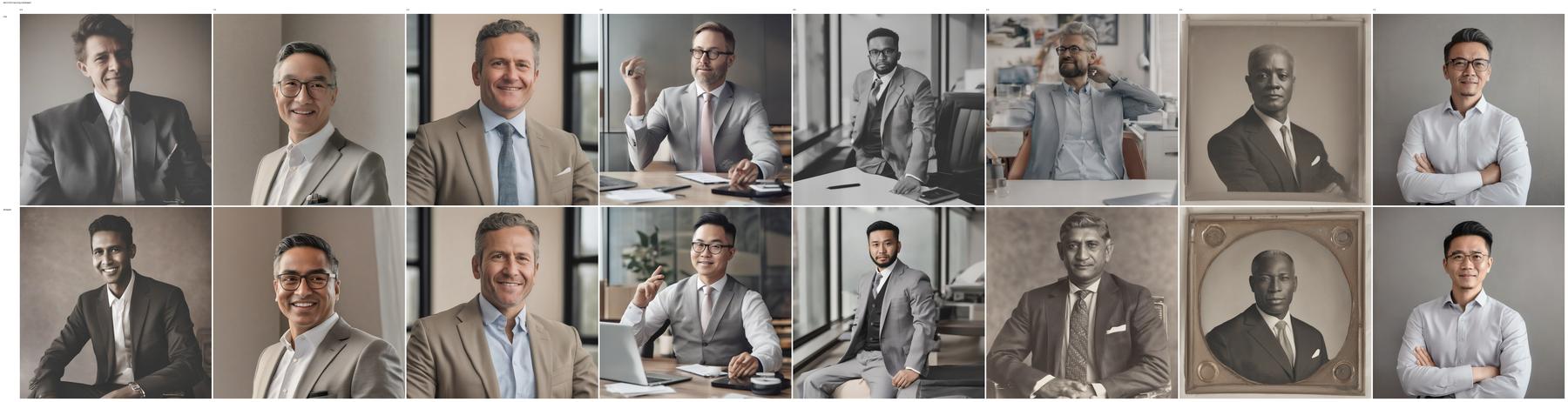}
  \caption{\textit{CEO}}
\end{subfigure}\hfill
\begin{subfigure}[t]{0.48\linewidth}
  \centering
  \includegraphics[width=\linewidth]{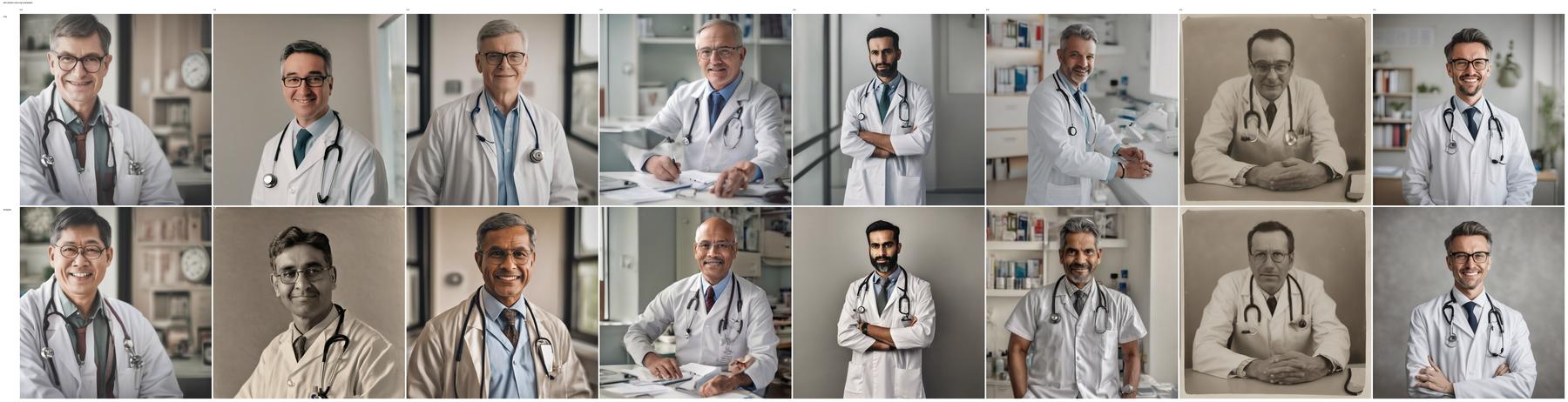}
  \caption{\textit{Doctor}}
\end{subfigure}

\begin{subfigure}[t]{0.48\linewidth}
  \centering
  \includegraphics[width=\linewidth]{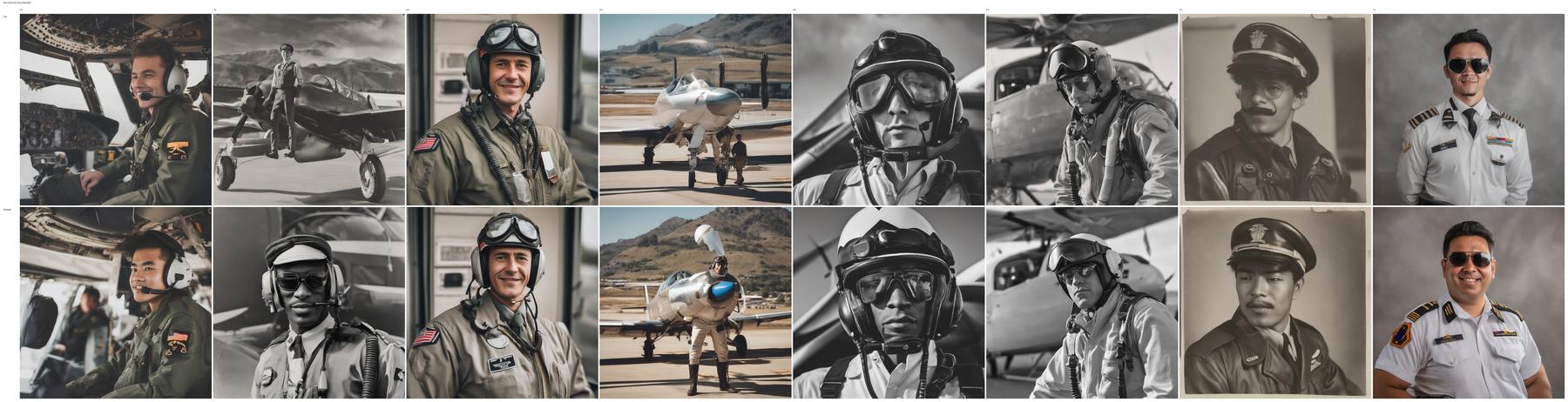}
  \caption{\textit{Pilot}}
\end{subfigure}\hfill
\begin{subfigure}[t]{0.48\linewidth}
  \centering
  \includegraphics[width=\linewidth]{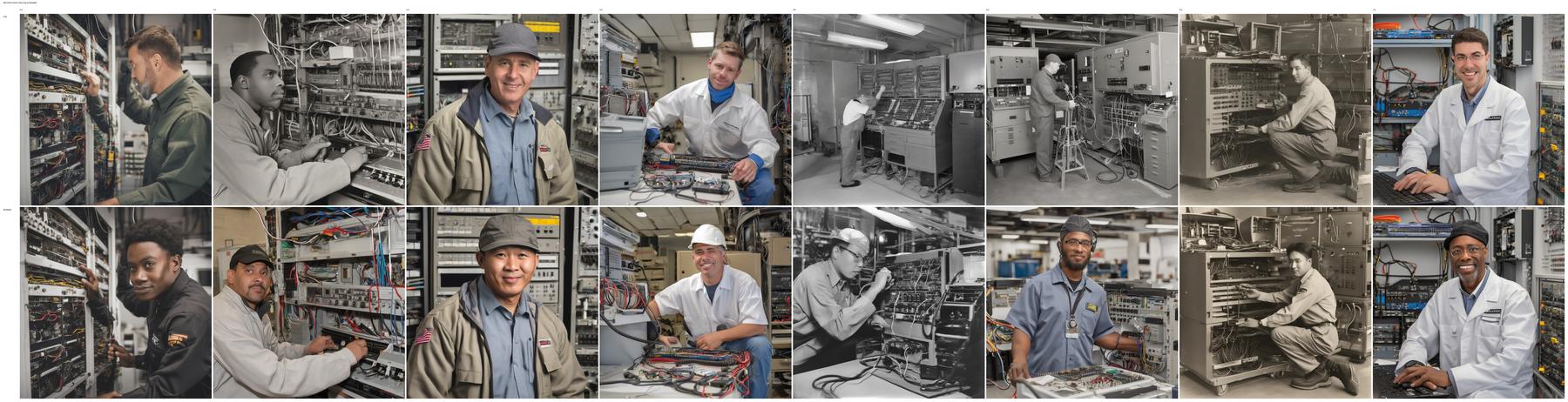}
  \caption{\textit{Technician}}
\end{subfigure}

\begin{subfigure}[t]{0.48\linewidth}
  \centering
  \includegraphics[width=\linewidth]{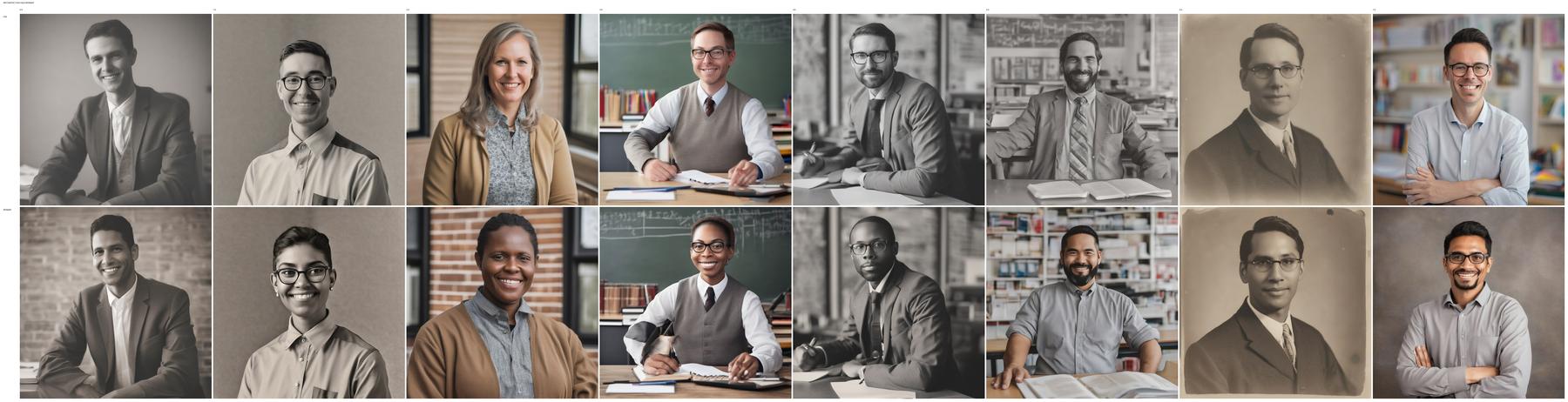}
  \caption{\textit{Teacher}}
\end{subfigure}\hfill
\begin{subfigure}[t]{0.48\linewidth}
  \centering
  \includegraphics[width=\linewidth]{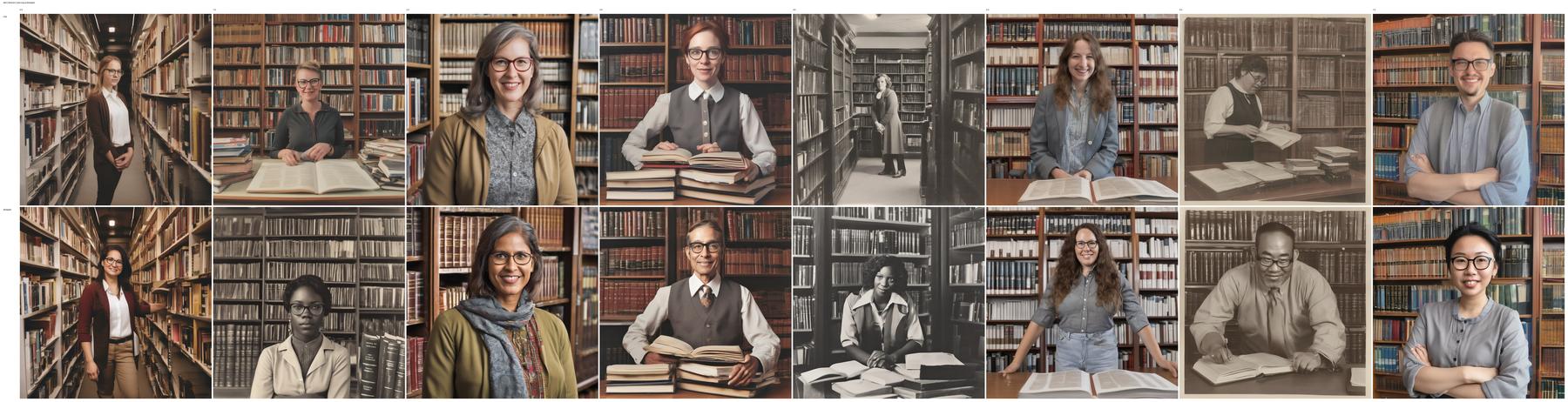}
  \caption{\textit{Librarian}}
\end{subfigure}

\begin{subfigure}[t]{0.48\linewidth}
  \centering
  \includegraphics[width=\linewidth]{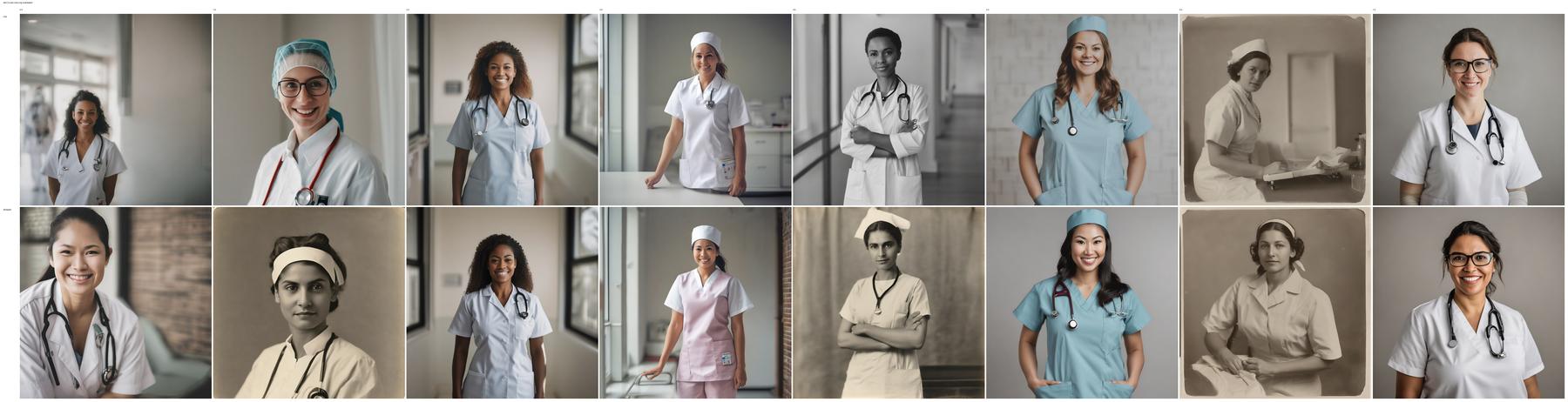}
  \caption{\textit{Nurse}}
\end{subfigure}\hfill
\begin{subfigure}[t]{0.48\linewidth}
  \centering
  \includegraphics[width=\linewidth]{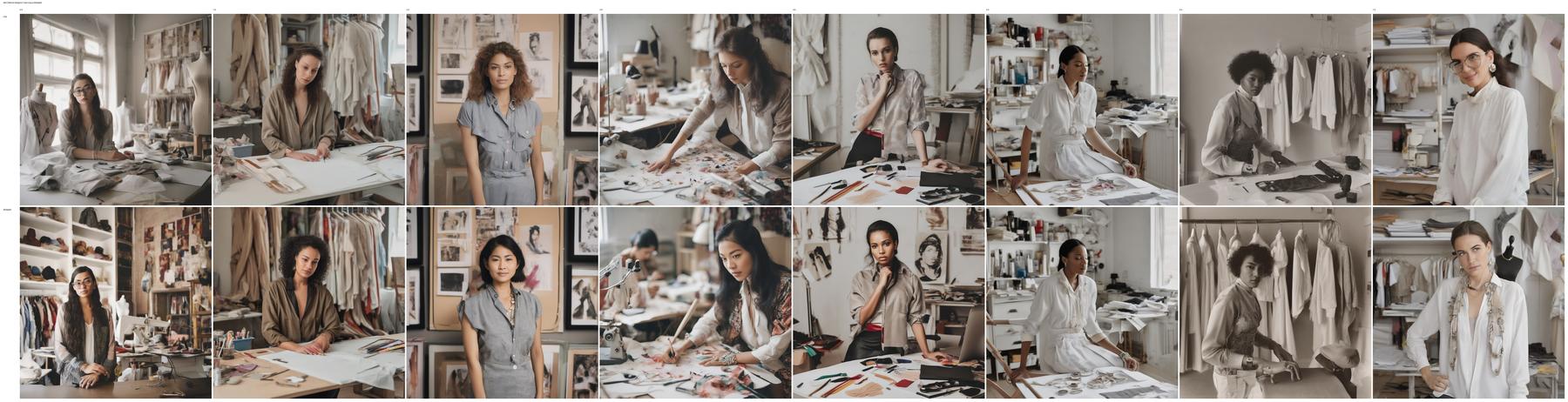}
  \caption{\textit{Fashion Designer}}
\end{subfigure}

\caption{Debiasing of \textit{race} concept on SDXL. Top: vanilla SDXL, bottom: EquiSteer}
\label{fig:qual_race_sdxl}
\end{figure}

\begin{figure}[th!]
\centering
\begin{subfigure}[t]{0.48\linewidth}
  \centering
  \includegraphics[width=\linewidth]{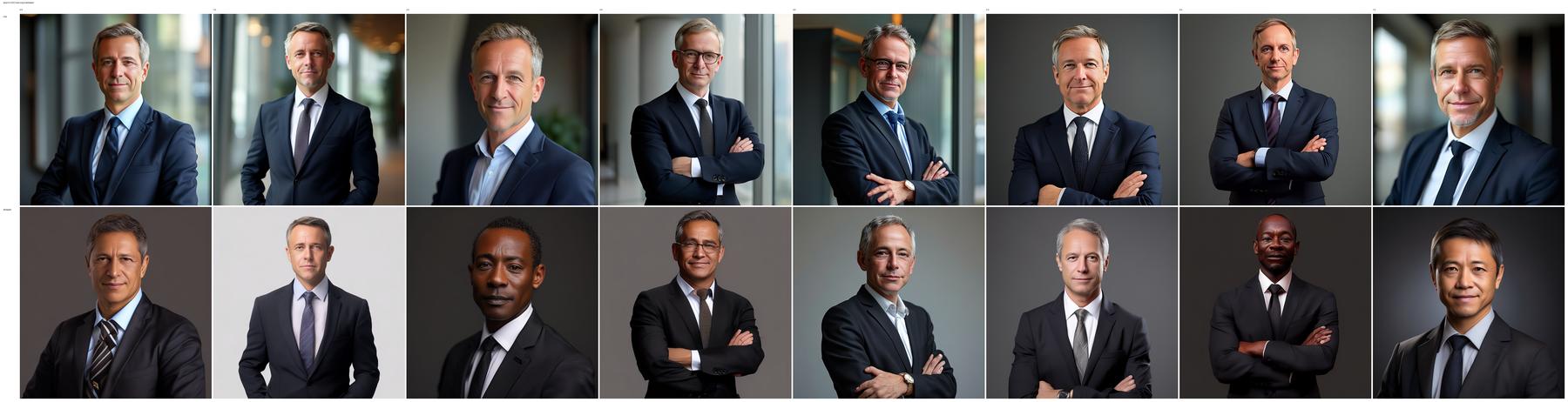}
  \caption{\textit{CEO}}
\end{subfigure}\hfill
\begin{subfigure}[t]{0.48\linewidth}
  \centering
  \includegraphics[width=\linewidth]{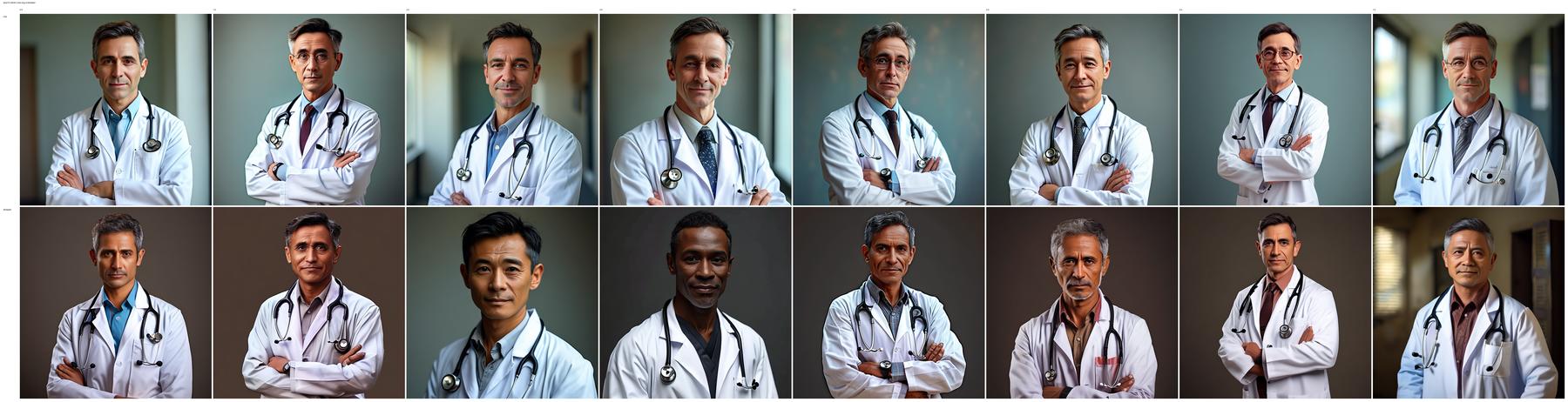}
  \caption{\textit{Doctor}}
\end{subfigure}

\begin{subfigure}[t]{0.48\linewidth}
  \centering
  \includegraphics[width=\linewidth]{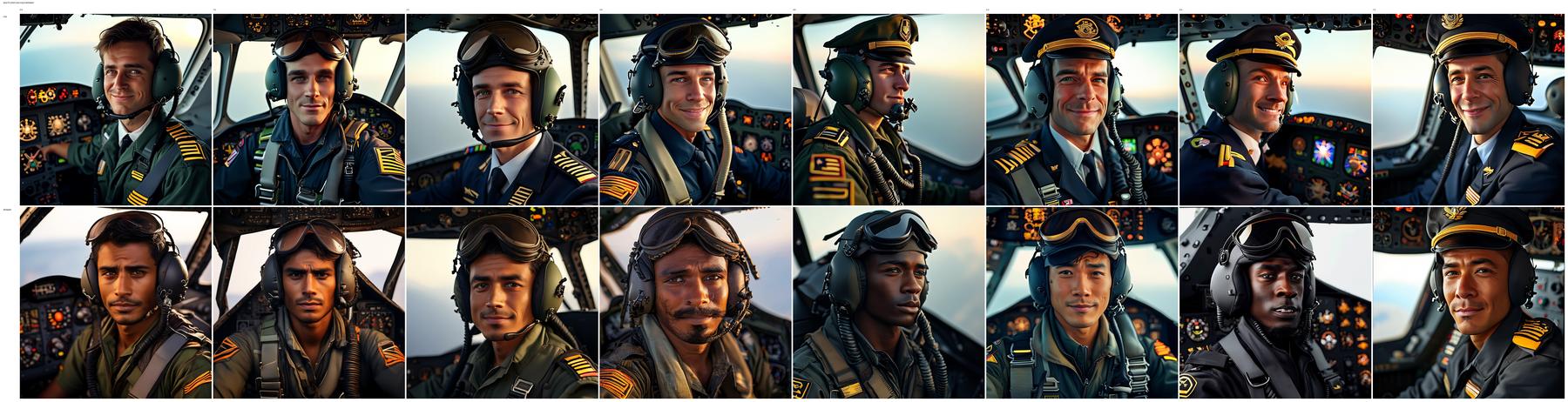}
  \caption{\textit{Pilot}}
\end{subfigure}\hfill
\begin{subfigure}[t]{0.48\linewidth}
  \centering
  \includegraphics[width=\linewidth]{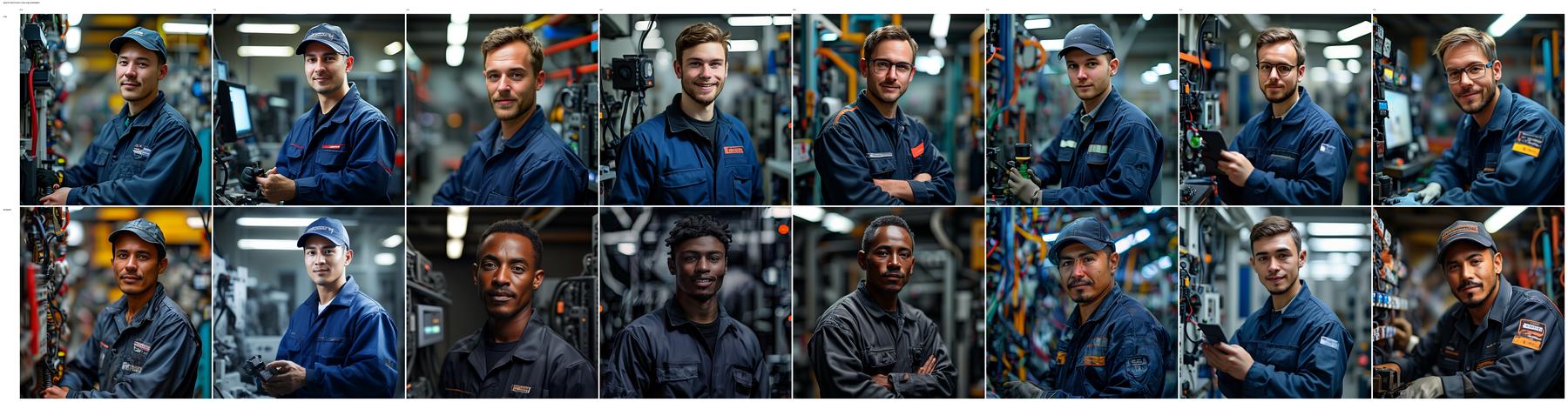}
  \caption{\textit{Technician}}
\end{subfigure}

\begin{subfigure}[t]{0.48\linewidth}
  \centering
  \includegraphics[width=\linewidth]{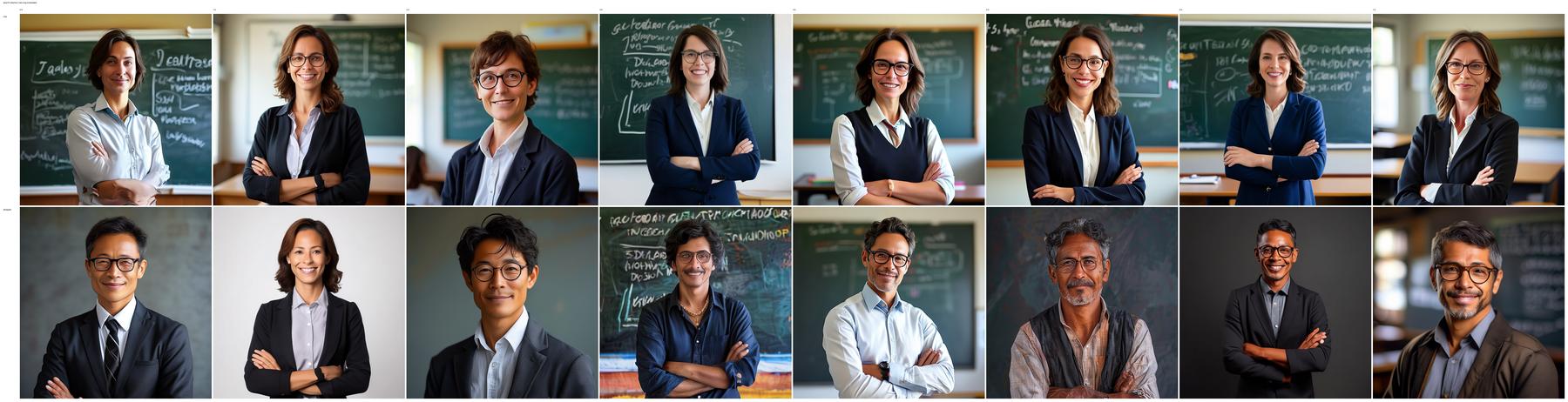}
  \caption{\textit{Teacher}}
\end{subfigure}\hfill
\begin{subfigure}[t]{0.48\linewidth}
  \centering
  \includegraphics[width=\linewidth]{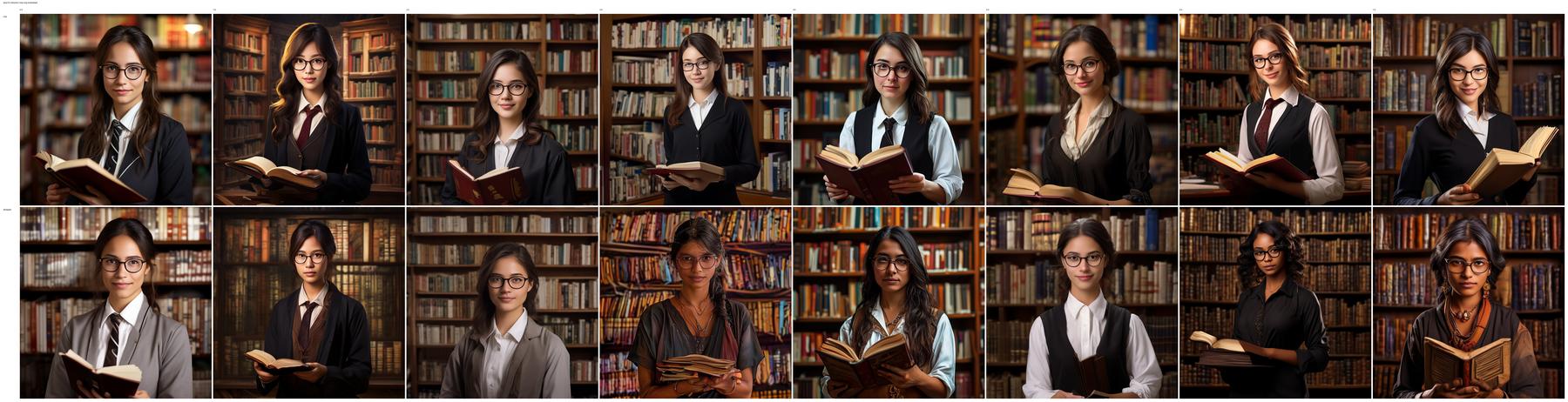}
  \caption{\textit{Librarian}}
\end{subfigure}

\begin{subfigure}[t]{0.48\linewidth}
  \centering
  \includegraphics[width=\linewidth]{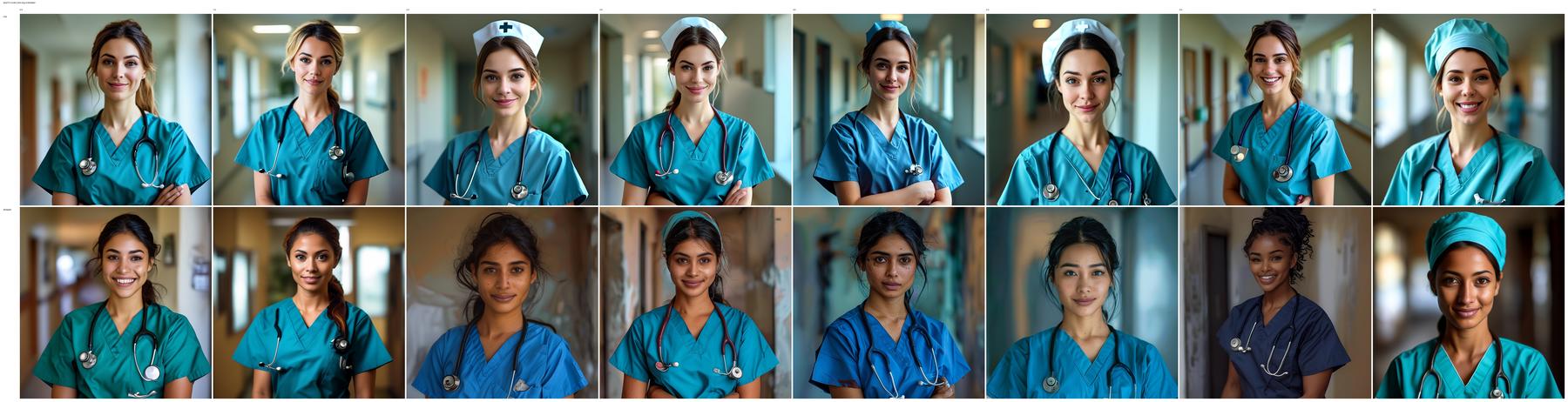}
  \caption{\textit{Nurse}}
\end{subfigure}\hfill
\begin{subfigure}[t]{0.48\linewidth}
  \centering
  \includegraphics[width=\linewidth]{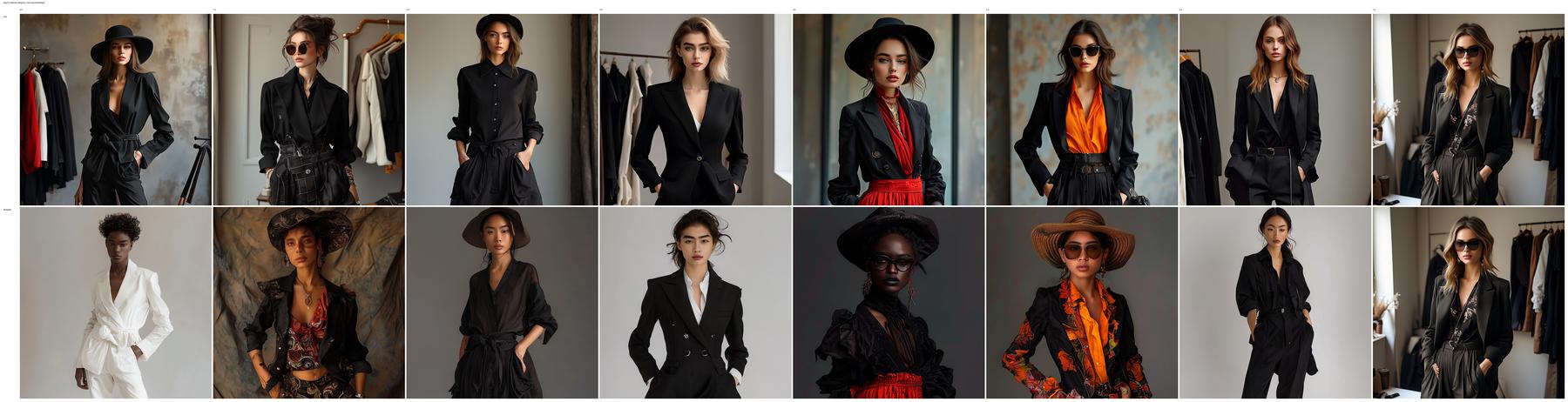}
  \caption{\textit{Fashion Designer}}
\end{subfigure}

\caption{Debiasing of \textit{race} concept on SANA-1.5. Top: vanilla SANA-1.5, bottom: EquiSteer}
\label{fig:qual_race_sana}
\end{figure}

In this section, on Fig.~\ref{fig:qual_race_sd15},~\ref{fig:qual_race_sdxl} and \ref{fig:qual_race_sana} we provide more qualitative results on debiasing \textit{race} concept.

\clearpage
\section{Ablations on EquiSteer components}
\label{sec:ablation_suppl}

In this section we ablate parameters of EquiSteer.

\subsection{EquiSteer components}

We ablate the contribution of EquiSteer’s main components: (i) the basic steering update (Eq.~\ref{eq:fairsteer_base}) with adaptive magnitude selection (Eq.~\ref{eq:thr_add}), (ii) orthogonalisation, and (iii) the gating mechanism (Eq.~\ref{eq:gate}). Tab.~\ref{tab:gender_parity_distance_ablation_v2} reports \textit{gender} debiasing results for three variants:
\begin{itemize}
    \item \textbf{add:} basic steering (Eq.~\ref{eq:fairsteer_base}) with adaptive magnitude (Eq.~\ref{eq:thr_add});
    \item \textbf{add \& erase:} \textbf{add} + orthogonalisation;
    \item \textbf{add \& erase \& gate:} full EquiSteer (adds gating).
\end{itemize}

The results highlight the role of each component. \textbf{add} improves parity on attribute-neutral prompts, but also alters generations for attribute-specific prompts. Adding orthogonalisation (\textbf{add \& erase}) further improves neutrality debiasing, yet substantially degrades attribute preservation, indicating that removing pre-existing attribute signals can conflict with prompts that explicitly specify an attribute. Finally, incorporating the gate (\textbf{add \& erase \& gate}) restores attribute preservation on attribute-specific prompts while maintaining strong debiasing performance on neutral prompts.

\begin{table*}[th!]
  \centering
  \caption{Gender distance-to-parity by profession for SD-1.5 under different EquiSteer components. For neutral prompts, $\Delta = |r-0.5|$ measures deviation from the parity target, so lower is better. For attribute-specific prompts, larger $\Delta$ indicates better preservation of the requested attribute, with the ideal value approaching $0.5$. The rightmost block corresponds to the full EquiSteer method.}
  \label{tab:gender_parity_distance_ablation_v2}
  \scriptsize
  \setlength{\tabcolsep}{2.5pt}
  \renewcommand{\arraystretch}{1.05}
  \begin{tabularx}{\textwidth}{l *{9}{>{\centering\arraybackslash}X}}
  \toprule
  & \multicolumn{3}{c}{add} & \multicolumn{3}{c}{add \& erase} & \multicolumn{3}{c}{add \& erase \& gate} \\
  \cmidrule(lr){2-4}\cmidrule(lr){5-7}\cmidrule(lr){8-10}
  \makecell[l]{Concept} &
  \makecell{neutral\\$\Delta$($\downarrow$)} & \makecell{female\\$\Delta$($\uparrow$)} & \makecell{male\\$\Delta$($\uparrow$)} &
  \makecell{neutral\\$\Delta$($\downarrow$)} & \makecell{female\\$\Delta$($\uparrow$)} & \makecell{male\\$\Delta$($\uparrow$)} &
  \makecell{neutral\\$\Delta$($\downarrow$)} & \makecell{female\\$\Delta$($\uparrow$)} & \makecell{male\\$\Delta$($\uparrow$)} \\
  \midrule
  CEO              & 0.120 & 0.500 & 0.160 & 0.040 & 0.310 & 0.130 & 0.017 & 0.497 & 0.500 \\
  doctor           & 0.020 & 0.430 & 0.450 & 0.010 & 0.390 & 0.210 & 0.000 & 0.487 & 0.500 \\
  fashion designer & 0.080 & 0.490 & 0.480 & 0.060 & 0.390 & 0.330 & 0.004 & 0.499 & 0.500 \\
  librarian        & 0.110 & 0.480 & 0.380 & 0.010 & 0.220 & 0.110 & 0.027 & 0.489 & 0.500 \\
  nurse            & 0.390 & 0.490 & 0.310 & 0.150 & 0.470 & 0.220 & 0.068 & 0.480 & 0.491 \\
  pilot            & 0.210 & 0.500 & 0.440 & 0.100 & 0.490 & 0.410 & 0.084 & 0.500 & 0.490 \\
  teacher          & 0.130 & 0.500 & 0.340 & 0.170 & 0.450 & 0.100 & 0.079 & 0.500 & 0.475 \\
  technician       & 0.260 & 0.460 & 0.470 & 0.220 & 0.430 & 0.330 & 0.125 & 0.480 & 0.500 \\
  \midrule
  Avg.\ $\Delta$   & 0.165 & 0.481 & 0.379 & 0.095 & 0.394 & 0.230 & 0.051 & 0.492 & 0.495 \\
  \bottomrule
  \end{tabularx}
  \end{table*}

\end{document}